\documentclass[journal]{IEEEtran}
\usepackage{amssymb}
\ifCLASSINFOpdf
\else
\fi
\usepackage{enumitem}
\usepackage{amsmath}
\usepackage{graphicx}
\usepackage{multirow}
\usepackage{makecell}
\usepackage{array}
\usepackage{caption}
\usepackage{subcaption}
\usepackage{algorithmic}
\usepackage{algorithm}

\floatname{algorithm}{Algorithm}

\begin{document}

\title{Towards Robust Optical–SAR Object Detection under Missing Modalities: A Dynamic Quality-Aware Fusion Framework}

\author{Zhicheng Zhao, Yuancheng Xu, Andong Lu, Chenglong Li, Jin Tang

\thanks{This work was supported in part by the National Natural Science Foundation of China (No. 62306005 and No. 62406002), in part by the Natural Science Foundation of Anhui Province (No. 2408085MF153 and 2208085J18), and in part by the Natural Science Foundation of Anhui Higher Education Institution (No. 2022AH040014). (Corresponding author: Andong Lu).}
\thanks{Zhicheng Zhao, Yuancheng Xu, and Chenglong Li are with Anhui Provincial Key Laboratory of Security Artificial Intelligence, Information Materials and Intelligent Sensing Laboratory of Anhui Province, School of Artificial Intelligence, Anhui University, Hefei 230601, China. Zhicheng Zhao is also with the 38th Research Institute, China Electronics Technology Group Corporation, Hefei 230088, China. (Email: zhaozhicheng@ahu.edu.cn, wa23301127@stu.ahu.edu.cn, lcl1314@foxmail.com, xiaoyun@ahu.edu.cn).}
\thanks{Andong Lu and Jin Tang are with Anhui Provincial Key Laboratory of Multi-modal Cognitive Computation, School of Computer Science and Technology, Anhui University, Hefei 230601, China. (Email: adlu\_ah@foxmail.com, tangjin@ahu.edu.cn).}}

\markboth{Journal of \LaTeX\ Class Files,~Vol.~13, No.~9, September~2014}%
{Shell \MakeLowercase{\textit{et al.}}: Bare Demo of IEEEtran.cls for Journals}

\maketitle
\begin{abstract}
Optical and Synthetic Aperture Radar (SAR) fusion-based object detection has attracted significant research interest in remote sensing, as these modalities provide complementary information for all-weather monitoring. However, practical deployment is severely limited by inherent challenges. Due to distinct imaging mechanisms, temporal asynchrony, and registration difficulties, obtaining well-aligned optical-SAR image pairs remains extremely difficult, frequently resulting in missing or degraded modality data. Although recent approaches have attempted to address this issue, they still suffer from limited robustness to random missing modalities and lack effective mechanisms to ensure consistent performance improvement in fusion-based detection. To address these limitations, we propose a novel Quality-Aware Dynamic Fusion Network (QDFNet) for robust optical-SAR object detection. Our proposed method leverages learnable reference tokens to dynamically assess feature reliability and guide adaptive fusion in the presence of missing modalities. In particular, we design a Dynamic Modality Quality Assessment (DMQA) module that employs learnable reference tokens to iteratively refine feature reliability assessment, enabling precise identification of degraded regions and providing quality guidance for subsequent fusion. Moreover, we develop an Orthogonal Constraint Normalization Fusion (OCNF) module that employs orthogonal constraints to preserve modality independence while dynamically adjusting fusion weights based on reliability scores, effectively suppressing unreliable feature propagation. Extensive experiments on the SpaceNet6-OTD and OGSOD-2.0 datasets demonstrate the superiority and effectiveness of QDFNet compared to state-of-the-art methods, particularly under partial modality corruption or missing data scenarios.
\end{abstract}

\begin{IEEEkeywords}
Optical-SAR fusion, Object detection, Missing modality, Remote sensing image
\end{IEEEkeywords}

\section{Introduction}
\IEEEPARstart{I}{n} recent years, optical-SAR object detection has drawn increasing attention in the field of remote sensing \cite{a,1}. Owing to the inherent complementarity between these modalities, optical imagery provides rich spectral and textural cues \cite{2,3}, while SAR imagery offers all-weather and day-night imaging capability that is resilient to illumination and atmospheric variations \cite{4,5}. Leveraging both modalities enables more comprehensive and robust object perception, significantly improving detection performance under complex environmental conditions \cite{6,7}. Consequently, optical-SAR object detection has demonstrated great potential in critical applications such as disaster response, environmental monitoring, and military reconnaissance \cite{8}.
\begin{figure}[t]
    \centering
    \includegraphics[width=1\linewidth]{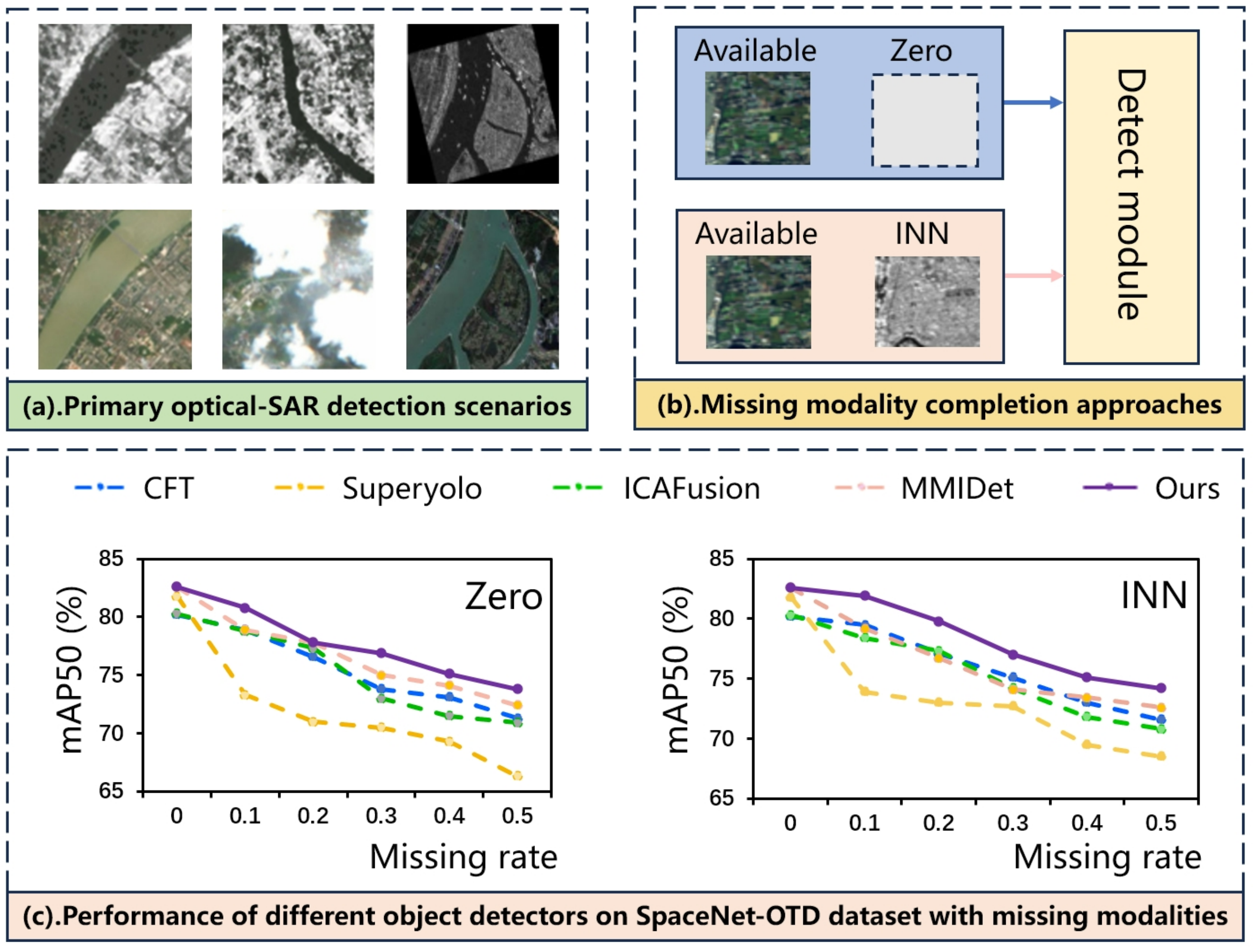}
    \caption{Illustration of two missing modality processing approaches and performance comparison of state-of-the-art methods on the SpaceNet6-OTD-Fog dataset \cite{19}.  (a) 
    Primary optical-SAR detection scenarios under various conditions. (b) Demonstration of two modality completion strategies: Zero (zero-filling for missing modalities) and INN (Inverse Design Network-based reconstruction). (c) Comparison of mAP50 performance for different object detectors (CFT~\cite{22}, SuperYOLO~\cite{16}, ICAFusion~\cite{25}, MMIDet~\cite{26}, and Ours) on the SpaceNet6-OTD-Fog dataset under increasing modality missing rates (0-0.5).}
    \label{fig:1}
\end{figure}

However, practical deployment faces significant challenges arising from real-world data imperfections. As illustrated in Figure~\ref{fig:1} (a), optical imagery is often obscured by cloud coverage, while discrepancies in imaging mechanisms, sensor viewing geometry, and acquisition-time asynchrony, along with the inherent difficulty of achieving accurate co-registration, frequently result in spatial misalignment between optical and SAR data. These factors restrict the availability of optical-SAR pairs. Consequently, most existing studies have focused on addressing alignment and reconstruction challenges. For example, Hong et al. \cite{9} propose a spatio-temporal partitioning method with multiple linear regression to reconstruct cloud-contaminated optical images, and Xiao et al. \cite{10459210} develop an affine deformable registration network to achieve multimodal image alignment. Despite these advancements, the missing modality problem has received insufficient attention in the optical-SAR object detection community.

One modality may be unavailable during inference due to acquisition failures, data corruption, or transmission losses. However, most existing dual-modality object detection methods \cite{11,12} assume fully aligned and complete multimodal inputs, limiting their effectiveness when a modality is missing or degraded. For instance, SuperYOLO \cite{16} performs feature concatenation at the input level, while E2E-MFD \cite{18} jointly optimizes fusion and detection in an end-to-end manner. Unfortunately, these approaches lack explicit mechanisms to address modality degradation, making them vulnerable to real-world imperfections such as sensor failure or atmospheric interference \cite{9,10}.
As illustrated in Figure~\ref{fig:1} (b), a straightforward strategy for handling missing modalities is to apply zero-filling or modality reconstruction.

Nevertheless, even when employing simple zero-filling or advanced invertible neural networks (INNs) for modality reconstruction, existing methods still exhibit substantial performance degradation under missing modality, as illustrated in Figure~\ref{fig:1} (c).
This observation empirically confirms that current fusion-based detectors struggle to extract discriminative representations from corrupted or reconstructed modalities and lack mechanisms to ensure stable performance under modality uncertainty.
Several recent efforts have explored robustness against modality perturbations in other domains \cite{Peng2022,9578794,yu2020}. For example, LF-MDet \cite{19} introduces adaptive feature modulation to mitigate single-modality dependencies and improve resistance to adversarial perturbations in RGB-infrared detection. Similar ideas have also been applied in cross-modal medical image fusion \cite{wei2023}. For instance, CT-PET-Fuse \cite{10975066} employs a cross-modal attention mechanism to reduce single-modality reliance and enhance robustness to noise perturbations in CT-PET fusion.
However, these methods are difficult to generalize to optical-SAR scenarios due to fundamentally different imaging mechanisms and the absence of explicit cross-modal quality assessment. Consequently, when a modality is missing or severely degraded, existing fusion-based frameworks struggle to extract discriminative features, leading to unreliable fusion quality and degraded detection performance.

To address these challenges, we propose a novel quality-aware multimodal fusion framework that dynamically assesses and adaptively integrates features from missing modalities. 
Specifically, we design a Dynamic Modality Quality Assessment (DMQA) module to extract reliable and discriminative features from compensated or degraded modalities.
Unlike conventional fusion methods that treat all modalities equally, DMQA introduces a set of learnable reference tokens that iteratively interact with input features to evaluate their reliability.
In each iteration, feature quality is evaluated from two complementary perspectives: magnitude stability, measured by the relative variance of feature vector norms to capture abnormal fluctuations \cite{20}, and directional consistency, evaluated through maximum cosine similarity with learned reference patterns to identify semantic deviations \cite{21}. 
By integrating both indicators \cite{wu2024}, DMQA precisely locates unreliable regions exhibiting significant magnitude or directional inconsistencies, thereby providing explicit quality guidance for subsequent fusion.

To further ensure robust integration across varying modality qualities, we develop an Orthogonal Constraint Normalization Fusion (OCNF) module that performs quality-aware adaptive fusion while preserving modality-specific information. OCNF employs modality-specific orthogonal weight matrices to project features into independent subspaces, effectively preventing feature space collapse and preserving modality-specific discriminative information \cite{baltruaitis2017multimodal}. Guided by reliability scores from DMQA, OCNF dynamically adjusts fusion weights, suppressing unreliable generated features and emphasizing trustworthy modality information. A normalization mechanism is further introduced to stabilize the fusion process by balancing feature scales.

In summary, this study addresses the critical problem of modality missing in optical-SAR object detection and proposes a quality-aware multimodal fusion framework tailored for real-world remote sensing scenarios. The major contributions are summarized as follows:
\begin{itemize}
\item We systematically investigate the underexplored issue of missing modalities in optical-SAR object detection and establish a quality-aware detection paradigm capable of handling missing multimodal inputs.

\item  We propose a Dynamic Modality Quality Assessment module that evaluates feature reliability from both magnitude and directional perspectives through iterative interaction between learnable reference tokens and input features.

\item We design an Orthogonal Constraint Normalization Fusion module that preserves modality-specific information while dynamically adjusting fusion weights based on reliability assessments.

\item Extensive experiments on two standard datasets demonstrate that QDFNet achieves superior robustness and detection accuracy under degraded or missing modality conditions.
\end{itemize}

\section{RELATED WORK}
\subsection{Multimodal Object Detection in Remote Sensing}
Multimodal object detection has become an essential component of modern remote sensing systems, as the integration of complementary modalities significantly enhances perception under complex environmental conditions. Optical-SAR fusion, in particular, has demonstrated strong potential for robust target detection. Fang et al.~\cite{22} proposed a cross-modal fusion Transformer that leverages cross-attention interactions to exchange global contextual information across modalities, while Liu et al.~\cite{23} introduced TarDAL, which employs a dual-layer optimization strategy to preserve discriminative modality-specific features. Zhang et al.~\cite{24} designed TINet with illumination-guided feature weighting and dual attention mechanisms, and Shen et al.~\cite{25} presented ICAFusion, integrating complementary information through iterative learning for efficient representation.

Although these methods achieve promising performance under ideal conditions with well-aligned and complete multimodal   inputs, they degrade significantly when facing modality degradation or absence. Recent efforts, such as MMI-Det~\cite{26} and LF-MDet~\cite{18}, enhance modality-invariant representation and robustness through information-guided optimization or frequency-domain learning. However, most existing detectors lack explicit mechanisms to assess or adapt to feature quality variations, limiting their effectiveness in real-world scenarios where missing or corrupted modalities frequently occur. These limitations motivate the development of a quality-aware framework that explicitly models modality reliability during detection.

\subsection{Multimodal Image Fusion Methods}
Image fusion plays a vital role in integrating complementary information from multiple sources for diverse vision applications \cite{27,28}. Recent years have witnessed significant progress with various advanced fusion strategies. In the optical-infrared domain, CDDFuse \cite{29} adopts a dual-branch Transformer-CNN framework with a correlation-driven loss for feature separation and fusion, while CrossFuse \cite{li2024crossfuse} introduces cross-attention to enhance modality complementarity and MH-FUNet \cite{30} employs self-supervised learning to improve robustness. In multi-spectral fusion, Yao et al. \cite{31} proposed a hierarchical Laplacian pyramid network preserving pixel-level saliency, and Yu et al. \cite{32} developed an unsupervised fusion model with group convolutions and 3D attention for hyperspectral-multispectral integration. For optical-SAR fusion, VSFF \cite{33} achieves complementary feature decomposition, Shao et al.~\cite{a} designed a pixel-saliency-based algorithm, and Zhang et al.~\cite{3} fused SAR texture with optical high-frequency details via wavelet transforms.

However, these fusion methods typically assume well-aligned and complete multimodal inputs, limiting their robustness in practical remote sensing scenarios. Challenges including resolution gaps, temporal misalignment, and modality degradation frequently arise under real conditions, yet existing approaches lack dynamic mechanisms to evaluate input quality or adaptively adjust fusion strategies. For example, current frameworks seldom assess feature reliability or modulate fusion weights when encountering missing or corrupted modalities, often leading to performance drops. To overcome these limitations, this paper proposes a quality-aware fusion framework that introduces learnable feature tokens and an orthogonal adaptive fusion module. Our method achieves real-time quality assessment and reliability-aware fusion, effectively preserving information integrity even with missing data. This work presents a novel integration of dynamic quality evaluation into end-to-end multimodal fusion, particularly suited for complex and degradation-prone imaging environments.

\subsection{Multimodal Learning under Modality Missing}
Multimodal learning under modality missing has emerged as a crucial research focus across multiple domains. To address this challenge, researchers have proposed various approaches in the literature. These approaches can be systematically categorized into two paradigms: generative methods and joint learning methods.\\
\noindent \textbf{Generative methods:} These approaches learn data distributions to synthesize missing modalities and compensate for information gaps~\cite{DeepMIH,xie2024generative,Yeh_2017_CVPR}. Kingma et al. \cite{36} introduced an invertible network for direct cross-modal mapping, while Zhang et al. \cite{37} adopted an adversarial strategy to generate representations of missing views. Shang et al. \cite{38} utilized GAN-based domain mappings and multimodal denoising autoencoders for view completion. Zhou et al. \cite{39} developed a feature enhancement generator to reconstruct 3D feature representations of absent modalities. DiCMoR \cite{40} applied category-based normalizing flows for accurate cross-modal distribution transformation, and IMDer \cite{41} leveraged score diffusion models for missing modality recovery. \\
\noindent\textbf{Joint learning methods:} These approaches fuse available modalities in latent space to learn shared representations for efficient multimodal integration~\cite{Liu_Wei_Lu_Sun_Wang_Zheng_2023,10906480}. Zhao et al. \cite{42} proposed a missing-modality imagination network for prediction using existing modalities, while Wei et al. \cite{43} introduced a separable multimodal learning framework to capture cross-modal complementarity. MTMSA \cite{44} addressed uncertain missing modalities via modality conversion and pre-training supervision, translating visual and auditory inputs into textual space. TMFormer \cite{45} modeled intra- and inter-modal dependencies by treating modalities as variable-length token sequences. DrFuse \cite{46} learned shared and unique representations of EHR and CXR data using a disease-aware attention mechanism, and PASSION \cite{47} applied self-distillation with preference-aware regularization to handle missing medical image segmentation. \\
\indent Although prior research has achieved remarkable results, it overlooks the detrimental impact of low-quality modalities on model performance. Therefore, we propose a framework that quantifies feature reliability through iterative refinement, preserves modality uniqueness via orthogonal constraints, and enables robust fusion through quality-aware adaptive weighting.

\section{PROPOSED METHOD}

\begin{figure*}[htb]
    \centering
    \includegraphics[width=\textwidth]{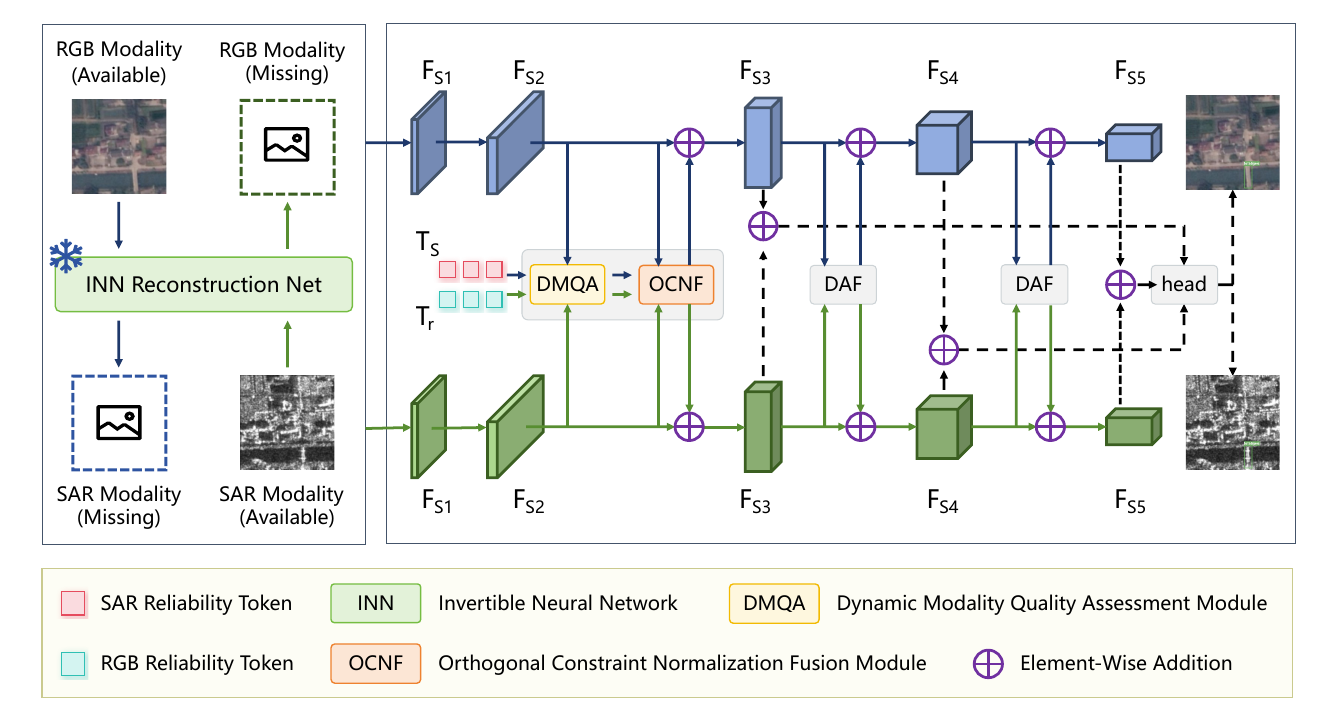}
    \caption{Framework of the proposed QDFNet. INN is a pre-trained invertible neural network used to generate randomly missing modalities~\cite{xing2021,decoupling2021}. The yellow region represents the DMQA module, which evaluates the reliability of multimodal features by quantifying feature quality across both magnitude and directional dimensions. The orange region denotes the OCNF module, which implements orthogonally constrained fusion of multimodal features and dynamically adjusts the fusion weights of different modalities based on the quality assessment provided by DMQA.}
    \label{fig:2}
\end{figure*}

In this section, we first present the overall architecture in Section III-A, followed by a detailed description of the Dynamic Modality Quality Assessment (DMQA) and Orthogonal Constraint Normalization Fusion (OCNF) modules in Sections III-B and III-C, respectively. Finally, the training objective is presented in Section~III-D.

\subsection{Network Overview}
To tackle the challenge of missing modalities in cross-modal object detection between optical and synthetic aperture radar (SAR) images, we propose a novel detection framework that integrates three core modules: an Invertible Neural Network (INN) preprocessing module, a Dynamic Modality Quality Assessment (DMQA) module, and an Orthogonal Constraint Normalization Fusion (OCNF) module. As illustrated in Figure~\ref{fig:2}, the framework takes optical and SAR images as input, where the pretrained INN preprocessing module first performs modality completion, reconstructing the missing optical or SAR counterparts to ensure complete modality pairs. The completed images are then fed into a dual-stream backbone network to extract modality-specific deep features. Subsequently, the DMQA module introduces an iterative token computation mechanism that dynamically identifies and refines unreliable regions within the feature maps through n-sampling iterations, thereby enhancing the robustness and consistency of cross-modal representations. During the fusion stage, the OCNF module applies an orthogonal constraint strategy to adaptively balance the contributions of different modalities, achieving effective and complementary feature fusion. The fused features are finally passed to the detection head to generate the final object detection results. The overall framework is trained in a step-wise manner, where the pretrained INN is fixed during the initial stages and the subsequent modules are jointly optimized for cross-modal detection.

\subsection{Dynamic Modality Quality Assessment}
Multimodal object detection relies heavily on the completeness and reliability of individual modalities.  Although generative approaches such as invertible neural networks (INNs) have been adopted to reconstruct absent modalities, the reconstructed features inevitably suffer from cascading reconstruction errors that cause local distortions in the synthesized content. More critically, existing reconstruction-based frameworks lack an explicit mechanism for estimating the reliability of synthesized features. As a result, unreliable regions are fused with equal importance as authentic ones, ultimately compromising fusion quality and detection robustness. 

To address these challenges, we propose a Dynamic Modality Quality Assessment (DMQA) mechanism that adaptively evaluates feature reliability through iterative token-based interaction. DMQA is motivated by the principle that reliable feature vectors tend to exhibit stable magnitude and consistent directional patterns. By jointly evaluating magnitude stability and directional consistency, DMQA provides fine-grained spatial reliability maps that guide subsequent quality-aware fusion.

\begin{figure}[htb]
    \centering
    \includegraphics[width=1\linewidth]{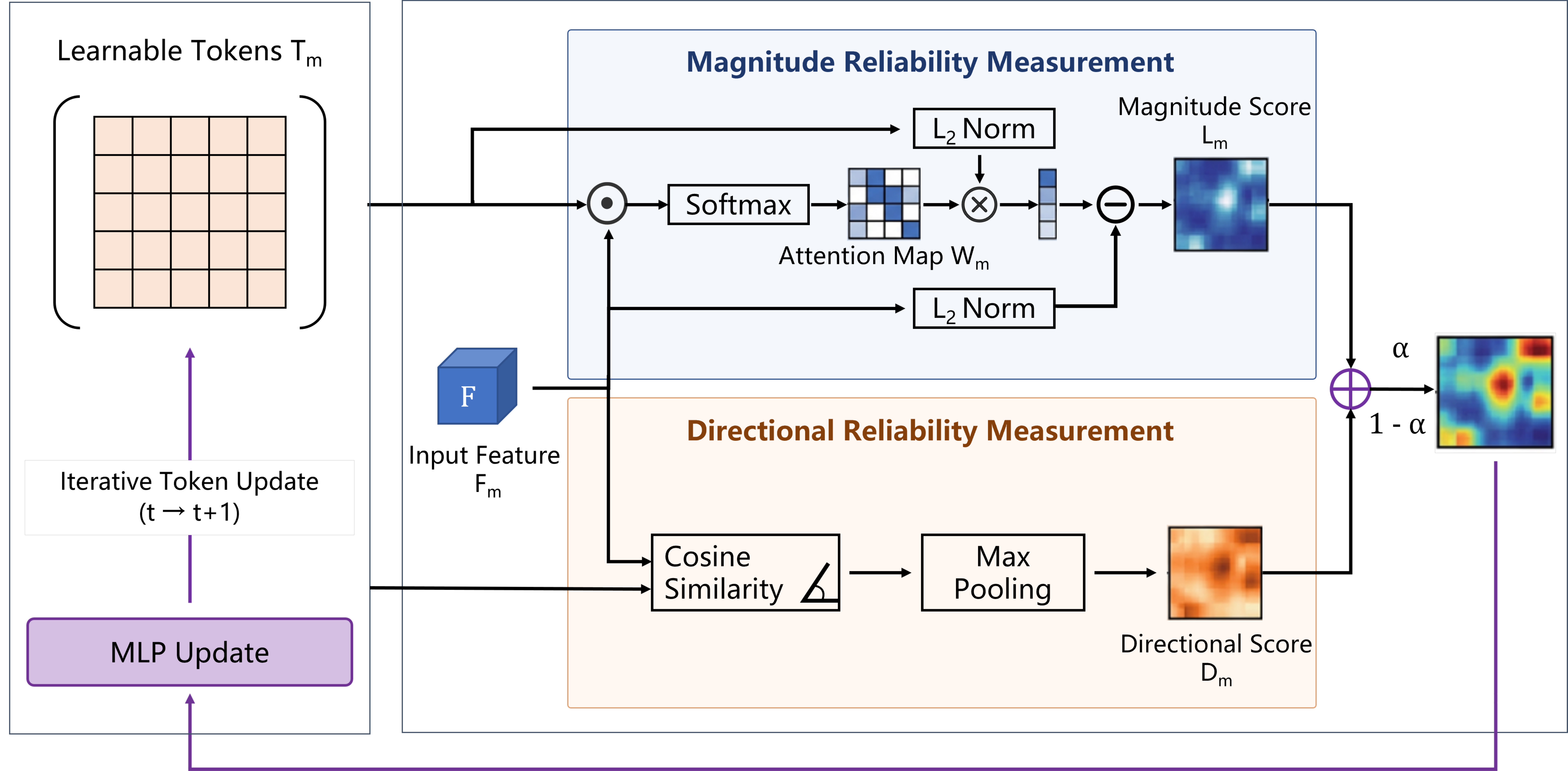}
    \caption{Overall workflow for the DMQA block which adaptively evaluates feature reliability through iterative token-based interaction by jointly measuring magnitude stability and directional consistency, providing fine-grained spatial reliability maps to guide quality-aware fusion.}
    \label{fig:3}
\end{figure}

As illustrated in Figure~\ref{fig:3}, given RGB and SAR modality feature maps denoted as $F_R, F_S \in \mathbb{R}^{B \times N \times C}$, where $B$ is the batch size, $N$ denotes the number of spatial locations, and $C$ is the feature dimension. To precisely identify unreliable regions characterized by inconsistent feature magnitudes and directional instability, we establish a constraint mechanism based on learnable reliability tokens $\mathbf{T}_{R}^{(0)}, \mathbf{T}_{S}^{(0)} \in \mathbb{R}^{K \times C}$ for RGB and SAR modalities, respectively. Where $K$ denotes the number of learnable reliability tokens.\\ 

\noindent{\textbf{Magnitude Reliability Measurement}}.
These tokens are optimized end-to-end to capture statistically stable and discriminative feature distributions across the training data, serving as adaptive reference patterns.
At iteration $t$, the similarity between the input features and the reference tokens is calculated as:
\begin{equation}
    \mathbf{W}_{m}^{(t)}(i)=\operatorname{softmax}\left(\frac{\mathbf{F}_{m}(i) \cdot\left(\mathbf{T}_{m}^{(t)}\right)^{T}}{\sqrt{C}}\right).
\end{equation}
Where \( m \in \{R, S\} \) denotes the modality type, with $R$
representing the RGB modality and $S$ representing the SAR modality. Here, \( \mathbf{W}_m^{(t)}(i) \in \mathbb{R}^{1 \times K} \) represents the attention weights at spatial position \( i \), computed using a scaled dot-product attention mechanism with \( C \) as the normalization factor. These weights quantify the alignment strength between the input features and the learned reference patterns, providing a reliable indicator of local feature consistency for the subsequent adaptive fusion process.
Based on the similarity weights, the token-guided expected feature magnitude is estimated as a weighted sum of the $\ell_{2}$-norms of the corresponding token vectors. For the reference modality, this is defined as:
\begin{equation}
    \mathbf{l}_{m}^{(i)}=\sum_{k=1}^{K} \mathbf{W}_{m}^{(t)}(i)[k] \cdot\left\|\mathbf{T}_{m}^{(t)}(k)\right\|_{2}.
\end{equation}
The deviation between the actual feature magnitude and its token-guided expectation reflects the inconsistency between the observed feature and the prediction implied by the semantically matched tokens. For each modality, the deviation is computed as:
\begin{equation}
    \delta_{m}^{(t)}(i)=\left|\left\|\mathbf{F}_{m}(i)\right\|_{2}-\mathbf{l}_{m}^{(i)}\right|.
\end{equation}
To ensure scale invariance within each batch, these deviations are normalized by the maximum deviation across all spatial positions, with a small constant $\epsilon$ added for numerical stability:
\begin{equation}
    \hat{\delta}_{m}^{(t)}(i)=\frac{\delta_{m}^{(t)}(i)}{\max _{j} \delta_{m}^{(t)}(j)+\epsilon}.
\end{equation}
Finally, the token-aware magnitude reliability is defined as the complement of the normalized deviation, assigning higher reliability scores to features that are more consistent with their token-based expectations:
\begin{equation}
\mathbf{L}_{m}^{(t)}(i)=1-\hat{\delta}_{m}^{(t)}(i).
\end{equation}
Here, \( \mathbf{L}_m^{(t)}(i) \) denotes the spatial-wise reliability score for modality \( m \), bounded within the interval $[0,1]$. A low relative variance, reflecting consistent feature strengths, results in a reliability score that tends toward unity. Conversely, a high degree of variability among feature lengths results in a reduced reliability score, reflecting lower spatial consistency and potential unreliability of features.\\

\noindent{\textbf{Directional Reliability Measurement}}.
Since magnitude alone may fail to capture semantic distortions, we further measure directional consistency using cosine similarity. Specifically, we compute the angular correlation between each spatial feature vector and all learned reference tokens. For each modality, the directional reliability score $D_m^{(t)}(i)$ for modality $m$ at position $i$ is defined as the maximum cosine similarity, indicating the closest semantic alignment between the feature and the token set:
\begin{equation}
    \mathbf{D}_{m}^{(t)}(i)=\max _{k} \frac{\mathbf{F}_{m}(i) \cdot \mathbf{T}_{m}^{(t)}(k)}{\left\|\mathbf{F}_{m}(i)\right\|_{2} \cdot\left\|\mathbf{T}_{m}^{(t)}(k)\right\|_{2}}.
\end{equation}
To obtain a unified reliability metric that reflects both magnitude and directional consistency, we define the joint reliability weight as a convex combination of the two measures. For the reference modality:
\begin{equation}
\mathbf{R}_{m}^{(t)}(i)=\alpha \mathbf{L}_{m}^{(t)}(i)+(1-\alpha) \mathbf{D}_{m}^{(t)}(i).
\end{equation}
Where $\alpha$ is a learnable parameter that controls the relative 
importance of magnitude and direction-based assessments. Directional consistency is derived from the maximum cosine similarity with the token set, indicating the most semantically aligned reference direction. High similarity implies stable and meaningful features. Conversely, if a feature vector shows low similarity to all reference directions, it may indicate that the feature does not contain explicit semantic information or is severely disturbed by noise.

To progressively refine the token representations, we incorporate these reliability scores into an iterative attention update. For each modality, the reliability-weighted attention map is computed as:
\begin{equation}
\mathbf{A}_{m}^{(t)}=\operatorname{softmax}\left(\frac{\mathbf{T}_{m}^{(t)}\left(\mathbf{F}_{m}\right)^{\top}}{\sqrt{C}} \odot \mathbf{R}_{m}^{(t)}\right).
\end{equation}
Where $\odot$ denotes element-wise modulation with broadcasting along the attention map.

The updated token representations are then obtained by applying a lightweight multi-layer perceptron (MLP) transformation to the reliability-weighted aggregated features and adding the result to the previous token embeddings via residual connection:
\begin{equation}
    \mathbf{T}_{m}^{(t+1)}=\operatorname{MLP}\left(\mathbf{A}_{m}^{(t)} \mathbf{F}_{m}\right)+\mathbf{T}_{m}^{(t)}.
\end{equation}
Finally, the final reliability scores combine evolved length and direction assessments after $I$ iterations:
\begin{equation}
    R_{m}=\beta L_{m}^{(I)}+(1-\beta) D_{m}^{(I)}.
\end{equation}
Where $\mathbf{L}_{m}^{(I)}$ represents the length reliability based on final iteration tokens, while $ \mathbf{D}_{m}^{(I)}$ denotes the corresponding direction reliability scores. The parameter $\beta$ is learnable and controls the trade-off between length and direction-based reliability. This dual-metric formulation provides a principled and robust mechanism for identifying high-quality features across modalities, thereby facilitating more effective multimodal fusion in remote sensing scenarios.

\begin{figure}[htb]
    \centering
    \includegraphics[width=1\linewidth]{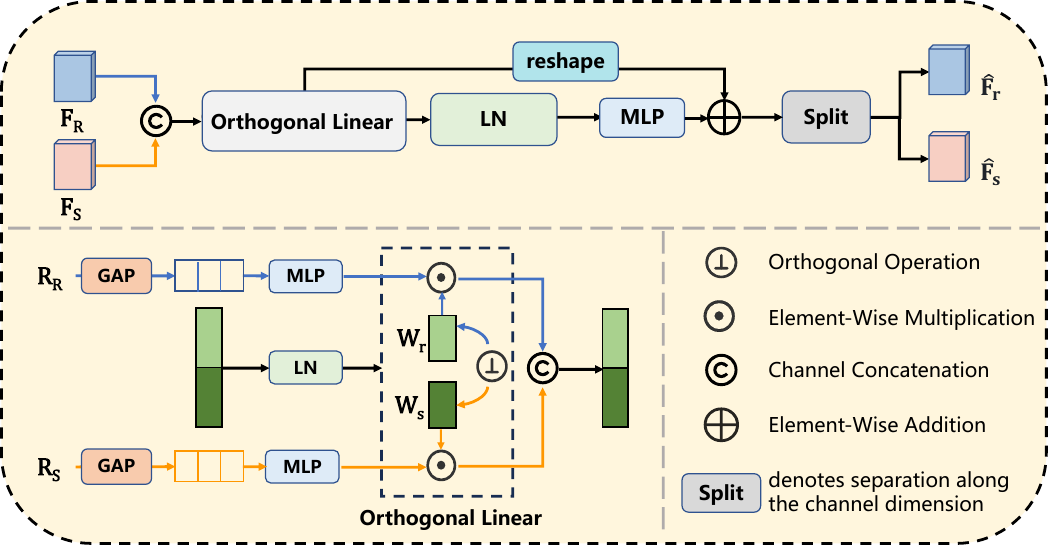}
    \caption{Overall workflow for the OCNF module enforces feature independence through orthogonal weight normalization and integrates quality-aware weighting guided by reliability scores to fuse features, preventing the propagation of low-quality features and noise.}
    \label{fig:4}
\end{figure}

\subsection{Orthogonal Constraint Normalization Fusion}
In multimodal scenarios with missing or degraded inputs, low-quality features can propagate across modalities during fusion and severely impair detection accuracy. To mitigate this issue, we propose the Orthogonal Constraint Normalization Fusion (OCNF) module, which enforces feature independence through orthogonal weight normalization and integrates quality-aware weighting guided by DMQA.

As illustrated in Figure~\ref{fig:4}, OCNF takes the optical and SAR feature maps $F_R, F_S \in \mathbb{R}^{C \times H \times W}$ as inputs and projects them into an orthogonal subspace using two learnable matrices $W_R$ and $W_S$. These projections satisfy:
\begin{equation}
W_R^T W_S = 0,
\end{equation}
\begin{equation}
\|W_R\|_F = \|W_S\|_F = 1.
\end{equation}
Where the constraints are imposed using orthogonal weight normalization \cite{06079,yang2024}. This process centralizes the shared weight matrix, computes its covariance, applies singular value decomposition, and normalizes columns by the inverse square roots of the eigenvalues. The resulting submatrices $W_R$ and $W_S$ are then used to obtain the orthogonalized features: $F_R' = W_R F_R$ and $F_S' = W_S F_S$, ensuring that modality-specific information remains decorrelated in the transformed feature space.

To adaptively fuse these independent features, OCNF incorporates reliability scores $R_R$ and $R_S$ produced by DMQA. A lightweight MLP first maps these scores to channel-wise reliability vectors $\tilde{R}_R, \tilde{R}_S \in \mathbb{R}^{1 \times C}$. Channel-wise softmax is then applied to derive modality-specific fusion weights:
\begin{equation}
\gamma_R = \frac{\exp(\tilde{R}_R^{(c)})}{\exp(\tilde{R}_R^{(c)}) + \exp(\tilde{R}_S^{(c)})},
\end{equation}
\begin{equation}
\gamma_S = \frac{\exp(\tilde{R}_S^{(c)})}{\exp(\tilde{R}_R^{(c)}) + \exp(\tilde{R}_S^{(c)})}.
\end{equation}
Where $c = 1, \dots, C$ denotes the channel dimension. The final fused feature is computed as:
\begin{equation}
F_{\text{fused}} = \gamma_R \odot F'_R + \gamma_S \odot F'_S.
\end{equation}
The symbol $\odot$ indicates element-wise multiplication applied across the entire feature map. This design enables fine-grained, channel-level comparison between modalities, suppressing unreliable information while enhancing high-quality features. By integrating orthogonal transformation with dynamic quality-aware weighting, this formulation maintains feature independence across modalities while preventing low-quality features and noise from degrading the fused representation. This mechanism is particularly effective in handling partial or missing modality data scenarios, where cross-modal noise propagation can significantly degrade detection performance.

\subsection{Loss Function}
After the DAF module, the enhanced features $\mathbf{\overset{\frown}{F}_r}$ and $\mathbf{\overset{\frown}{F}_s}$ are aggregated to produce the fused feature map $\mathbf{P}_i$. The fused features from the last three stages are subsequently fed into the detection head for supervised training. The overall training objective is defined as:
\begin{equation}
\mathcal{L}_{\text{total}} = \lambda_1 \mathcal{L}_{\text{box}} + \lambda_2\mathcal{L}_{\text{cls}} + \lambda_3 \mathcal{L}_{\text{obj}}.
\end{equation}
Where $\mathcal{L}_{\text{box}}$ denotes the bounding box regression loss, $\mathcal{L}_{\text{cls}}$ is the classification loss, and $\mathcal{L}_{\text{obj}}$ represents the objectiveness for optimizing localization confidence. The coefficients $\lambda_1$, $\lambda_2$, and $\lambda_3$ control the relative importance of each loss component during optimization.

\section{EXPERIMENTAL RESULTS AND ANALYSIS}
In this section, we begin by describing the datasets and evaluation metrics adopted in our experiments. Next, we detail the experimental setup and implementation. We then compare our approach with existing popular methods. Finally, ablation studies are conducted to verify the contribution and effectiveness of the major components.

\subsection{Datasets and Evaluation Metrics}

\noindent{\textbf{SpaceNet6-OTD-Fog Dataset}}. The SpaceNet6-OTD-Fog dataset is an enhanced benchmark derived from SpaceNet 6 for oil tank detection in high-resolution SAR imagery. It consists of Gaofen-3 SAR images paired with manually co-registered optical counterparts collected from Google Earth. To better reflect real-world operational conditions such as adverse weather and optical degradation, we introduce synthetic fog to the optical modality, forming a challenging SAR-optical pair setting. The final dataset contains 620 training pairs and 200 testing pairs, each annotated with oil tank instances.

\begin{table*}[t]
\centering
\caption{Performance comparison on SpaceNet6-OTD-Fog dataset. Comparison on random missing protocol. The values reported in each cell denote mAP50/mAP. Bold is the best.}
\resizebox{\linewidth}{1.2\height}{
\begin{tabular}{c|c|c|c|c|c|c|c|c}
\hline
Datasets & MR & CFT & CALNet & Superyolo & ICAFusion & TFDet & MMIDet & QDFNet(Ours) \\
\hline
\hline
\multirow{6}{*}{\makecell{SpaceNet6-OTD-Fog \\ (Zero)}} 
& 0.0 & 80.2 / \underline{42.7} & 79.6 / 41.8 & \underline{81.8} / 42.5 & 80.3 / 39.8 & 80.2 / 41.0 & \textbf{82.6} / 42.1 & \textbf{82.6} / \textbf{44.9} \\
& 0.1 & 78.6 / \underline{40.8} & 78.4 / 40.3 & 73.3 / 39.0 & 78.8 / 37.9 & 75.3 / 37.7 & \underline{78.9} / 40.2 & \textbf{80.8} / \textbf{42.9} \\
& 0.2 & 76.6 / 38.7 & \underline{77.3} / \underline{38.8} & 71.0 / 37.4 & \underline{77.3} / 35.6 & 72.2 / 36.9 & \textbf{77.8} / 38.4 & \textbf{77.8} / \textbf{39.4} \\
& 0.3 & 73.8 / \underline{37.5} & 74.8 / 36.6 & 70.5 / 37.3 & 73.0 / 34.0 & 68.0 / 33.9 & \underline{75.2} / 36.3 & \textbf{76.6} / \textbf{38.3} \\
& 0.4 & 72.8 / \underline{36.3} & 72.6 / 34.6 & 69.3 / 36.2 & 71.5 / 33.1 & 66.2 / 32.0 & \underline{74.1} / 36.1 & \textbf{75.0} / \textbf{38.2} \\
& 0.5 & 71.3 / 35.3 & 69.6 / 33.5 & 66.3 / 35.5 & 70.9 / 32.6 & 63.7 / 31.6 & \underline{72.4} / \underline{35.6} & \textbf{73.8} / \textbf{37.1} \\
\hline

\noalign{\vskip 0.6mm}
\hline

\multirow{5}{*}{\makecell{SpaceNet6-OTD-Fog \\ (INN)}}
& 0.1 & \underline{79.5} / \underline{40.9} & 79.3 / 39.3 & 73.9 / 37.7 & 78.4 / 37.3 & 76.2 / 38.1 & 79.2 / 40.2 & \textbf{81.9} / \textbf{42.9} \\
& 0.2 & 77.1 / \underline{39.7} & \underline{78.2} / 38.5 & 73.0 / 37.6 & 77.3 / 35.6 & 73.6 / 36.1 & 77.2 / 38.1 & \textbf{79.8} / \textbf{41.5} \\
& 0.3 & 75.2 / \underline{38.1} & 74.8 / 36.6 & 72.7 / 37.0 & 74.2 / 34.4 & 71.3 / 34.1 & \underline{75.5} / 37.2 & \textbf{77.0} / \textbf{39.7} \\
& 0.4 & 73.0 / \underline{36.8} & 73.4 / 34.9 & 69.5 / 36.6 & 71.8 / 33.2 & 68.8 / 32.9 & \underline{74.1} / 35.3 & \textbf{75.1} / \textbf{38.2} \\
& 0.5 & 71.6 / \underline{35.4} & 70.4 / 33.0 & 68.5 / 32.4 & 70.8 / 31.9 & 65.3 / 31.1 & \underline{72.1} / 35.2 & \textbf{74.2} / \textbf{37.6} \\
\hline
\end{tabular}
}
\label{tab:1}
\end{table*}

\begin{table*}[t]
\centering
\caption{Performance comparison on  OGSOD-2.0 dataset. Comparison on random missing protocol. The values reported in each cell denote mAP50/mAP. Bold is the best.}
\resizebox{\linewidth}{1.2\height}{
\begin{tabular}{c|c|c|c|c|c|c|c|c}
\hline
Datasets & MR & CFT & CALNet & Superyolo & ICAFusion & TFDet & MMIDet & QDFNet(Ours) \\
\hline
\hline
\multirow{6}{*}{\makecell{ OGSOD-2.0 \\ (Zero)}} 
& 0.0 &  \underline{79.0} / \underline{41.5} & 76.8 / 38.9 & 76.4 / 39.1 & \textbf{79.6} /\textbf{ 41.9} & 75.2 / 40.1 & 77.7 / 39.3 & \textbf{79.6} / \textbf{41.9} \\
& 0.1 & 74.1 / 37.7 & 72.1 / 36.2 & 72.1 / 35.5 & \underline{76.0} / \underline{38.5} & 70.1 / 39.3 & 74.0 / 38.1 & \textbf{76.2} /\textbf{ 38.8} \\
& 0.2 & 68.9 / 35.0 & 66.9 / 33.2 & 68.7 / 32.2 & \underline{71.7} / \underline{37.0} & 69.1 / \underline{37.0} & \underline{71.7} / 36.4 & \textbf{72.3} / \textbf{37.1} \\
& 0.3 & 65.5 / 33.1 & 61.6 / 30.5 & 63.0 / 31.6 & 66.9 / 35.0 & 64.4 / \textbf{36.3} & \underline{67.6} / 32.4 & \textbf{69.7} / \underline{35.1} \\
& 0.4 & 59.3 / 29.8 & 58.4 / 29.8 & 58.4 / 32.1 & \underline{61.5} / \underline{33.2} & 59.7 / 33.0 & 58.7 / 28.6 & \textbf{65.0} /\textbf{ 33.5} \\
& 0.5 & 52.6 / 24.6 & 55.7 / 27.0 & 52.1 / 28.3 & \underline{58.1} / \underline{29.7} & 55.3 / 29.0 & 52.8 / 25.3 & \textbf{61.8} / \textbf{30.1} \\
\hline

\noalign{\vskip 0.6mm}
\hline

\multirow{5}{*}{\makecell{OGSOD-2.0 \\ (INN)}}
& 0.1 & 73.3 / 37.0 & 72.2 / 36.3 & 72.9 / 36.0 & \underline{76.5} / \underline{38.3} & 70.1 / \textbf{39.1} & 75.6 / 38.1 & \textbf{76.9} / \underline{38.3} \\
& 0.2 & 67.7 / 33.5 & 69.9 / 38.5 & 68.6 / 32.2 & \underline{73.4} / \textbf{37.6} & 69.0 / 36.8 & 71.9 / 35.9 & \textbf{75.1}/ \underline{37.2} \\
& 0.3 & 64.3 / 31.4 & 65.8 / 34.6 & 63.0 / 31.6 & \underline{69.7} / \underline{35.7} & 63.8 / 34.2 & 68.5 / 35.0 & \textbf{70.1} / \textbf{35.9} \\
& 0.4 & 59.1 / 29.0 & 60.3 / \underline{32.9} & 58.4 / 30.1 & \underline{65.9} / 32.5 & 59.7 / 31.9 & 63.6 / 31.7 & \textbf{68.3} / \textbf{33.2} \\
& 0.5 & 55.1 / 26.4 & 57.2 / 26.0 & 52.1 / 28.3 & \underline{62.0} / \underline{31.2} & 57.1 / 30.3 & 58.4 / 29.4 & \textbf{64.8} / \textbf{32.2} \\
\hline
\end{tabular}
}
\label{tab:2}
\end{table*}

\noindent{\textbf{OGSOD-2.0 Dataset}}.
The OGSOD-2.0 dataset contains multimodal remote sensing imagery, pairing optical data from Sentinel-2 with SAR observations from Sentinel-1 \cite{ruan2025ogsod}. It provides a large-scale benchmark with 21,115 image pairs, including 18,768 for training and 2,347 for testing. The dataset covers four representative object categories, including bridges, harbors, oil tanks, and playgrounds, which reflect diverse structural patterns and imaging conditions.

\noindent{\textbf{Evaluation Metrics}}.
We adopt mean Average Precision (mAP) as the primary evaluation metric for multimodal object detection. For fair comparison, we report both mAP50 and mAP. mAP50 measures AP at a single IoU threshold of 0.5, reflecting detection accuracy and coarse localization. In contrast, mAP averages AP across IoU thresholds from 0.50 to 0.95 at 0.05 intervals, providing a more comprehensive assessment of localization precision and overall detection robustness.

\subsection{Implementation Details}
We evaluate all methods on multimodal object detection datasets under a random modality dropout protocol that simulates realistic modality-missing scenarios where sensor data may be unavailable due to environmental or technical conditions. Each optical-SAR image pair either retains both modalities or has one modality dropped under random modality dropout. The probability of missing the optical modality is set equal to that of missing the SAR modality for each predefined Missing Rate (MR), and at least one modality is always retained, ensuring $a_{i} \geq 1$ for every sample. The missing rate (MR) quantifies the proportion of missing modalities, defined as $MR = 1 - \frac{\sum_{i = 1}^{L} a_{i}}{L\times M}$, where $a_{i}$ is the number of available modalities for the $i$th sample, $L$ is the total number of samples, and $M$ is the number of modalities. For two modalities, the theoretical upper bound of MR is $\frac{M-1}{M}$, and we adopt values from $0.0$ to $0.5$ at increments of $0.1$, where $0.5$ approximates this bound.
We use the same MR during training, validation, and testing, following previous work. All experiments are implemented in PyTorch and conducted on an RTX A40 GPU with 48 GB memory. We use a batch size of 16 and train the network for 120 epochs until convergence.

\subsection{Comparisons with State-of-the-Art Methods}
Existing multimodal object detectors typically rely on complete multimodal inputs, and their performance degrades noticeably when facing missing modality conditions. In this work, we present a systematic investigation of the missing modality problem in multimodal object detection with optical and SAR data. Our goal is to provide a thorough evaluation of existing algorithms under different missing modality scenarios and to examine the effectiveness of various compensation strategies. The evaluated methods include several representative multimodal detectors, such as CFT~\cite{22}, CALNet~\cite{he2023multispectral}, SuperYOLO~\cite{16}, ICAFusion~\cite{25}, TFDet~\cite{tfdet} and MMIDet~\cite{26}. As shown in Table \ref{tab:1} and Table \ref{tab:2}, comparing different algorithms under the same compensation strategy, together with evaluating each algorithm across multiple strategies, enables a comprehensive assessment of the robustness of current multimodal object detectors under missing modality conditions.

\subsubsection{Evaluation on Modality-complete Datasets}

We conducted comprehensive experiments to evaluate our proposed method against several representative multimodal object detection algorithms on two modality-complete datasets, SpaceNet6-OTD-Fog and OGSOD-2.0. As summarized in Table~\ref{tab:1} and Table~\ref{tab:2}, QDFNet achieves consistently superior or comparable results under modality-complete conditions, demonstrating its effectiveness in exploiting multimodal information.
On the SpaceNet6-OTD-Fog dataset, which contains complete optical and SAR modalities, QDFNet demonstrates clear advantages over representative multimodal detectors. Compared with the baseline CFT, QDFNet improves mAP50 by 2.4\% and mAP by 2.2\%, confirming its stronger ability to integrate complementary features from different modalities. In comparison with MMIDet, QDFNet obtains similar performance in terms of mAP50 while achieving a 2.8\% gain in mAP, which indicates enhanced detection capability for objects with complex structures. Furthermore, QDFNet consistently outperforms ICAFusion and SuperYOLO. Specifically, QDFNet yields improvements of 1.3\% and 4.8\% over ICAFusion, and 0.8\% and 2.4\% over SuperYOLO in mAP50 and mAP, respectively. These results verify the generality of the proposed feature fusion strategy and its robustness across different backbone architectures and fusion mechanisms.
On the OGSOD-2.0 dataset, QDFNet achieves the best overall results among all evaluated methods. It obtains an mAP50 of 79.6\% and an mAP of 41.9\%, which are comparable to the top performing ICAFusion and confirm the effectiveness of the proposed fusion mechanism under full modality conditions. Compared with the baseline CFT, QDFNet demonstrates consistent improvements of 0.6\% in mAP50 and 0.4\% in mAP. In addition, QDFNet surpasses other competitive approaches, including CALNet and SuperYOLO, with gains of 2.8\% and 3.0\% in mAP50, and 3.2\% and 2.8\% in mAP, respectively. These results show that QDFNet provides more reliable feature representation and stronger object discrimination capability, leading to steady improvements across different network designs and multimodal configurations.

\subsubsection{Evaluation on Modality-missing Datasets}

We evaluate multimodal object detectors under missing modality scenarios using SpaceNet6-OTD-Fog and OGSOD-2.0. Two complementary strategies are considered: (i) Zero filling, where unavailable modalities are masked, representing a direct operational baseline. (ii) INN reconstruction, where missing modalities are first generated via invertible networks before detection. This design enables a consistent assessment of both resilience to missing inputs and the ability to exploit reconstructed modality cues.

\begin{table*}[h]
\centering
\caption{Ablation study of module contributions on the SpaceNet6-OTD-Fog dataset under the Zero-filling setting. Performance is evaluated using mAP50 and mAP.}
\label{tab:3}

\small 
\setlength{\tabcolsep}{12pt} 
\begin{tabular}{c|c|c|c|c|c}
\hline
Datasets & MR & Baseline & w/ DMQA & w/ OCNF & Ours \\
\hline
\hline
\multirow{6}{*}{\makecell{SpaceNet6-OTD-Fog \\ (Zero)}} 
& 0.0 & 80.2 / 42.7 & 81.3 / 43.5 & 80.9 / 43.3 & \textbf{82.6} / \textbf{44.9} \\
& 0.1 & 78.6 / 40.8 & 79.4 / 41.3 & 79.2 / 41.3 & \textbf{80.8} / \textbf{42.9} \\
& 0.2 & 76.6 / 39.4 & 77.5 / 40.5 & 77.1 / 40.4 & \textbf{77.8} / \textbf{39.4} \\
& 0.3 & 73.8 / 37.5 & 74.7 / 38.6 & 74.5 / 38.1 & \textbf{76.6} / \textbf{38.3} \\
& 0.4 & 72.8 / 36.3 & 73.6 / 37.1 & 73.5 / 36.9 & \textbf{75.0} / \textbf{38.2} \\
& 0.5 & 71.3 / 35.3 & 72.2 / 36.3 & 71.9 / 36.1 & \textbf{73.8} / \textbf{37.1} \\
\hline
\end{tabular}
\end{table*}

\begin{table*}[h]
\centering
\caption{Ablation study of module contributions on the SpaceNet6-OTD-Fog dataset under INN reconstruction setting. Performance is evaluated using mAP50 and mAP.}
\label{tab:4}

\small 
\setlength{\tabcolsep}{12pt} 
\begin{tabular}{c|c|c|c|c|c}
\hline
Datasets & MR & Baseline & w/ DMQA & w/ OCNF & Ours \\
\hline
\hline
\multirow{6}{*}{\makecell{SpaceNet6-OTD-Fog \\ (INN)}} 
& 0.0 & 80.2 / 42.7 & 81.3 / 43.5 & 80.9 / 43.3 & \textbf{82.6} / \textbf{44.9} \\
& 0.1 & 79.1 / 41.5 & 79.9 / 41.6 & 79.7 / 41.5 & \textbf{81.9} / \textbf{42.9} \\
& 0.2 & 77.1 / 39.7 & 78.0 / 40.6 & 77.7 / 40.5 & \textbf{79.8} / \textbf{41.5} \\
& 0.3 & 75.2 / 38.1 & 76.4 / 39.1 & 76.2 / 38.8 & \textbf{77.0} / \textbf{39.7} \\
& 0.4 & 73.0 / 36.8 & 74.1 / 37.5 & 73.8 / 37.3 & \textbf{75.1} / \textbf{38.2} \\
& 0.5 & 71.6 / 35.4 & 72.6 / 36.3 & 72.0 / 36.0 & \textbf{74.2} / \textbf{37.6} \\
\hline
\end{tabular}
\end{table*}

\begin{figure*}[!ht]
    \centering
    \begin{tabular}{>{\centering\arraybackslash}m{0.15\textwidth}
                    >{\centering\arraybackslash}m{0.15\textwidth}
                    >{\centering\arraybackslash}m{0.15\textwidth}
                    >{\centering\arraybackslash}m{0.15\textwidth}
                    >{\centering\arraybackslash}m{0.15\textwidth}
                    >{\centering\arraybackslash}m{0.15\textwidth}}
        \includegraphics[width=\linewidth]{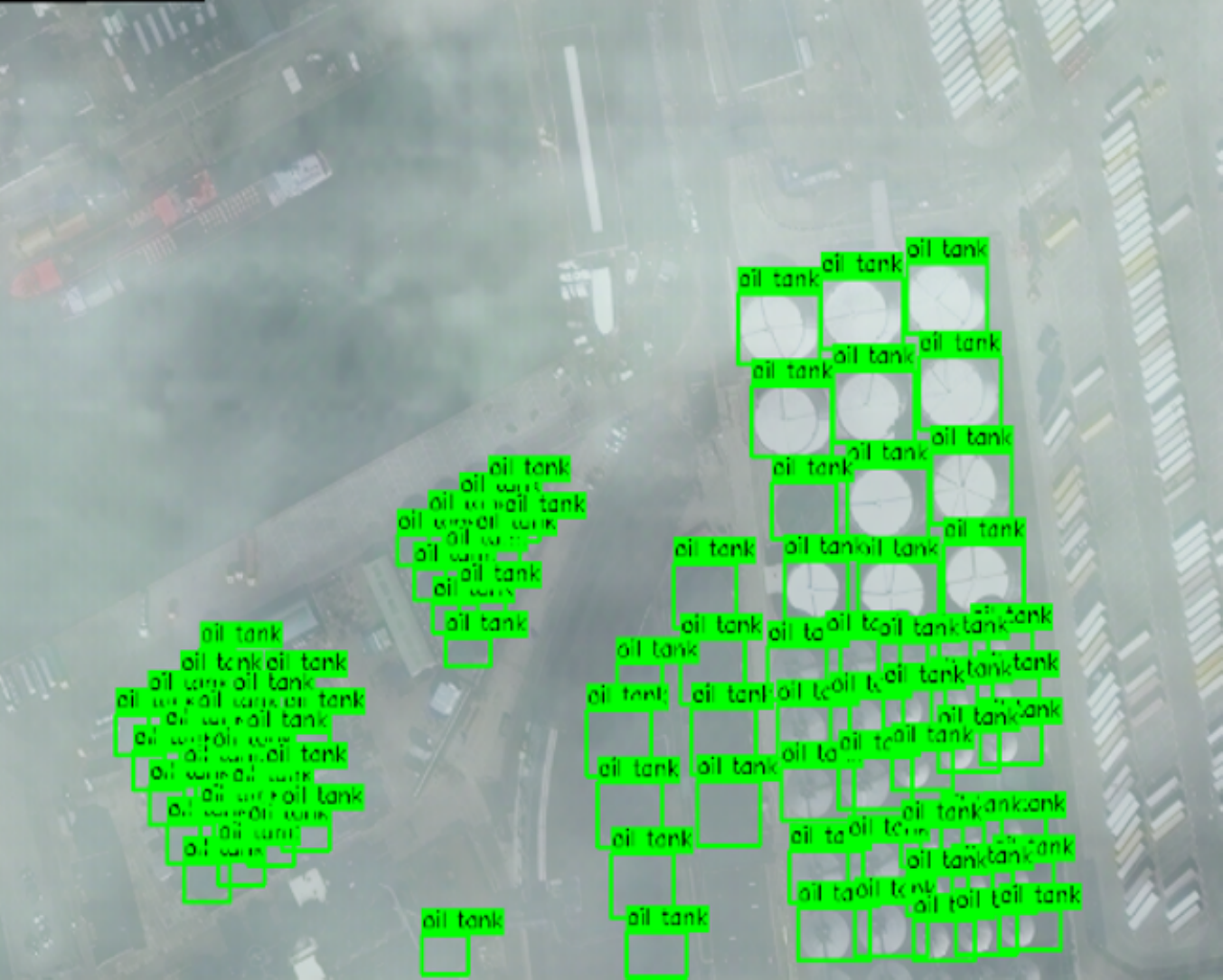} & 
        \includegraphics[width=\linewidth]{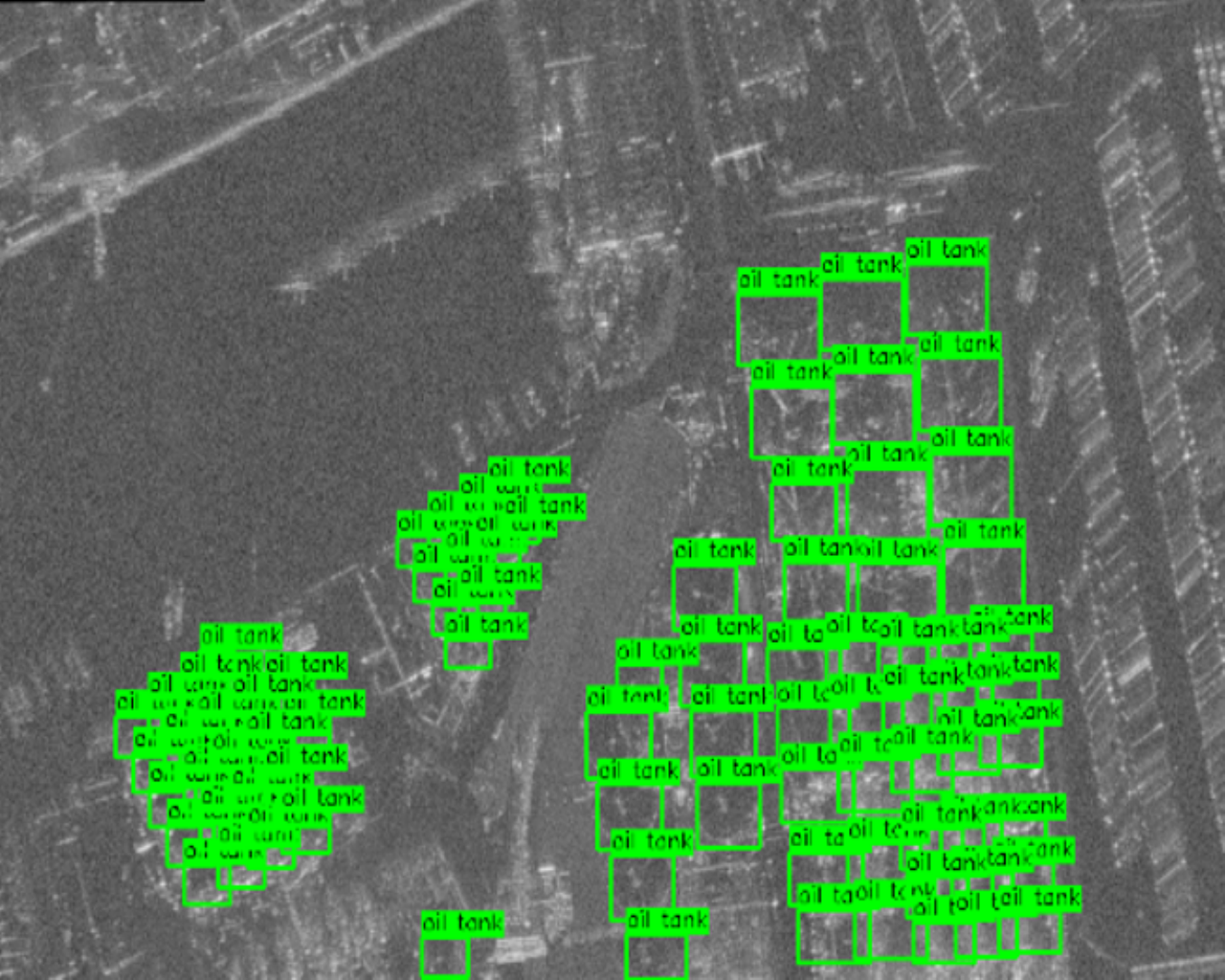} & 
        \includegraphics[width=\linewidth]{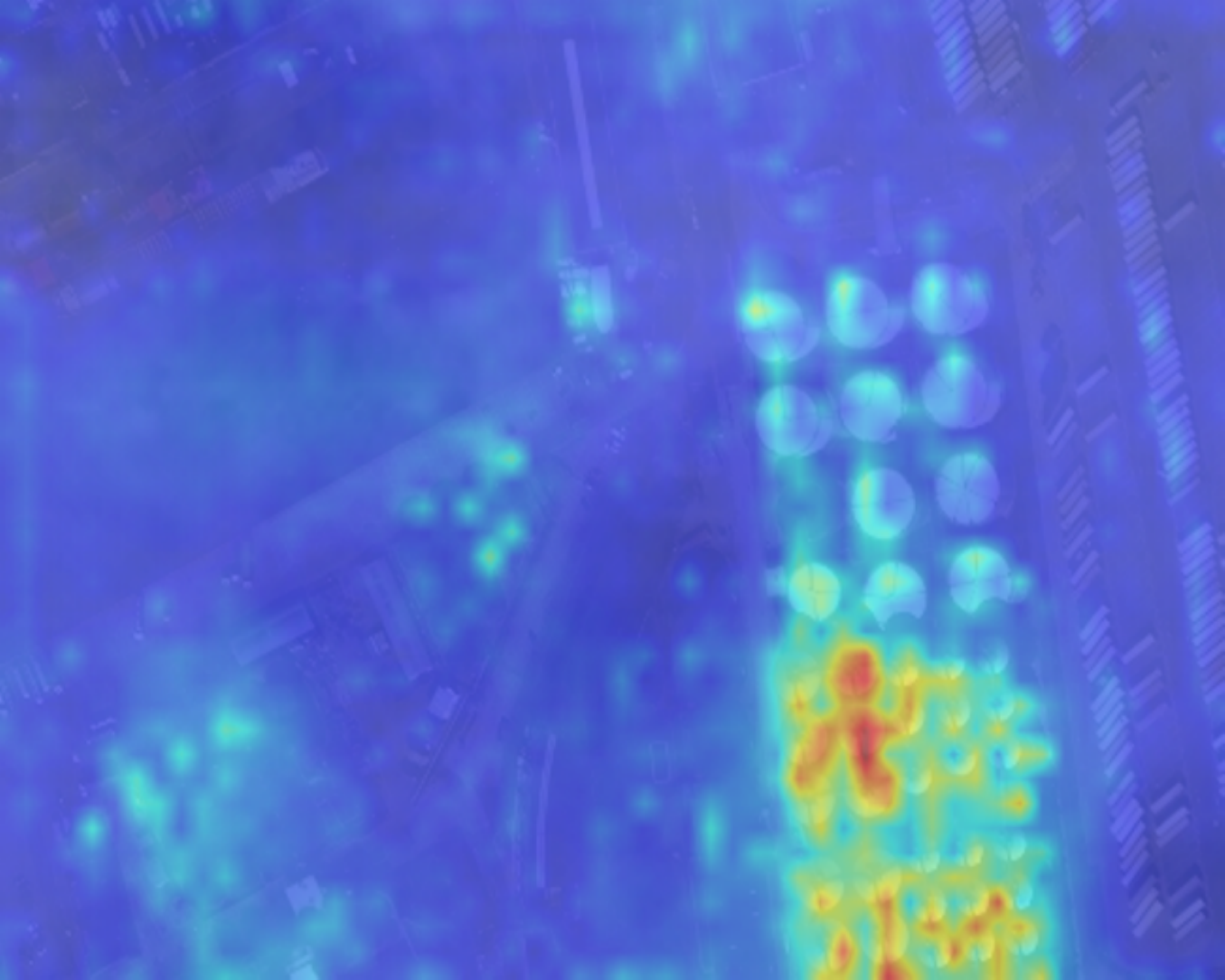} & 
        \includegraphics[width=\linewidth]{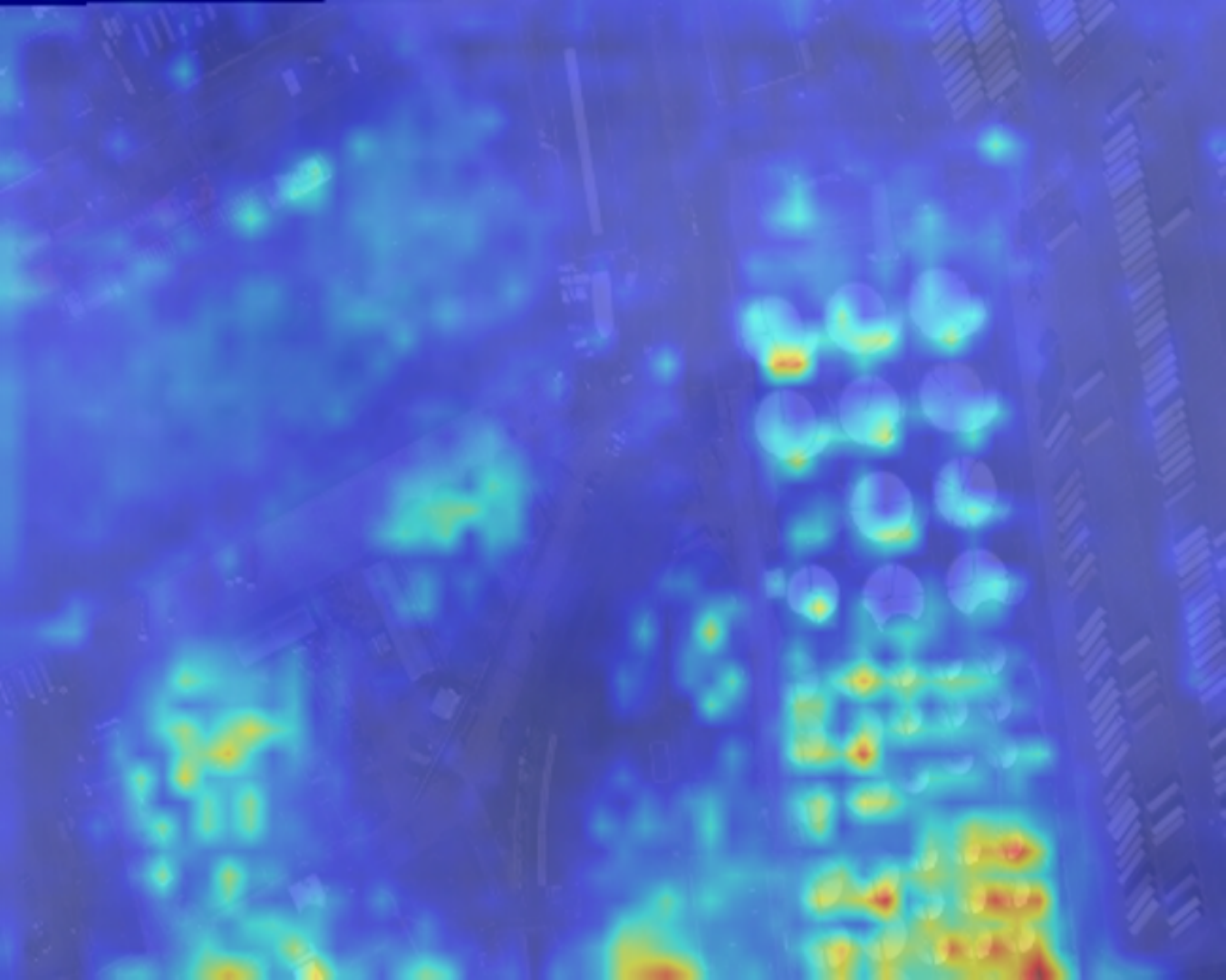} & 
        \includegraphics[width=\linewidth]{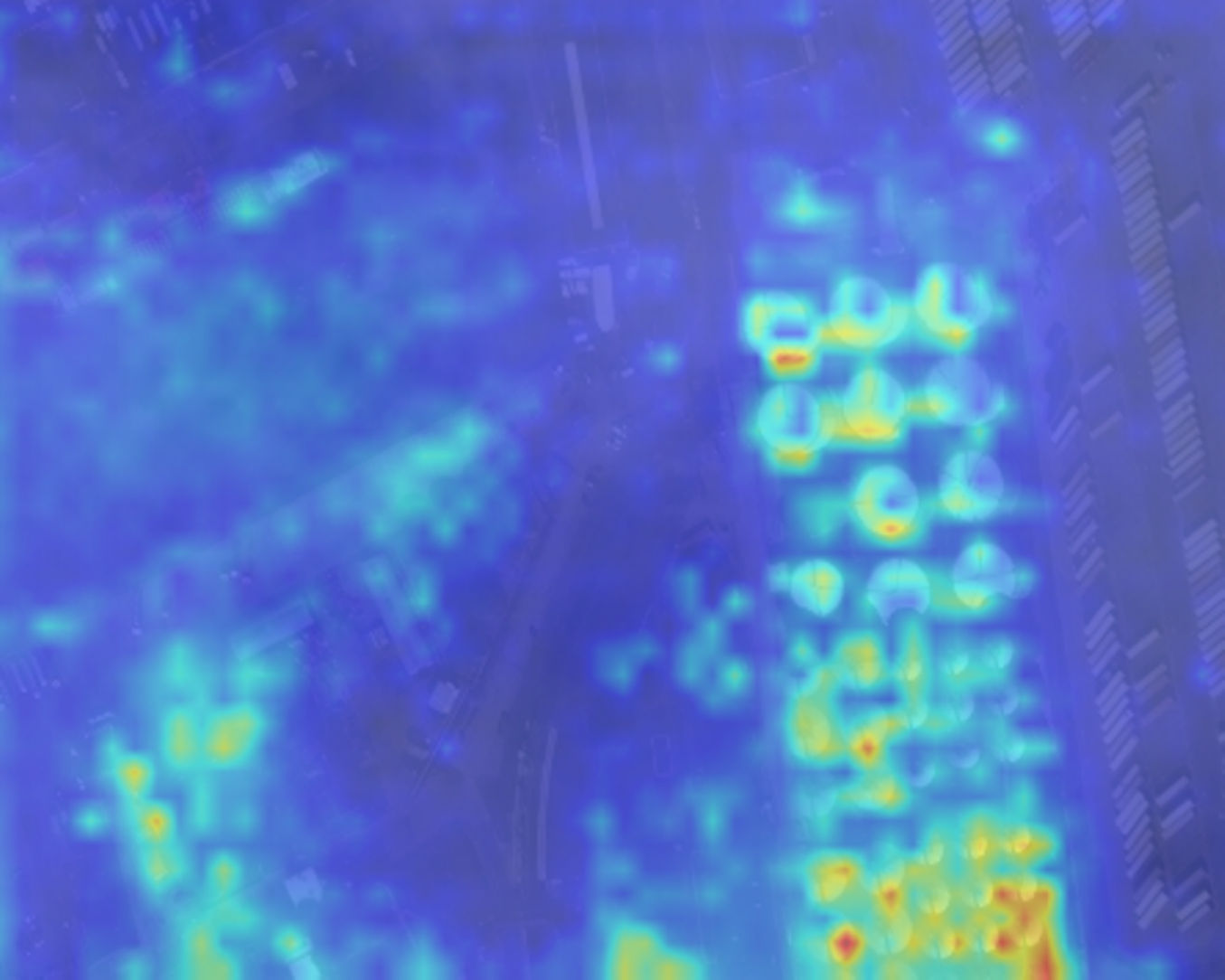} & 
        \includegraphics[width=\linewidth]{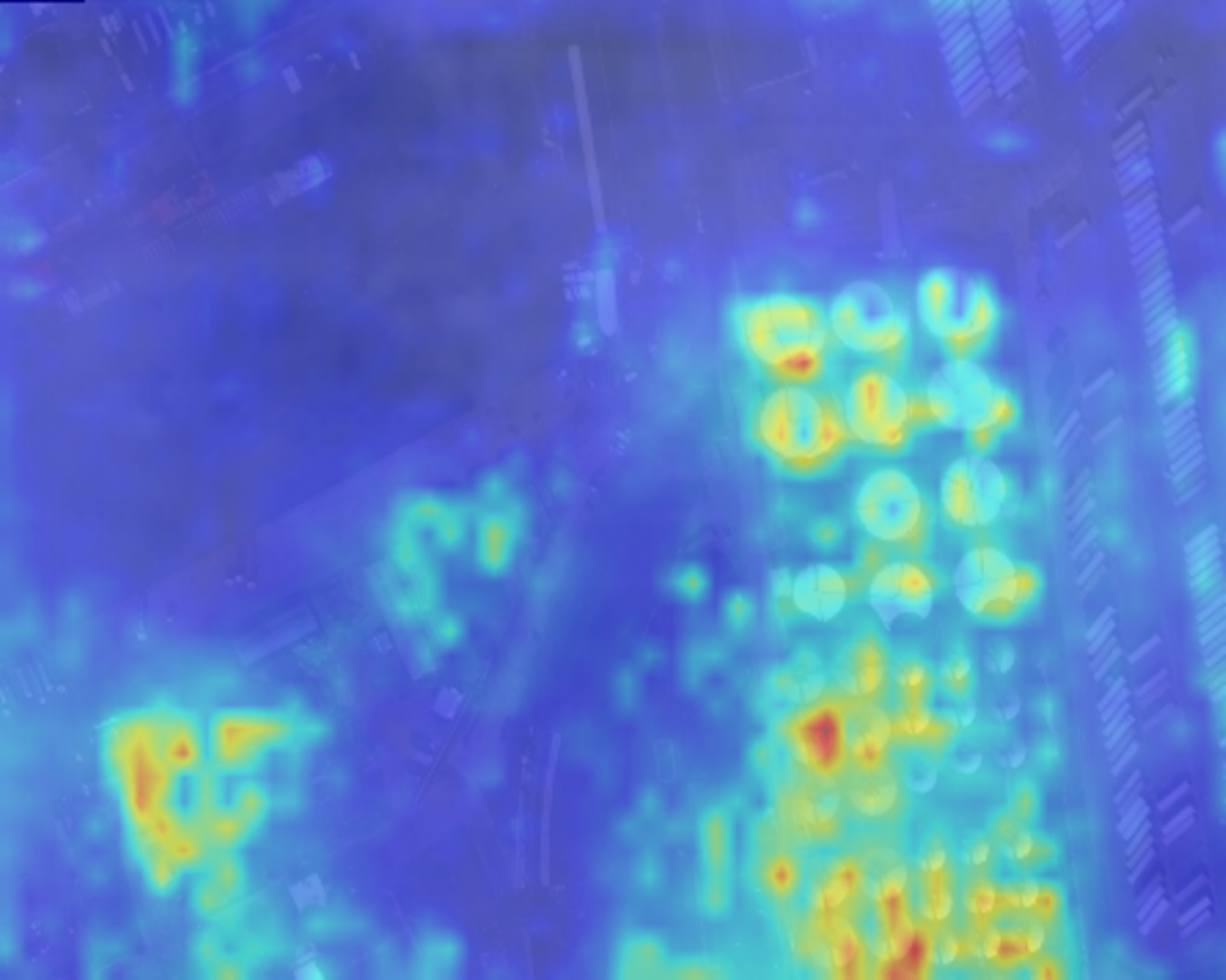} \\

        \includegraphics[width=\linewidth]{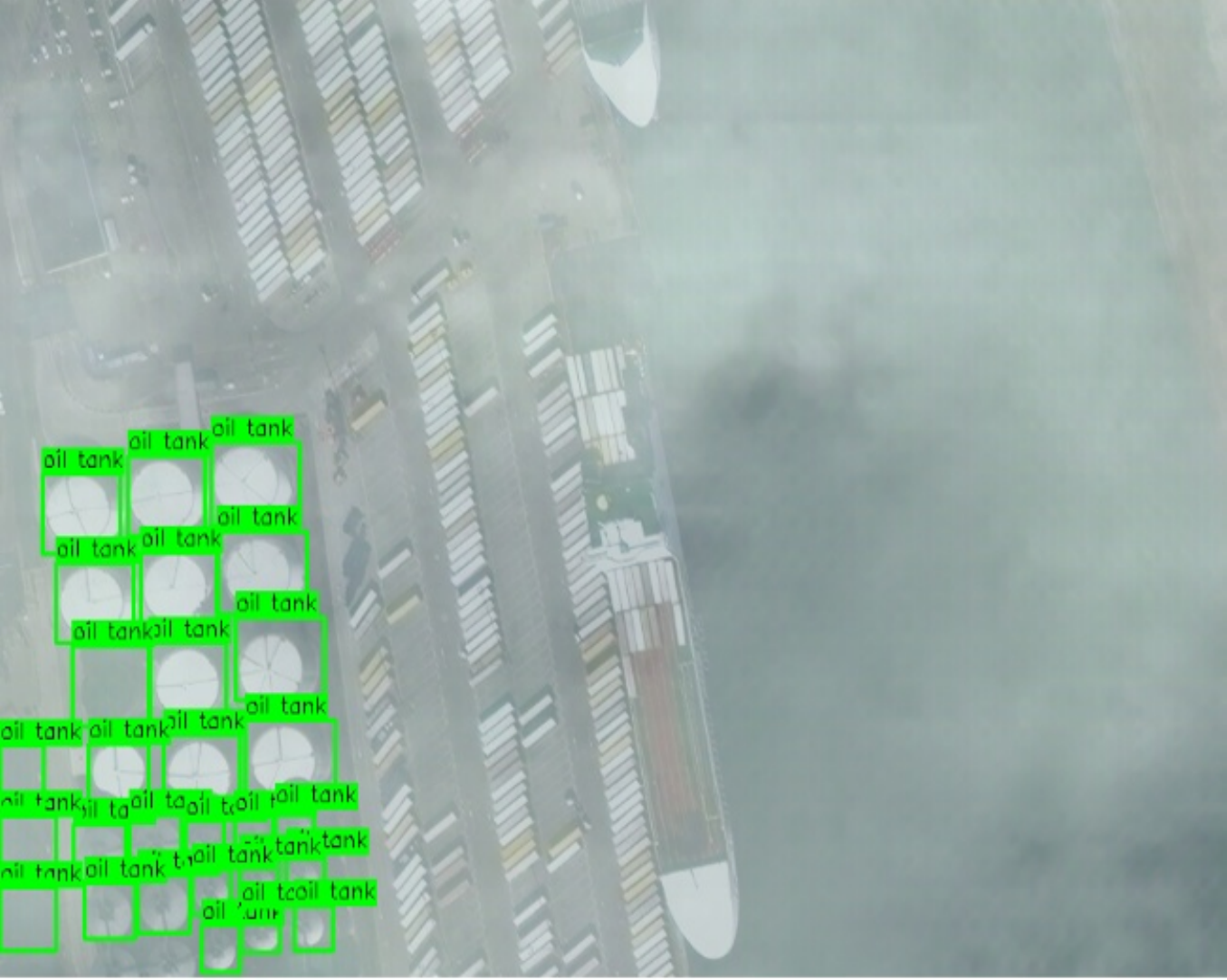} & 
        \includegraphics[width=\linewidth]{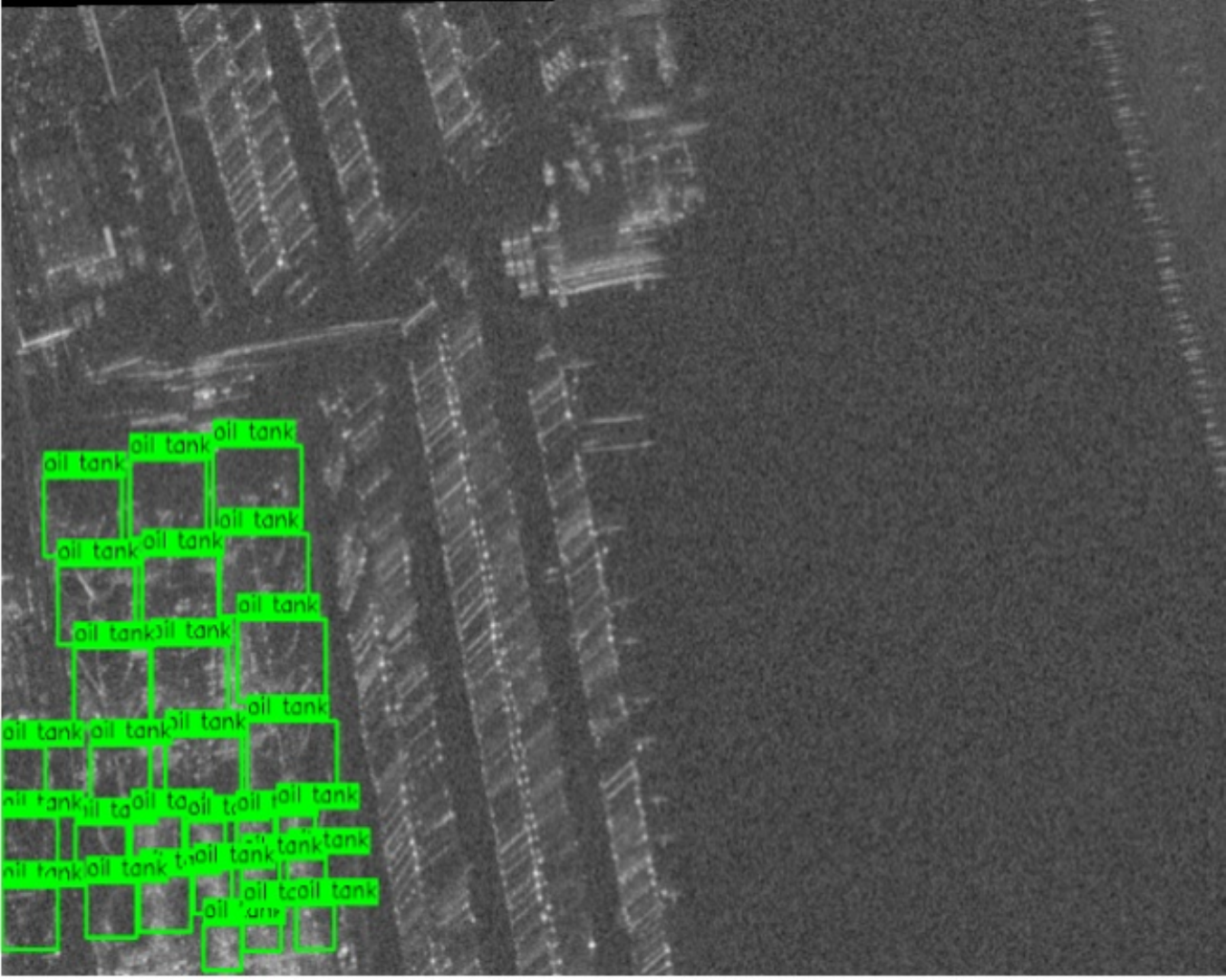} & 
        \includegraphics[width=\linewidth]{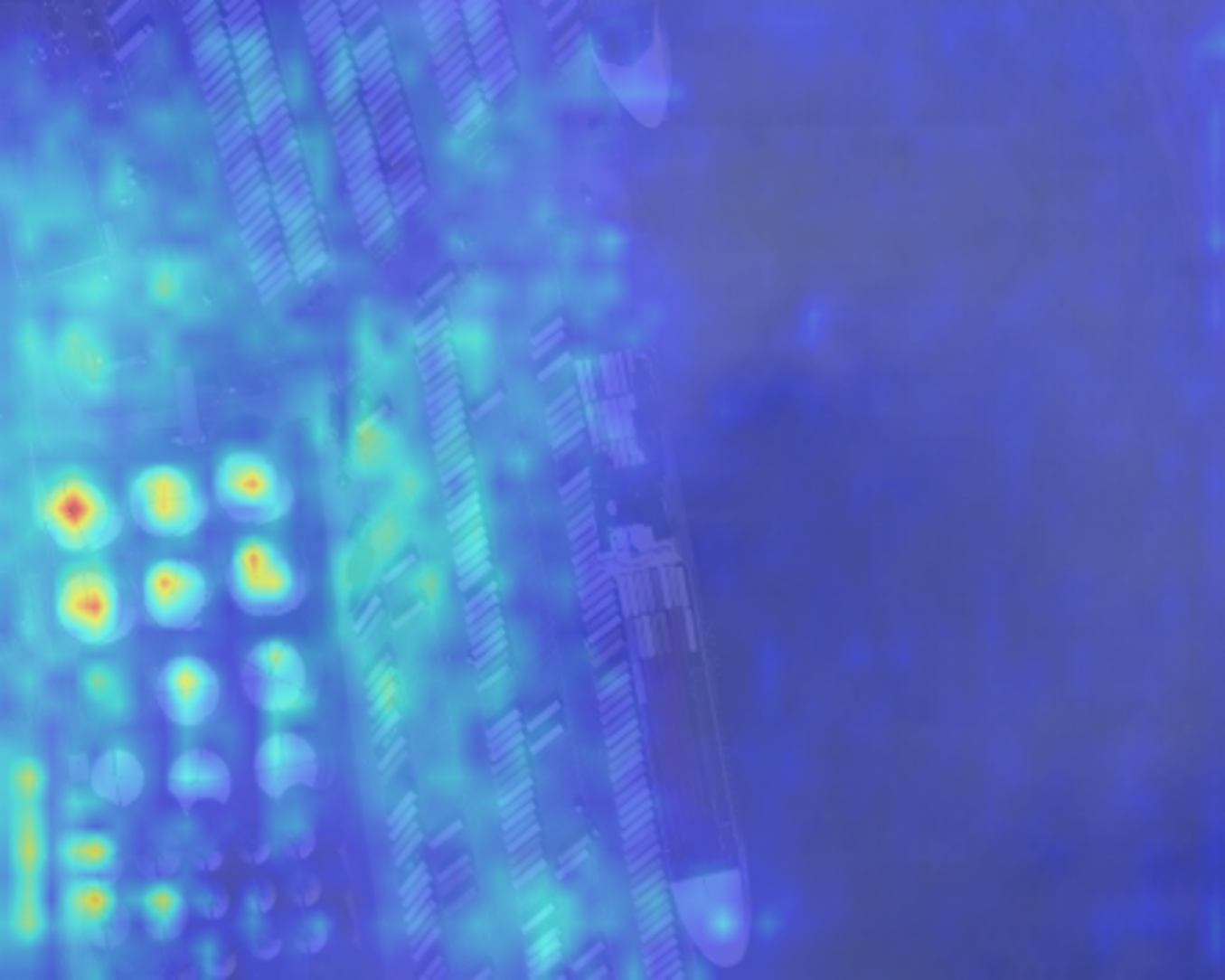} & 
        \includegraphics[width=\linewidth]{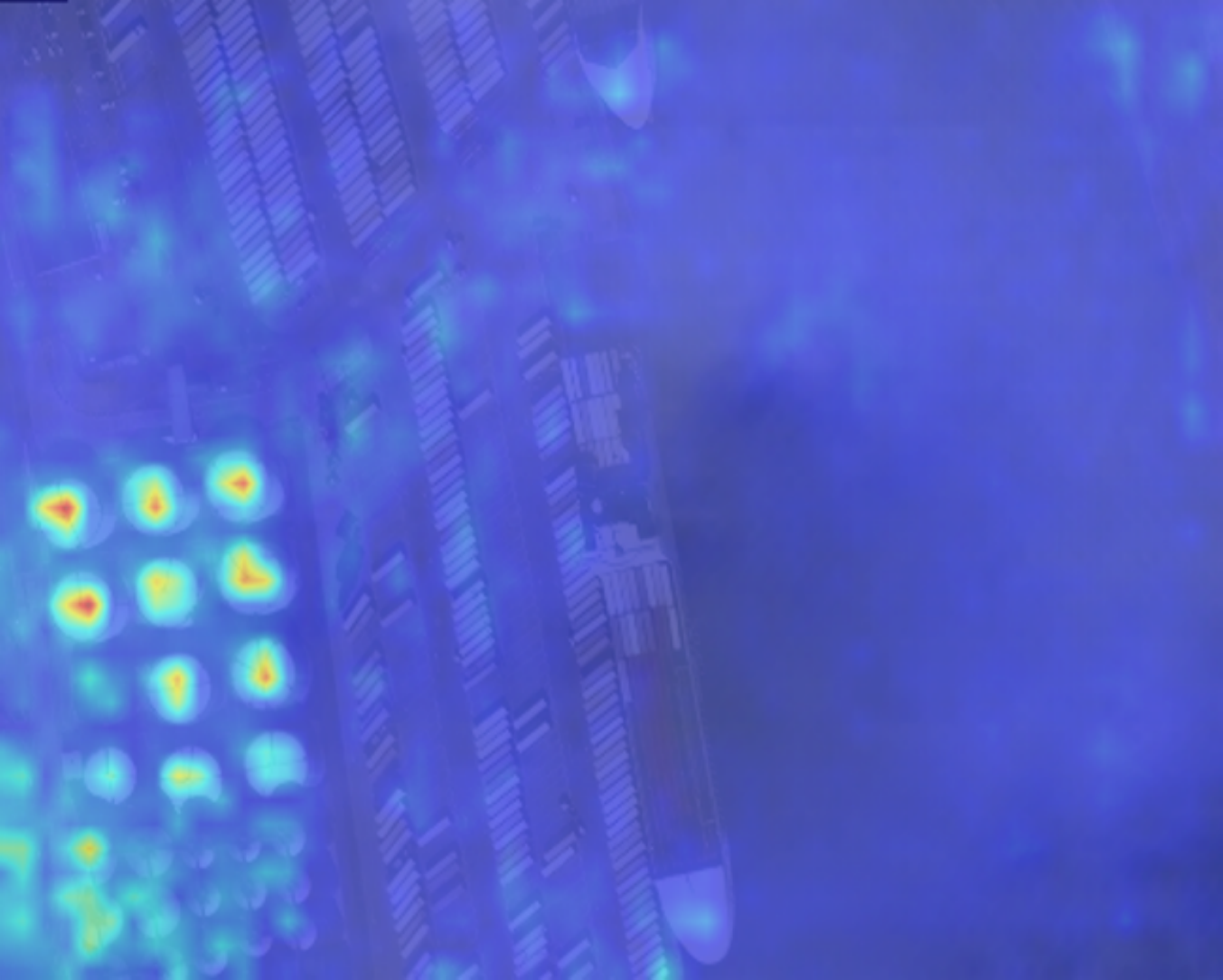} & 
        \includegraphics[width=\linewidth]{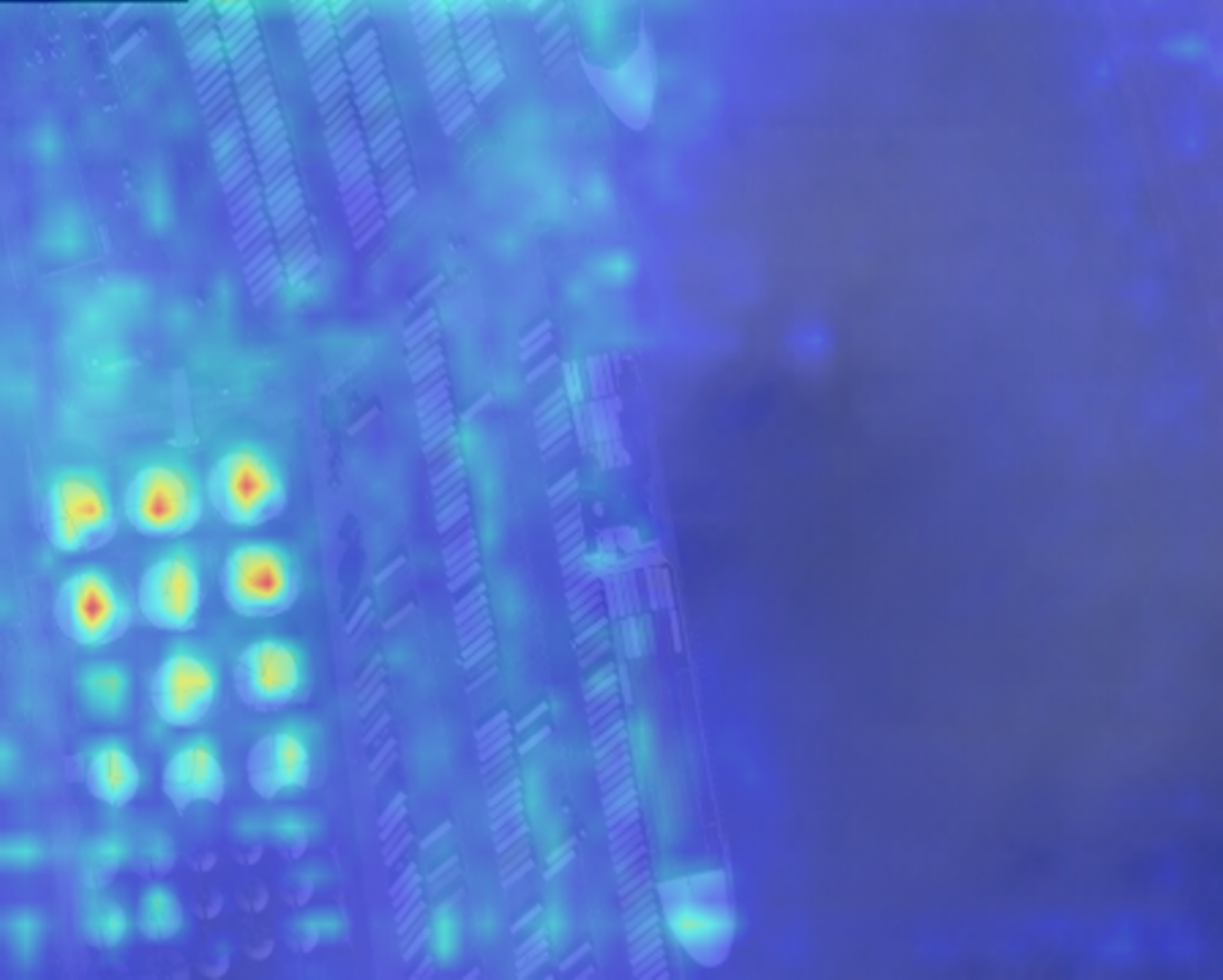} & 
        \includegraphics[width=\linewidth]{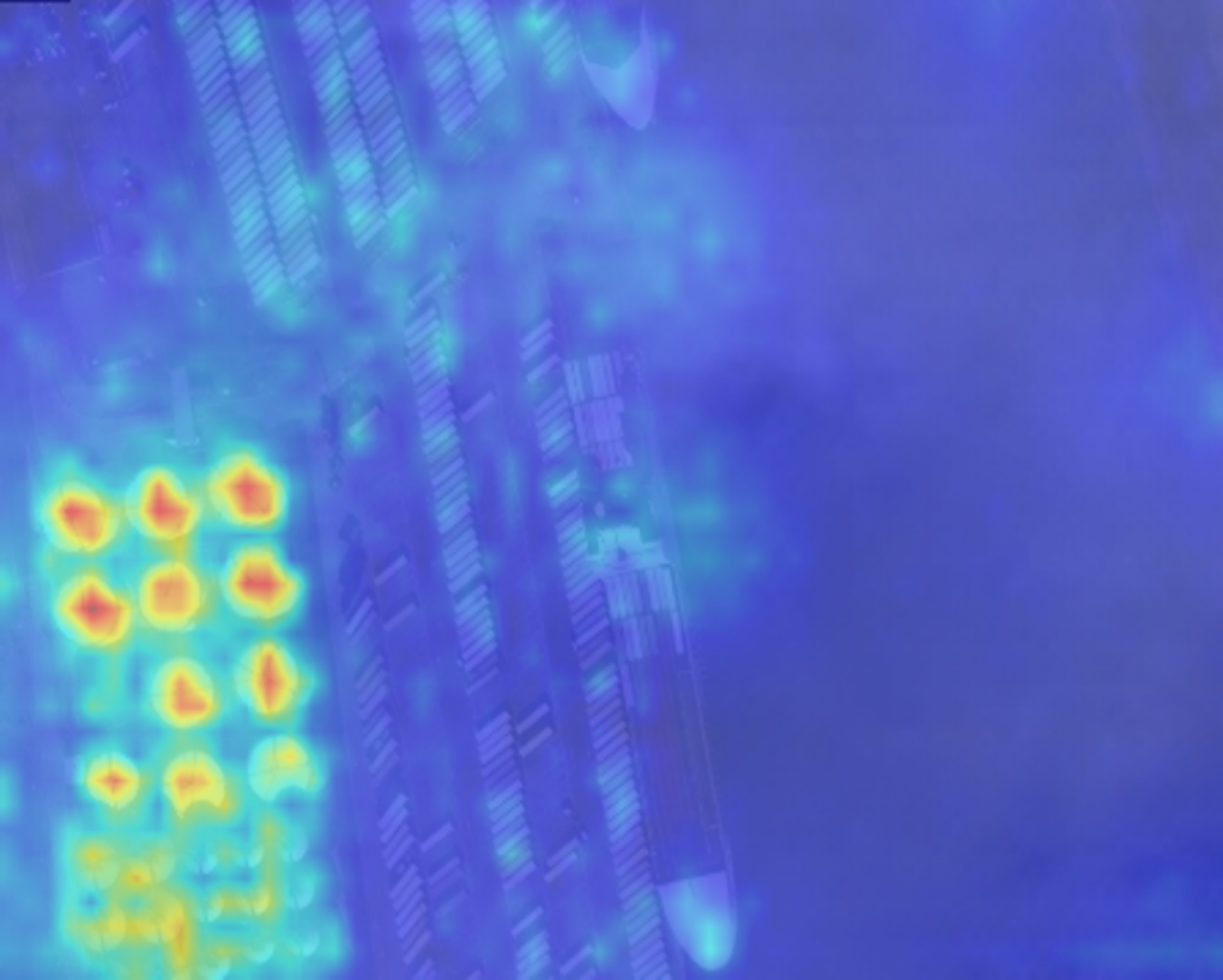} \\

        (a) & (b) & (c) & (d) & (e) & (f) \\
    \end{tabular}
    \caption{Visualization of the learning attention map using Grad-CAM under the scenario of MR = 0.3 and Zero-filling. (a) RGB GT Labels. (b) SAR GT Labels. (c) Baseline. (d) w/ DMQA. (e) w/ OCNF. (f) Ours. }
    \label{fig:5}
\end{figure*}

\begin{figure*}[!ht]
    \centering
    \begin{tabular}{>{\centering\arraybackslash}m{0.15\textwidth}
                    >{\centering\arraybackslash}m{0.15\textwidth}
                    >{\centering\arraybackslash}m{0.15\textwidth}
                    >{\centering\arraybackslash}m{0.15\textwidth}
                    >{\centering\arraybackslash}m{0.15\textwidth}
                    >{\centering\arraybackslash}m{0.15\textwidth}}
        \includegraphics[width=\linewidth]{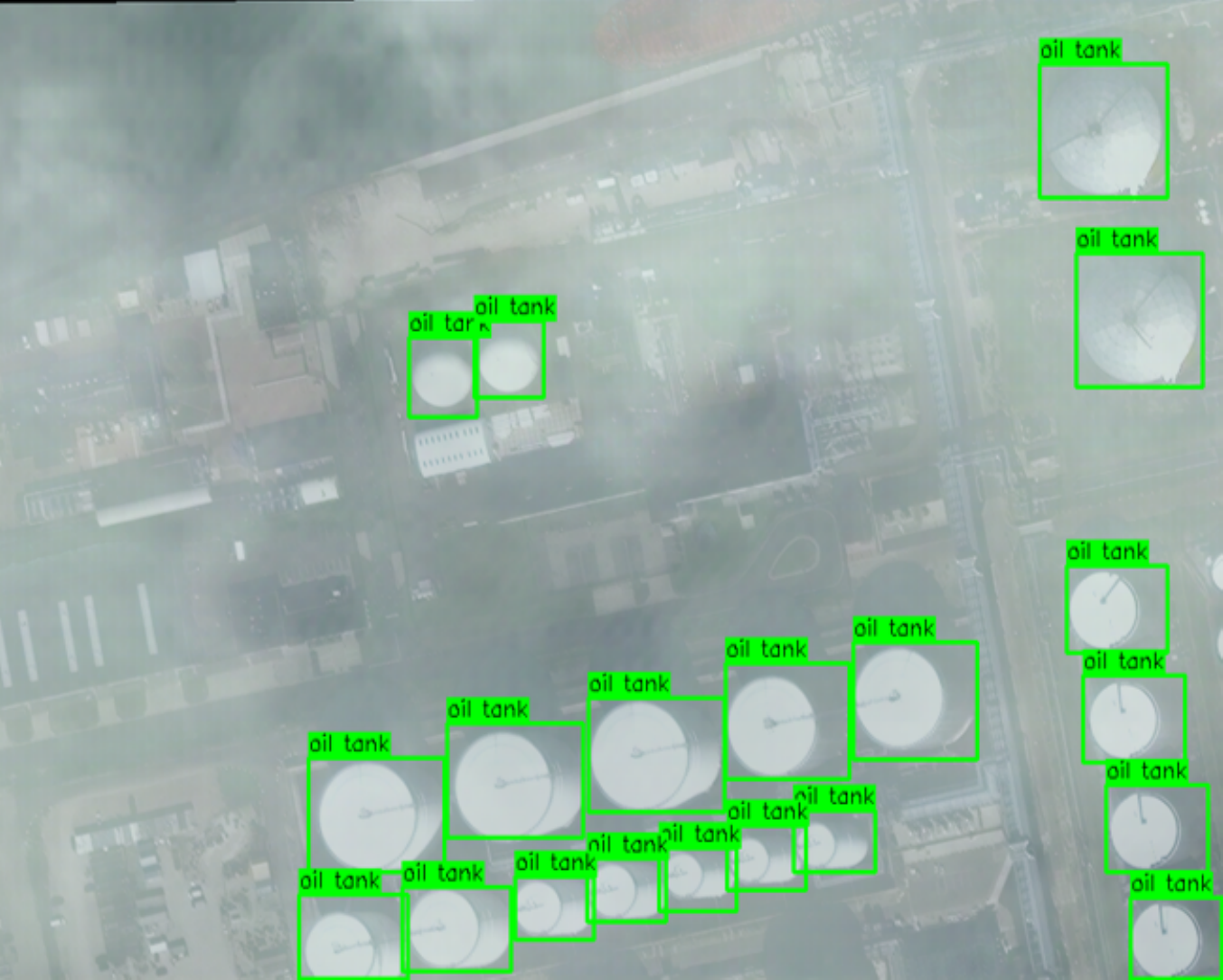} & 
        \includegraphics[width=\linewidth]{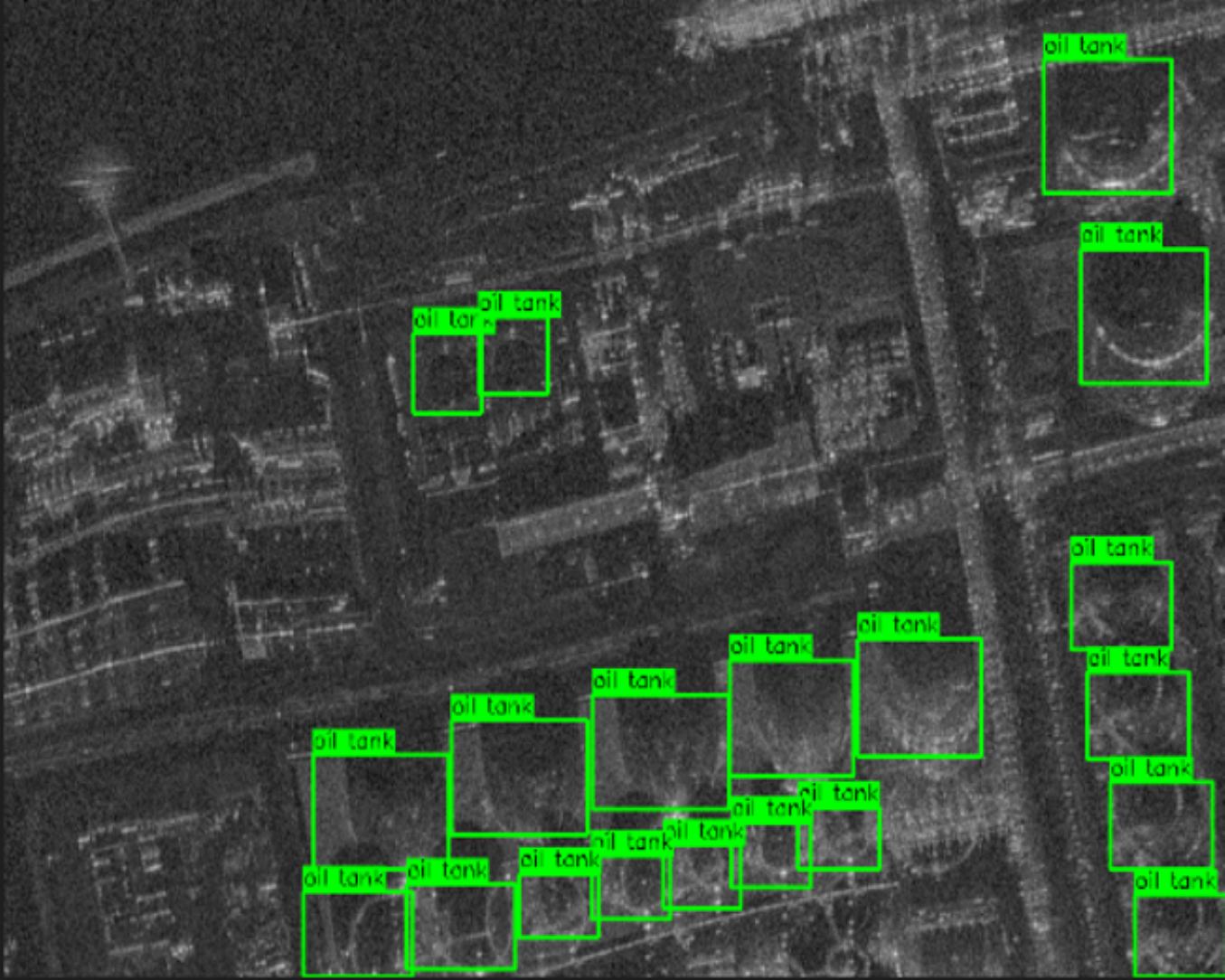} & 
        \includegraphics[width=\linewidth]{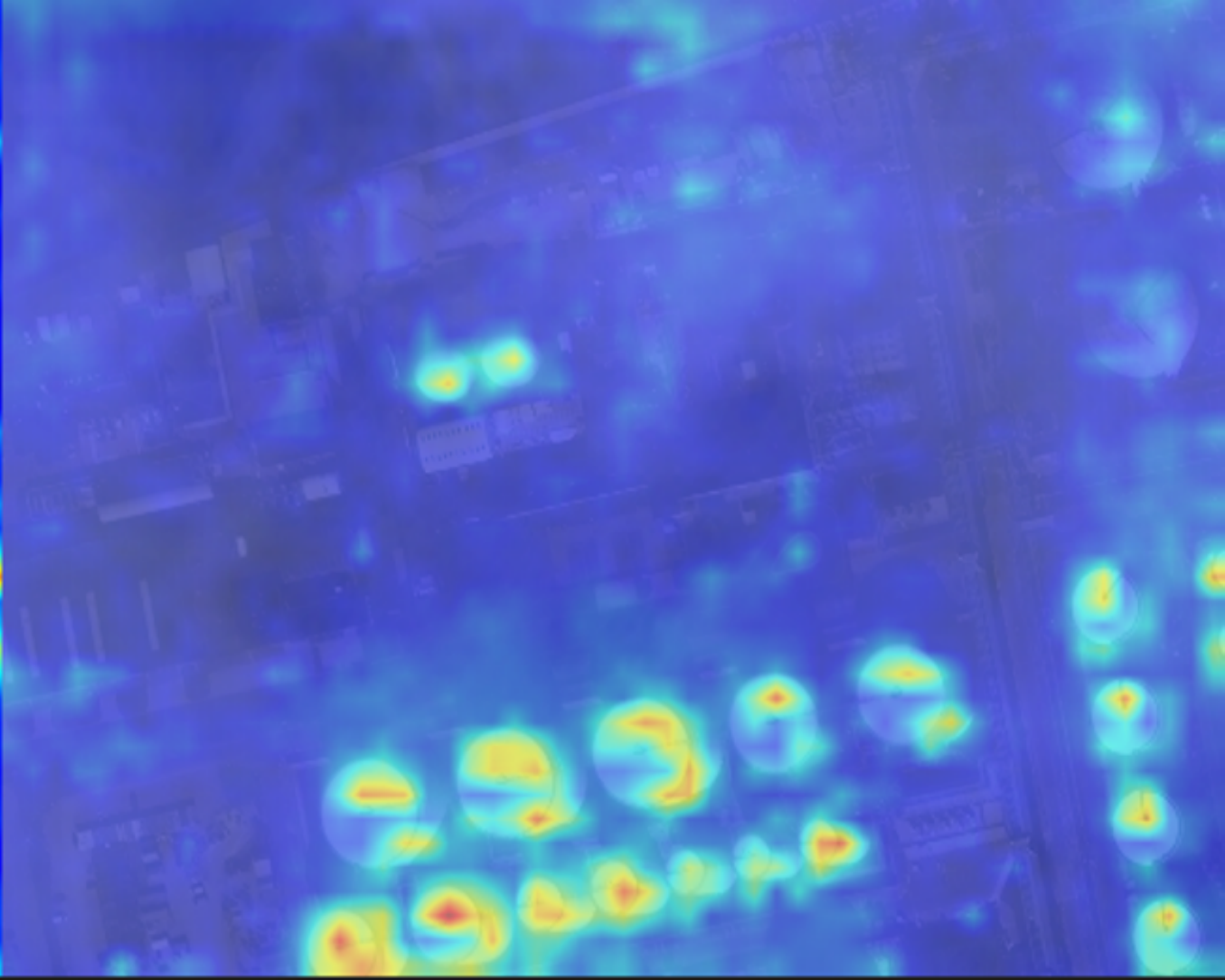} & 
        \includegraphics[width=\linewidth]{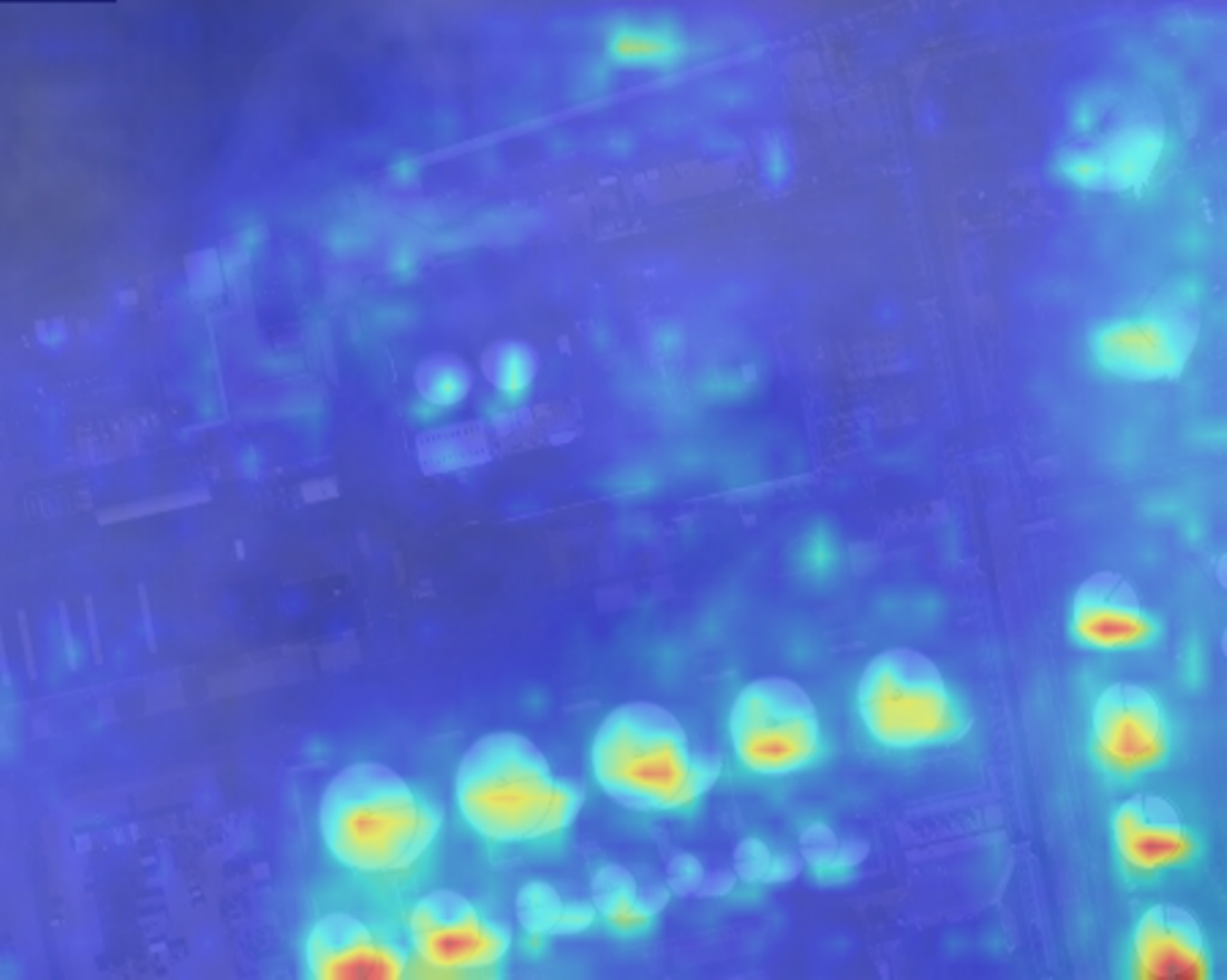} & 
        \includegraphics[width=\linewidth]{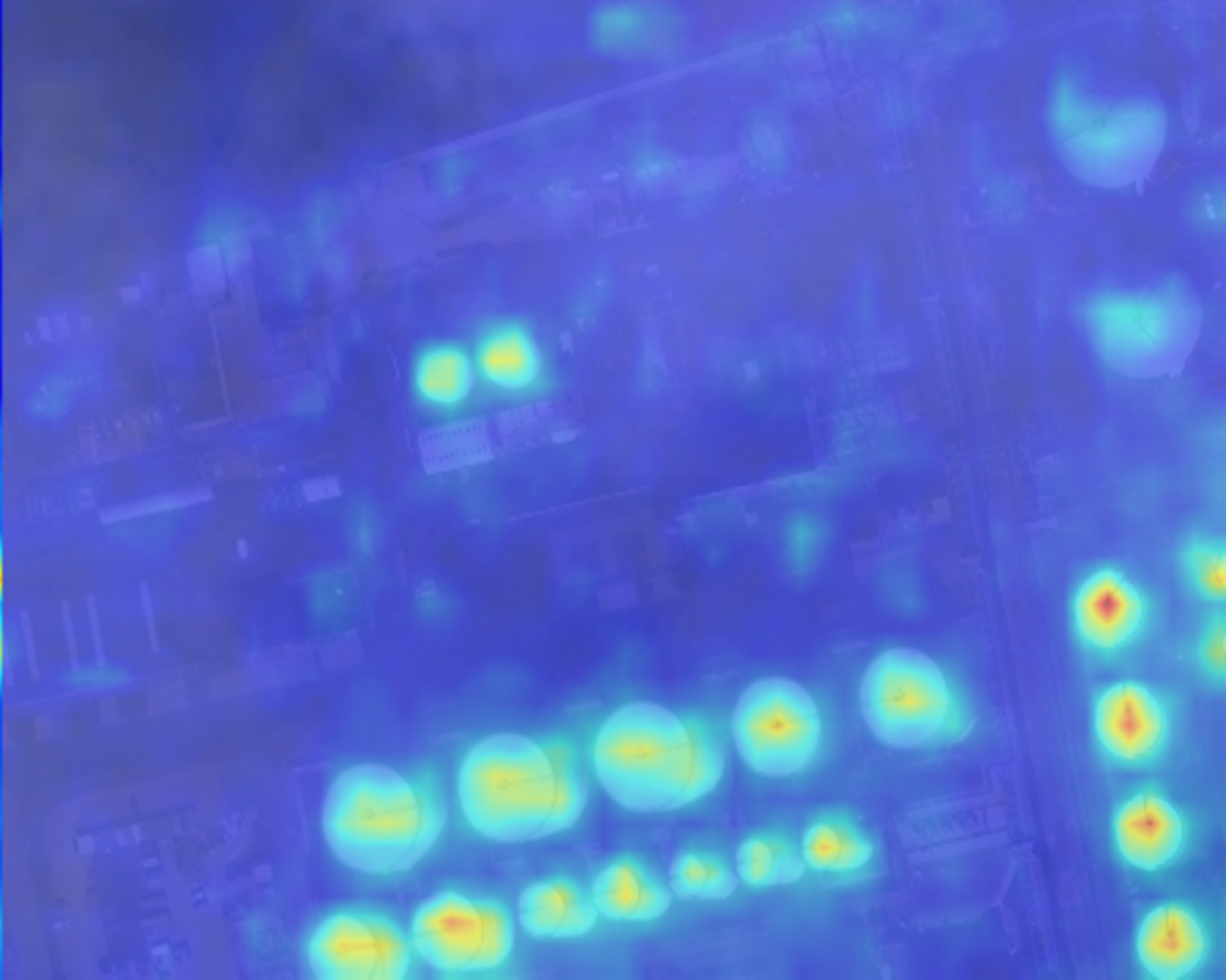} & 
        \includegraphics[width=\linewidth]{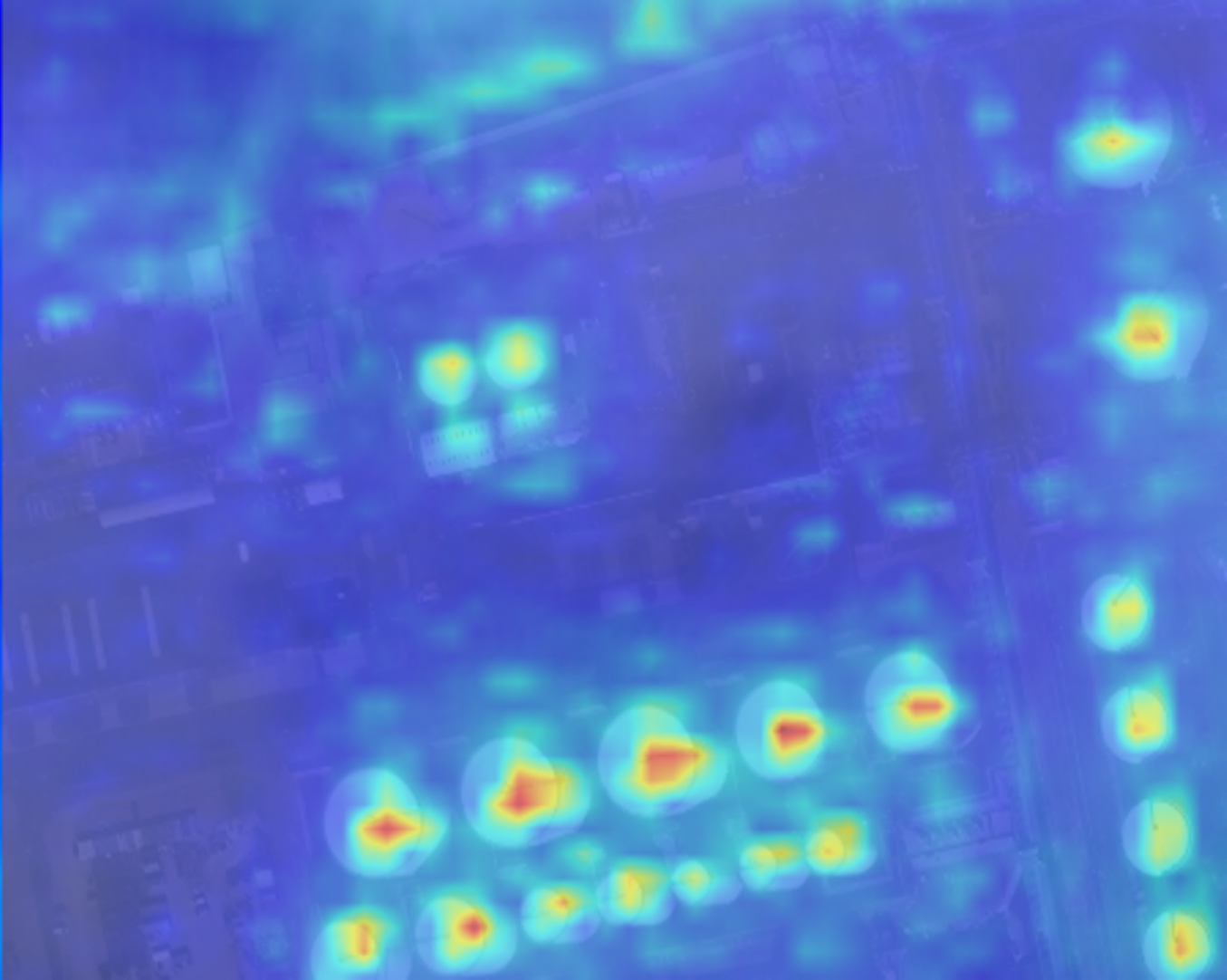} \\

        \includegraphics[width=\linewidth]{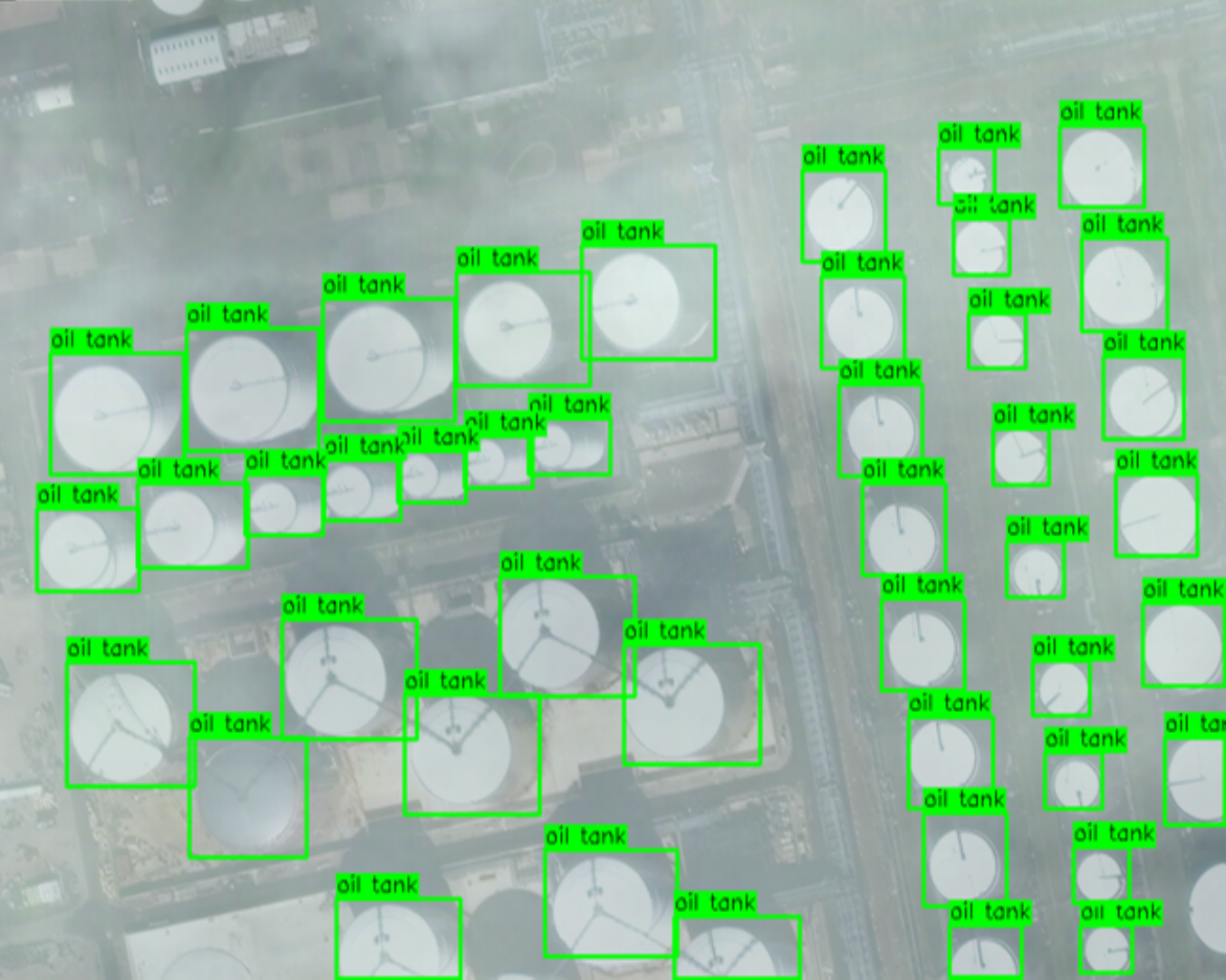} & 
        \includegraphics[width=\linewidth]{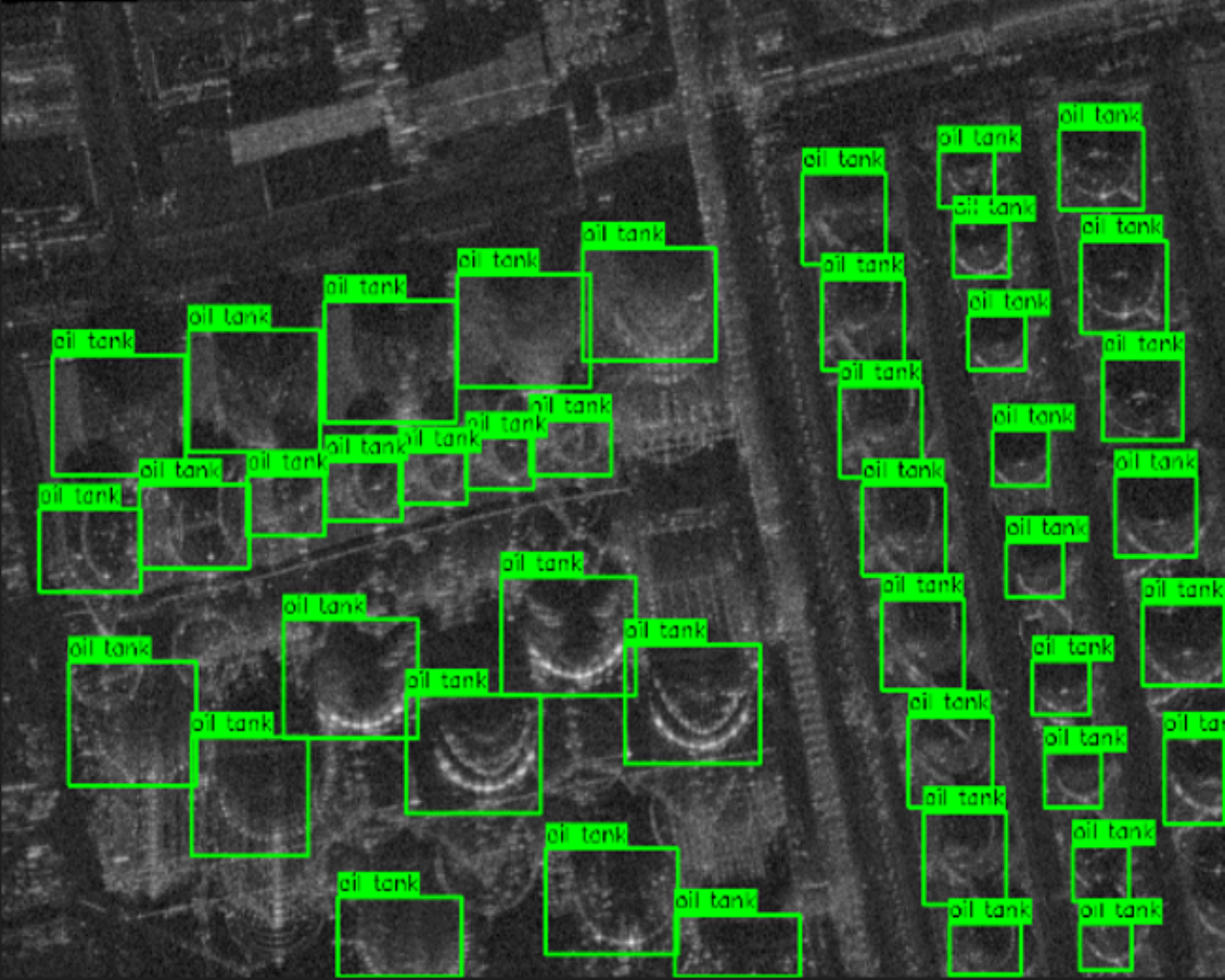} & 
        \includegraphics[width=\linewidth]{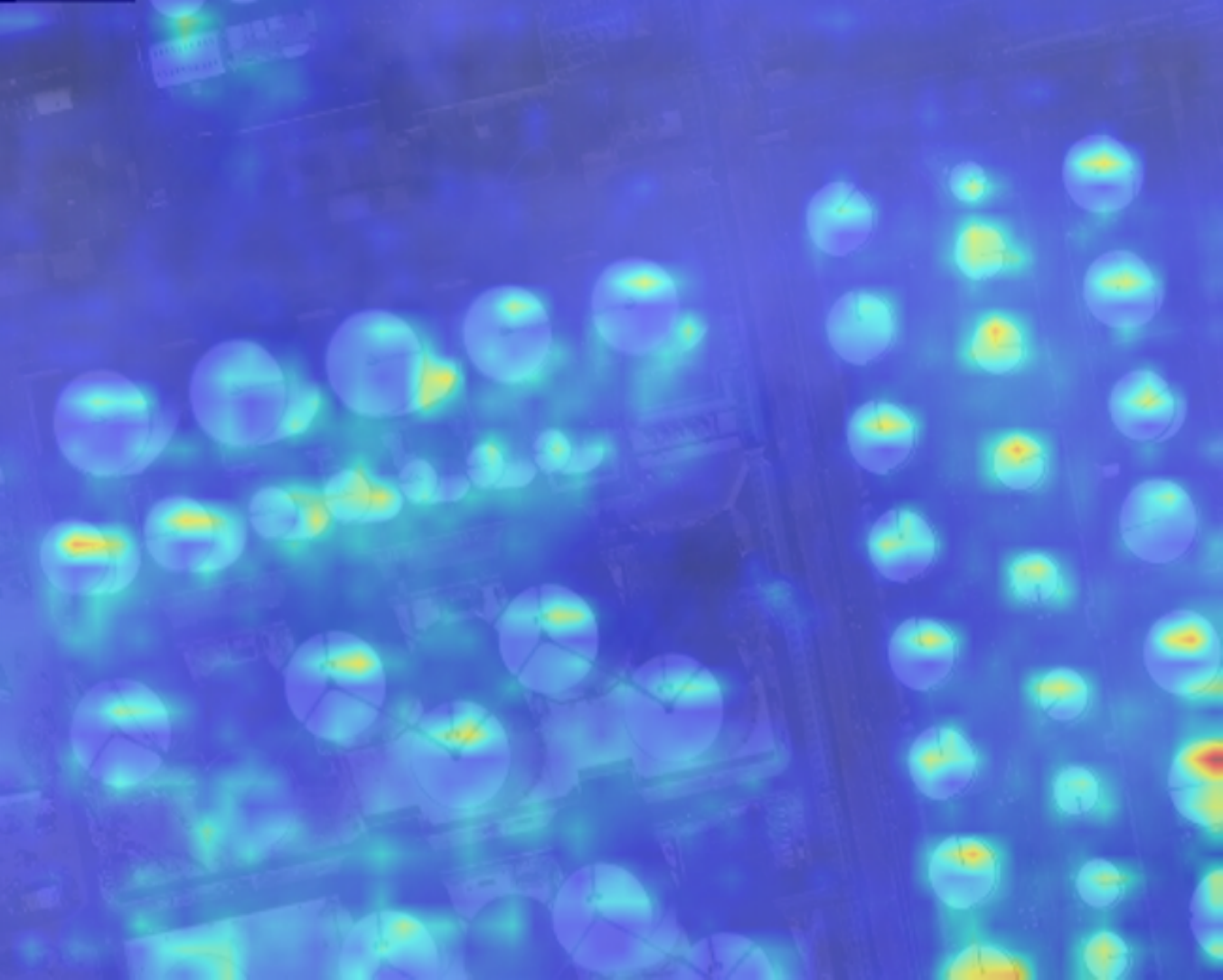} & 
        \includegraphics[width=\linewidth]{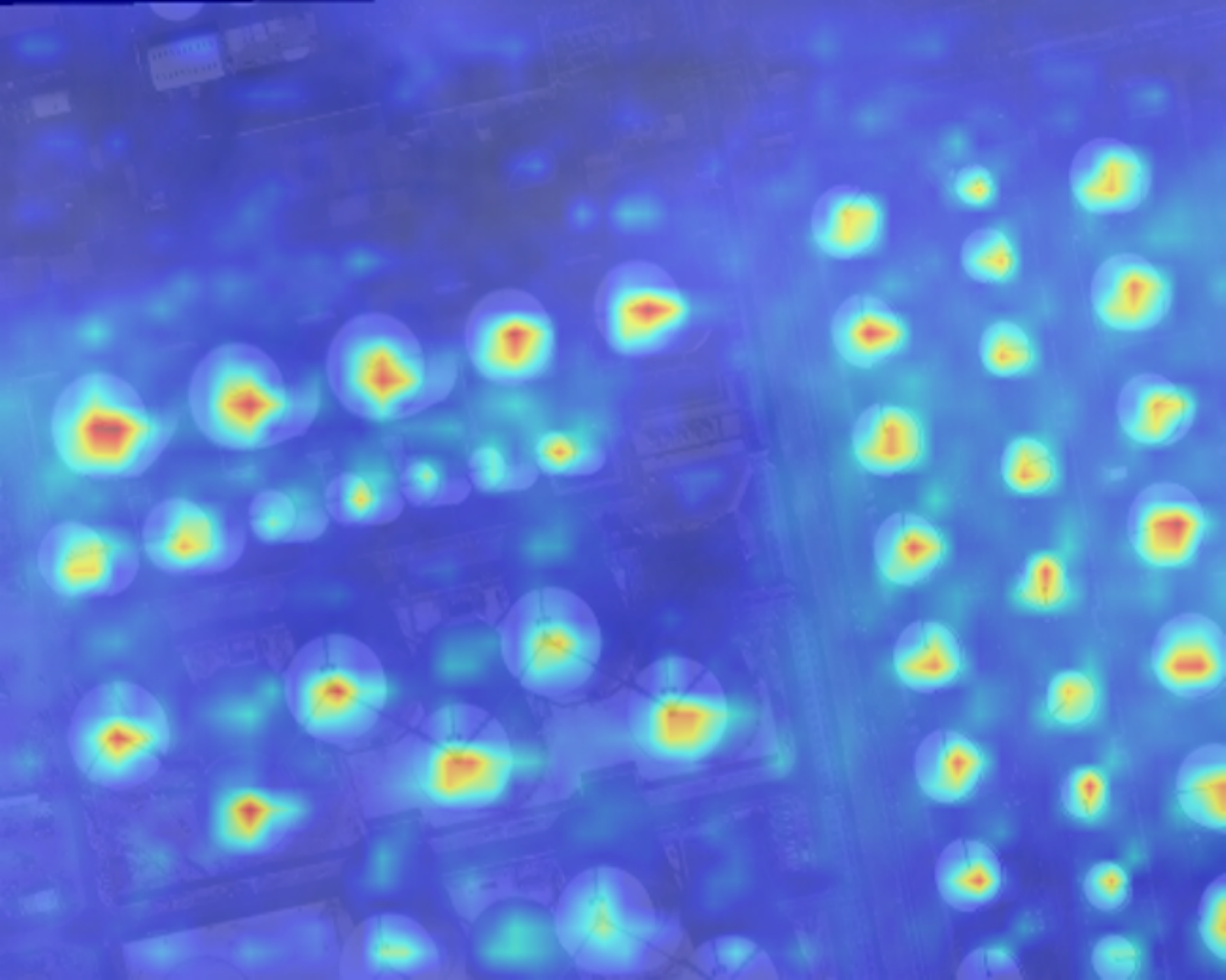} & 
        \includegraphics[width=\linewidth]{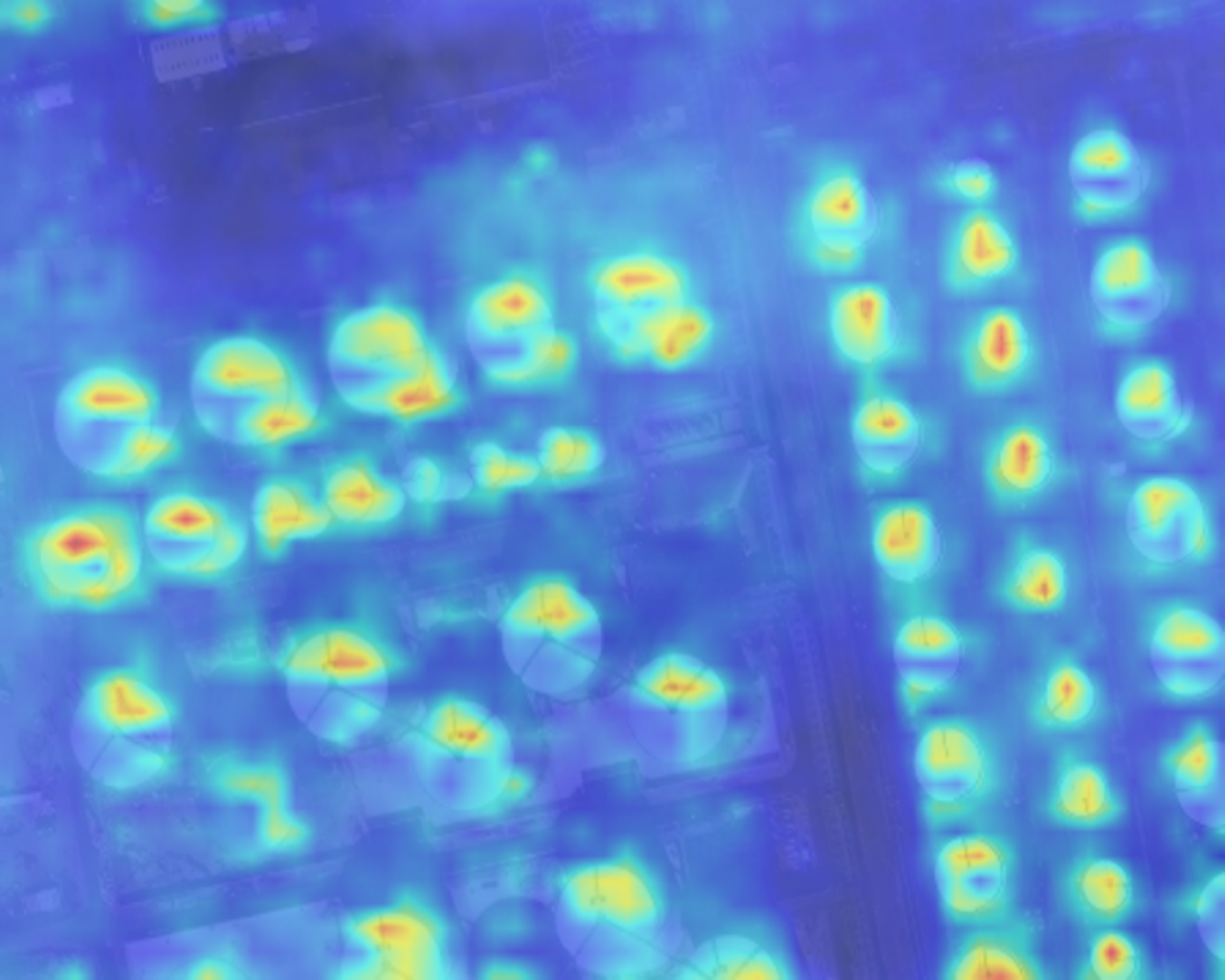} & 
        \includegraphics[width=\linewidth]{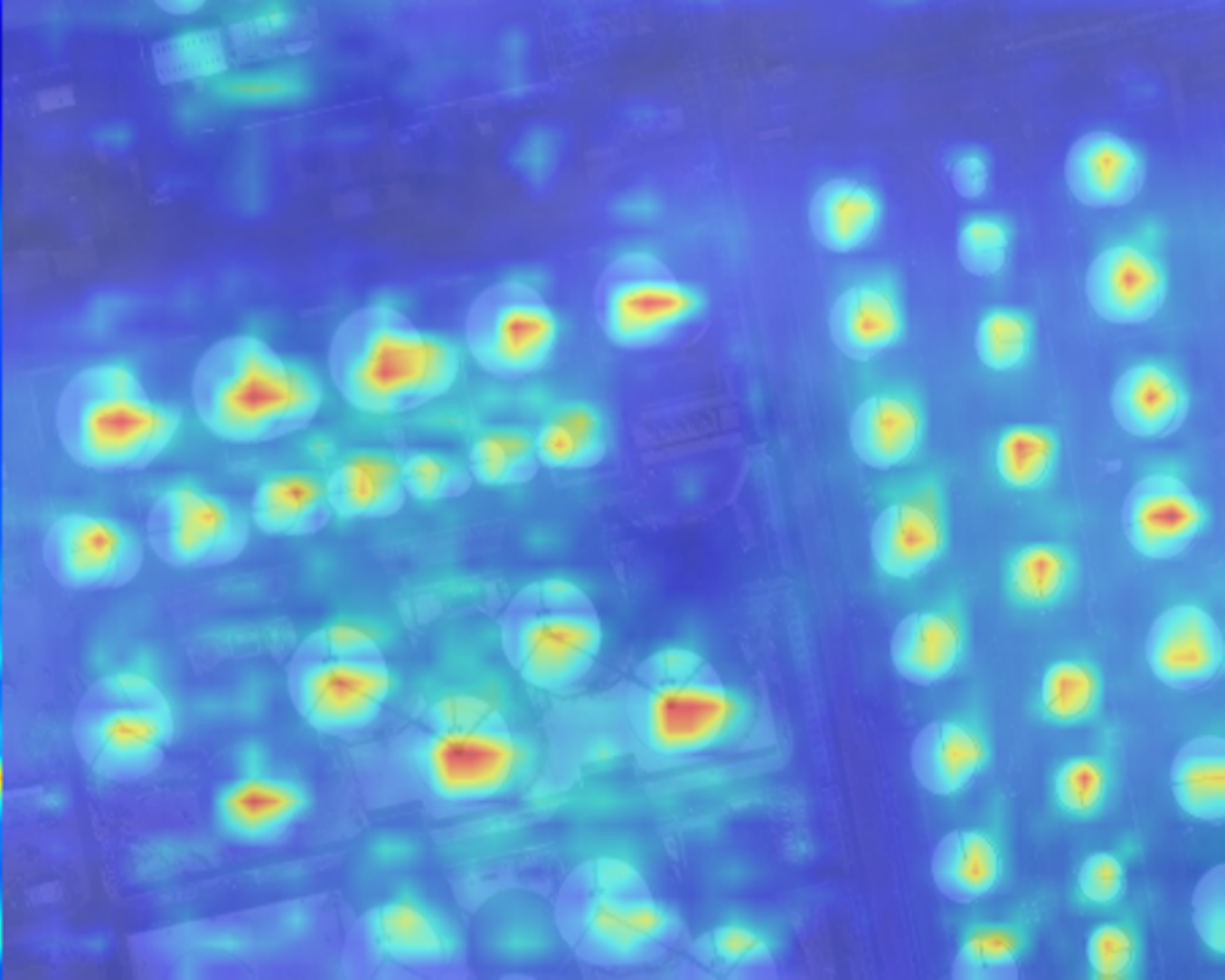} \\

        (a) & (b) & (c) & (d) & (e) & (f) \\
    \end{tabular}
    \caption{Visualization of the learning attention map using Grad-CAM under the scenario of MR = 0.3 and INN reconstruction. (a) RGB GT Labels. (b) SAR GT Labels. (c) Baseline. (d) w/ DMQA. (e) w/ OCNF. (f) Ours. }
    \label{fig:6}
\end{figure*}

\noindent\textbf{Performance on SpaceNet6-OTD-Fog dataset}. 
In the Zero-filling setting, as reported in Table~\ref{tab:1}, QDFNet consistently achieves the highest mAP50 and mAP across all missing rates, showing strong robustness to missing modalities. At MR = 0.1, QDFNet achieves 80.8\%/42.9\%, and still maintains 73.8\%/37.1\% at MR = 0.5. In comparison, MMIDet shows a larger performance drop, decreasing from 78.9\%/40.2\% to 72.4\%/35.6\%.
ICAFusion and CALNet remain competitive when the missing rate is low, reporting 78.8\%/37.9\% and 78.4\%/40.3\% at MR = 0.1. However, their performance decreases faster as the missing rate increases. TFDet and SuperYOLO are more sensitive to missing modalities: TFDet drops from 75.3\%/37.7\% to 63.7\%/31.6\%, and SuperYOLO decreases from 73.3\%/39.0\% to 66.3\%/35.5\% when MR increases from 0.1 to 0.5. These results indicate that QDFNet maintains accuracy more effectively when inputs are missing, while other detectors experience larger performance degradation.Under the INN reconstruction setting, missing modality is first reconstructed through invertible neural networks before being fed into the detector

Under the INN reconstruction setting, missing modality is first reconstructed through invertible neural networks before being fed into the detector. As presented in Table~\ref{tab:2}, most models benefit from reconstruction, suggesting that partial semantic recovery mitigates performance loss. However, the ability to utilize reconstructed information differs across detectors. QDFNet again achieves the best results at all missing rates, with stable performance trends. At MR = 0.1, QDFNet reaches 81.9\%/42.9\%, and still leads at MR = 0.5 with 74.2\%/37.6\%. MMIDet decreases to 72.1\%/35.2\%, and TFDet falls to 65.3\%/31.1\%.
The results show that QDFNet can better exploit reconstructed features and preserve discriminative representations, especially at high missing rates. In contrast, CFT and CALNet suffer more obvious performance drops, indicating limited effectiveness of their reconstruction strategies.

\noindent\textbf{Performance on OGSOD-2.0 dataset}. 
In the Zero-filling setting, as presented in Table~\ref{tab:1}, QDFNet achieves the best results across all missing rates on OGSOD-2.0. The advantage becomes clearer as the missing rate increases. At MR = 0.4, QDFNet obtains 65.0\%/33.5\%, exceeding ICAFusion by 3.5 points in mAP50 with 61.5\%/33.2\%. At MR = 0.5, QDFNet reaches 61.8\%/30.1\%, while ICAFusion shows 58.1\%/29.7\%, indicating a continued performance margin under severe modality missing. CFT and MMIDet experience a significant performance drop, from 79.0\%/41.5\% to 52.6\%/24.6\%, and from 77.7\%/39.3\% to 52.8\%/25.3\%, respectively. 

Under the INN reconstruction setting, as shown in Table~\ref{tab:2}, QDFNet again achieves the highest mAP50 and mAP at all missing rates. At MR = 0.1, QDFNet records 76.9\%/38.3\% and maintains leading performance at MR = 0.5 with 64.8\%/32.2\%. In contrast, MMIDet shows a more substantial decline, decreasing from 75.6\%/38.1\% to 58.4\%/29.4\%, reflecting weaker adaptation to reconstructed modalities. ICAFusion performs well at low missing rates, reaching 76.0\%/38.5\% at MR = 0.1, but its performance drops faster as MR increases, suggesting limited robustness in its compensation mechanism. TFDet and SuperYOLO exhibit pronounced sensitivity to missing modality information. TFDet drops from 70.1\%/39.1\% to 57.1\%/30.3\%, and SuperYOLO from 72.9\%/36.0\% to 52.1\%/28.3\% as the missing rate increases from 0.1 to 0.5. 

Overall, results on OGSOD-2.0 show that QDFNet provides consistently higher robustness to missing modalities. QDFNet not only matches the best detectors when all modalities are available, but also maintains a clear margin as the missing rate increases. Competing methods such as ICAFusion, CFT, MMIDet, TFDet, and SuperYOLO show more severe performance drops, especially at high missing rates or when relying on reconstructed inputs. These results confirm that QDFNet preserves more discriminative information and makes better use of missing or reconstructed features.

\begin{figure}[htb]
    \centering
    \includegraphics[width=1\linewidth]{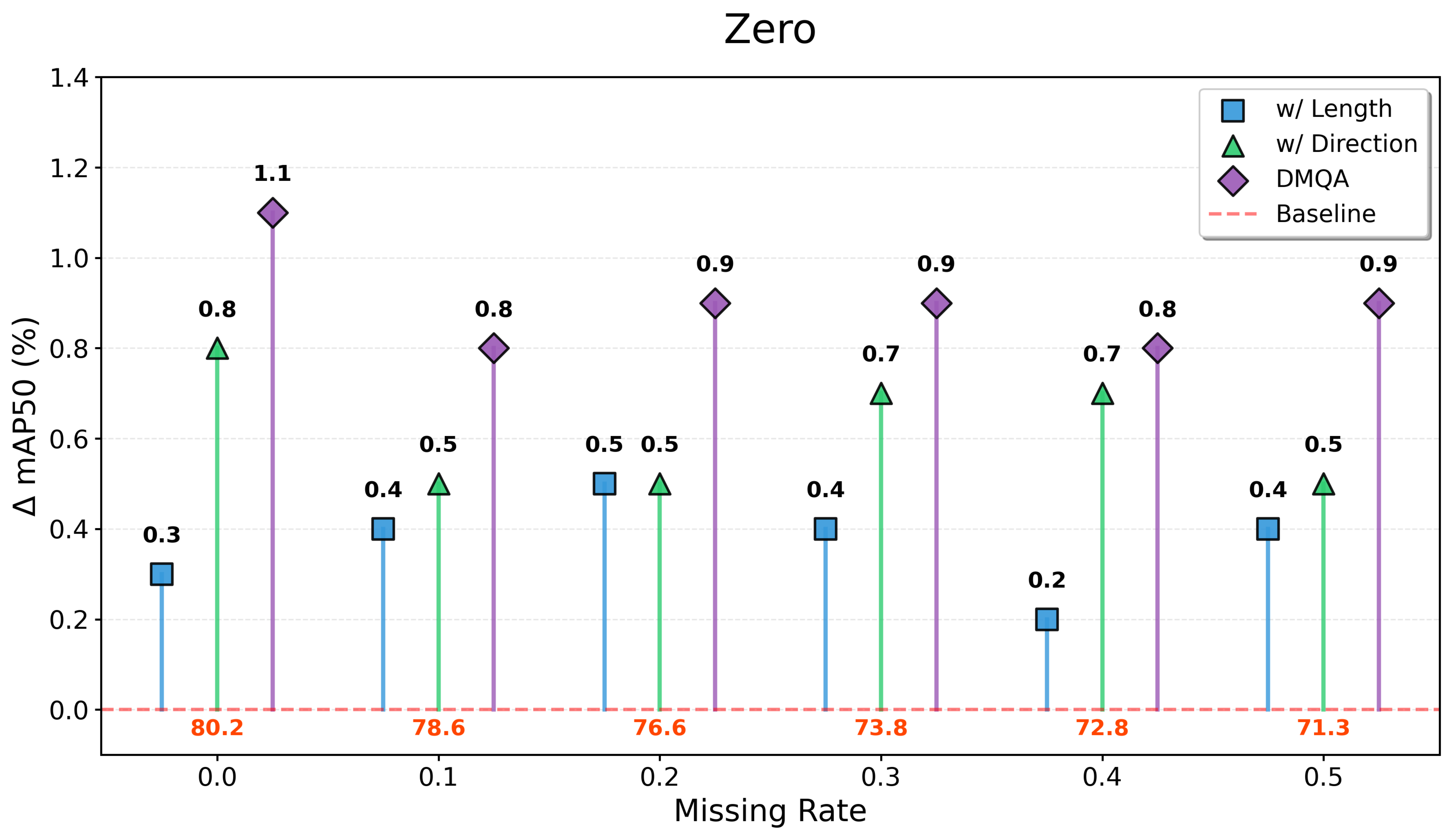}
    \caption{Ablation improvements of the DMQA components over the Baseline under the Zero-filling setting. The orange numerical values shown below the zero line represent the absolute Baseline performance.}
    \label{fig:7}
\end{figure}
\begin{figure}[htb]
    \centering
    \includegraphics[width=1\linewidth]{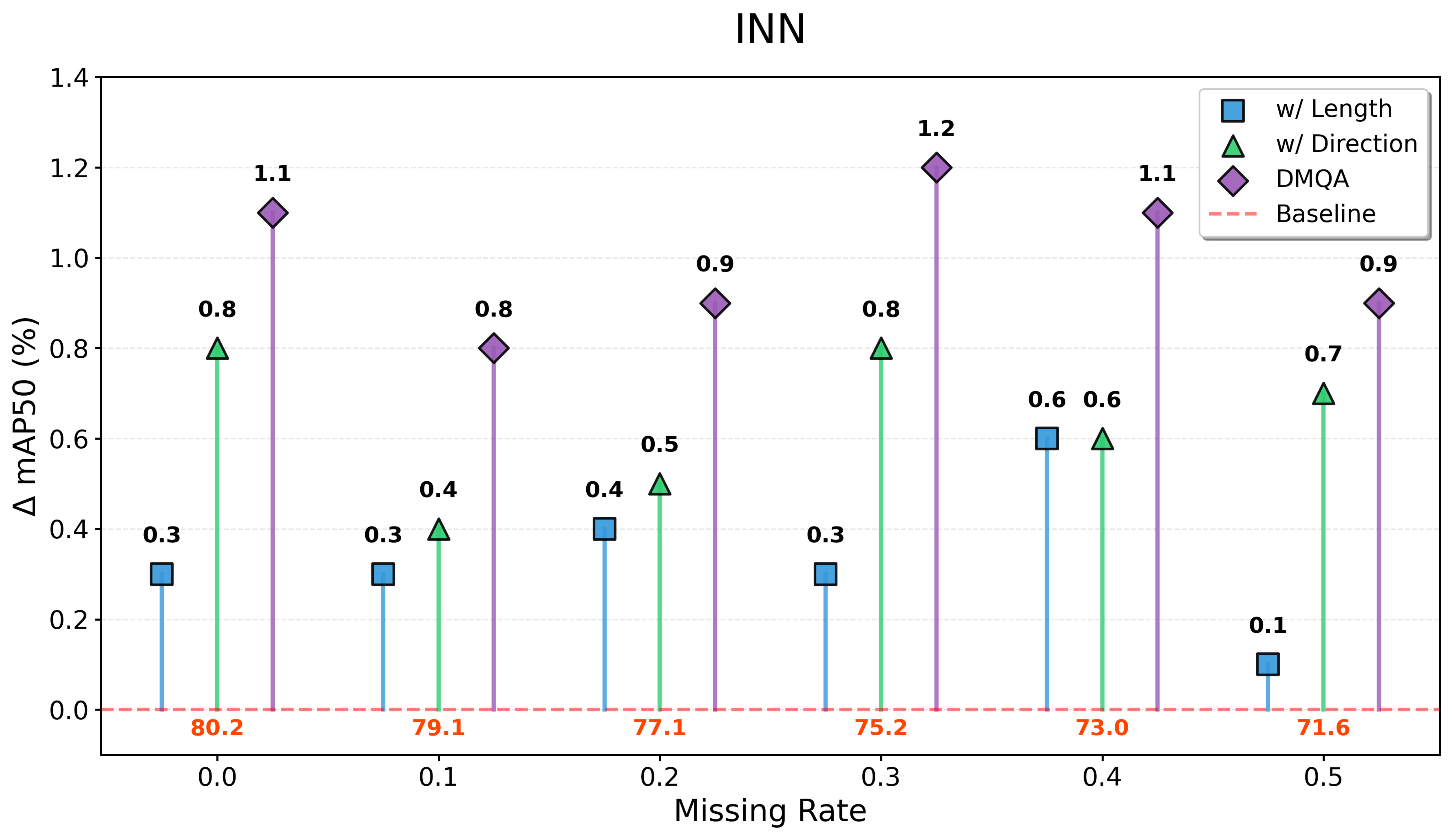}
    \caption{Ablation improvements of the DMQA components over the Baseline under the INN reconstruction setting. The orange numerical values shown below the zero line represent the absolute Baseline performance.}
    \label{fig:8}
\end{figure}

\subsection{Ablation Study}
To evaluate the contribution of each component, ablation studies are performed on the SpaceNet6-OTD-Fog dataset, as summarized in Table~\ref{tab:3} and Table~\ref{tab:4}. We adopt mAP50 and mAP as evaluation metrics, corresponding to average precision at IoU 0.5 and averaged over multiple IoU thresholds, respectively. 


\begin{figure}[!ht]
    \centering
    \includegraphics[width=1\linewidth]{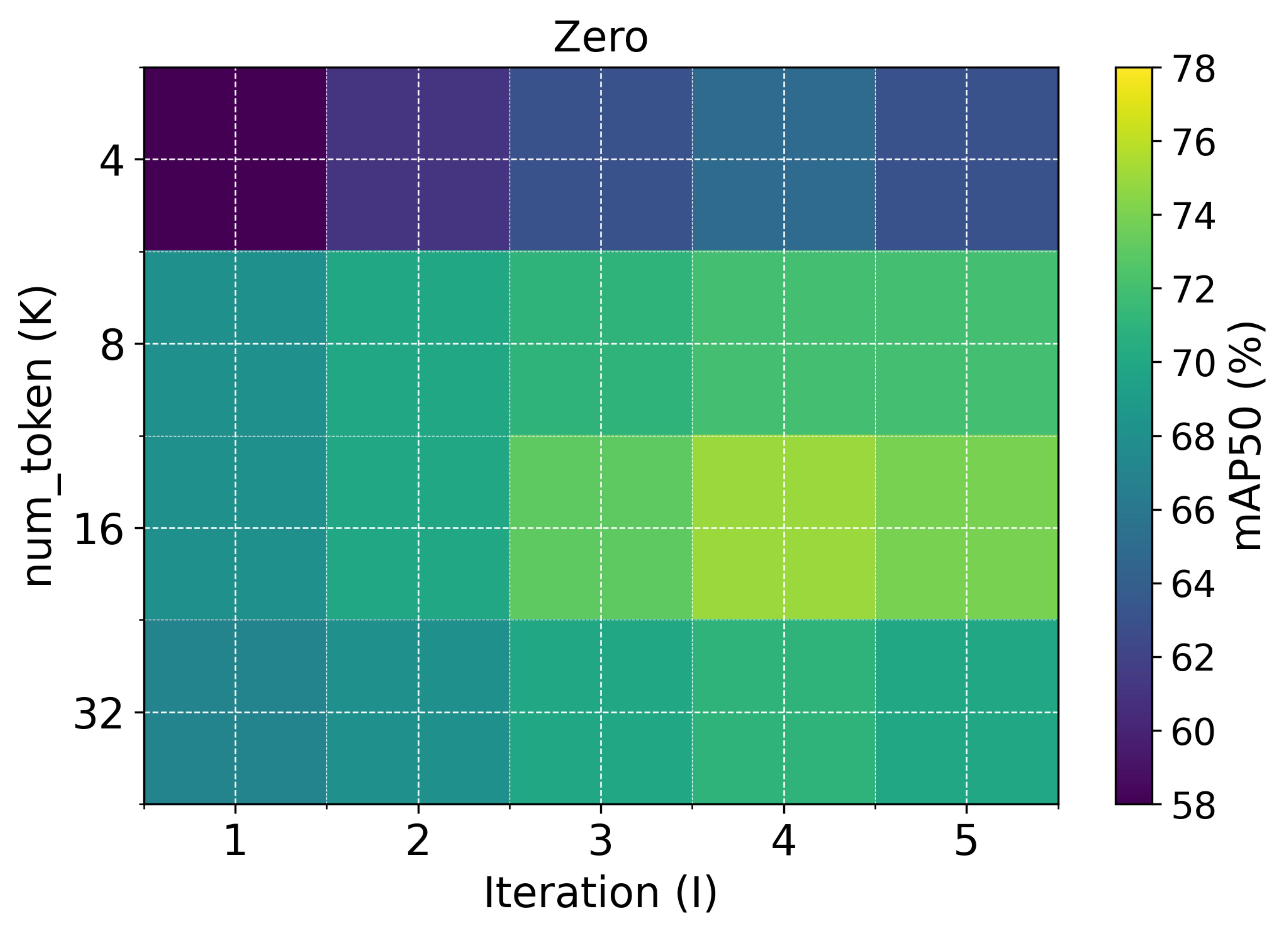}
    \caption{Evaluation of the sensitivity of the DMQA module to the iteration number (I) and number of reliability tokens (K) on the SpaceNet6-OTD-Fog dataset under the Zero-filling setting and MR = 0.3.}
    \label{fig:9}
\end{figure} 

\begin{figure}[!ht]
    \centering
    \includegraphics[width=1\linewidth]{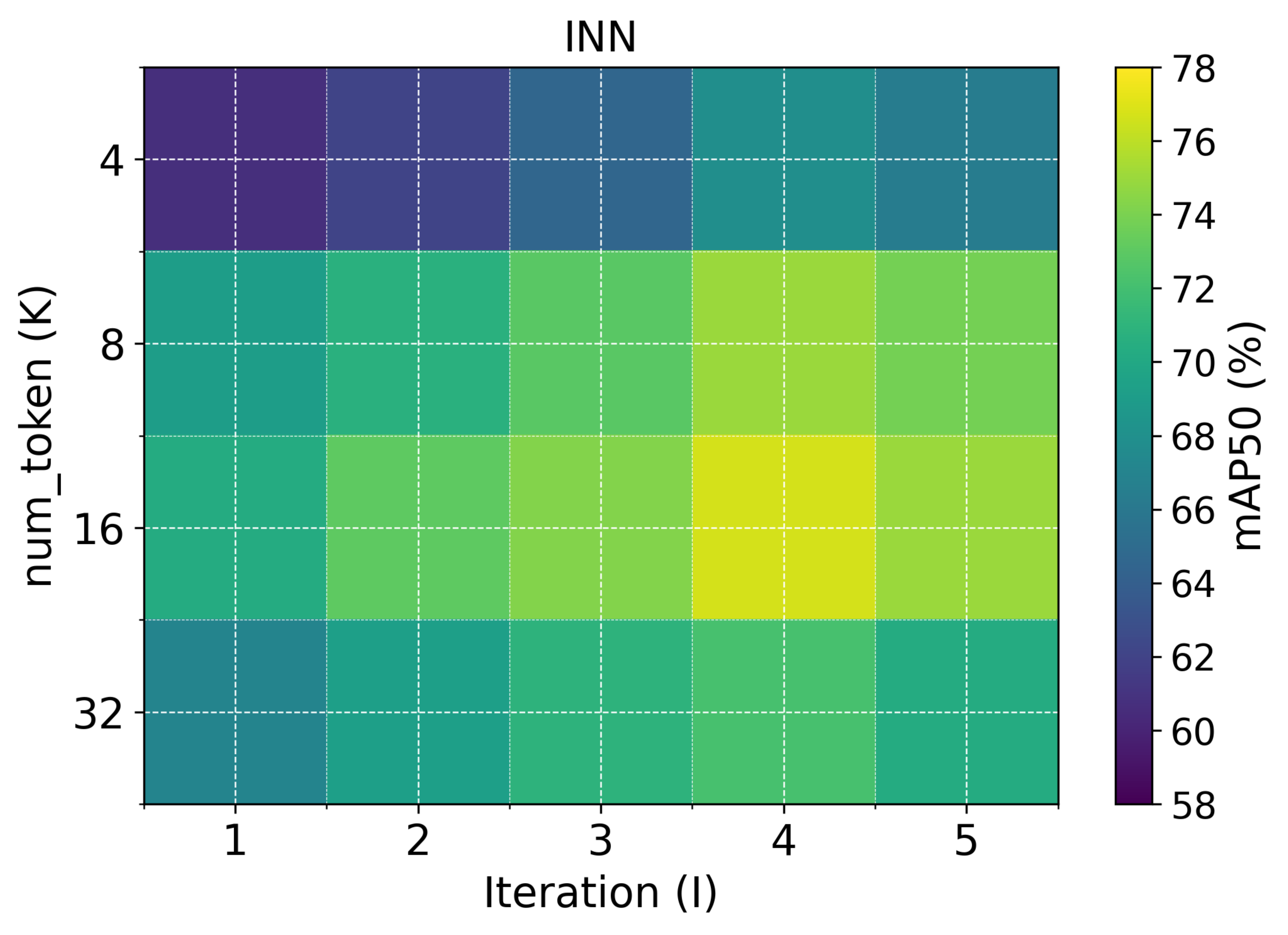}
    \caption{Evaluation of the sensitivity of the DMQA module to the iteration number (I) and number of reliability tokens (K) on the SpaceNet6-OTD-Fog dataset under the INN reconstruction setting and MR = 0.3.}
    \label{fig:10}
\end{figure} 

\begin{figure*}[!ht]
    \centering
    \begin{tabular}{>{\centering\arraybackslash}m{0.18\textwidth}
                    >{\centering\arraybackslash}m{0.18\textwidth}
                    >{\centering\arraybackslash}m{0.18\textwidth}
                    >{\centering\arraybackslash}m{0.18\textwidth}
                    >{\centering\arraybackslash}m{0.18\textwidth}}
        \includegraphics[width=\linewidth]{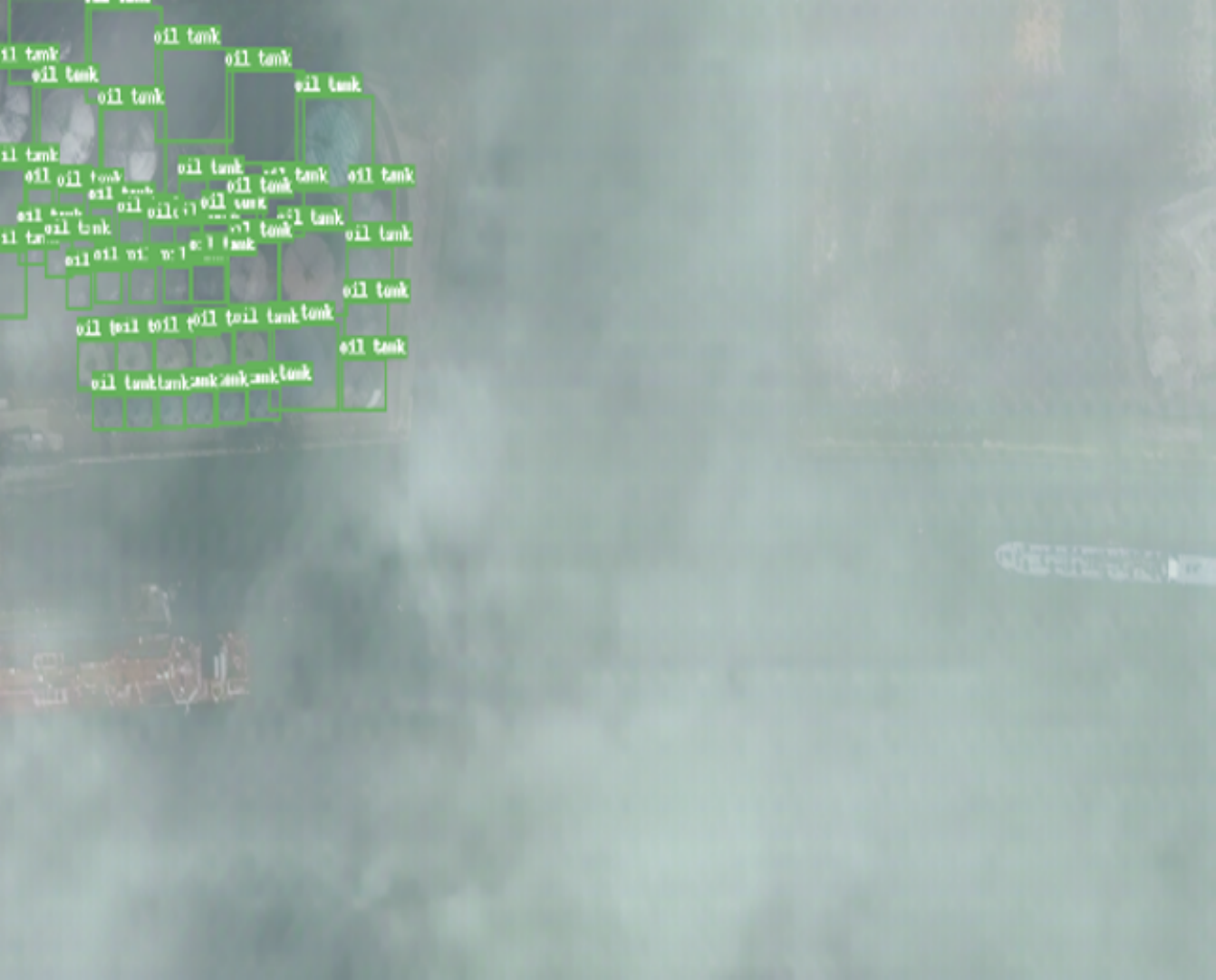} & 
        \includegraphics[width=\linewidth]{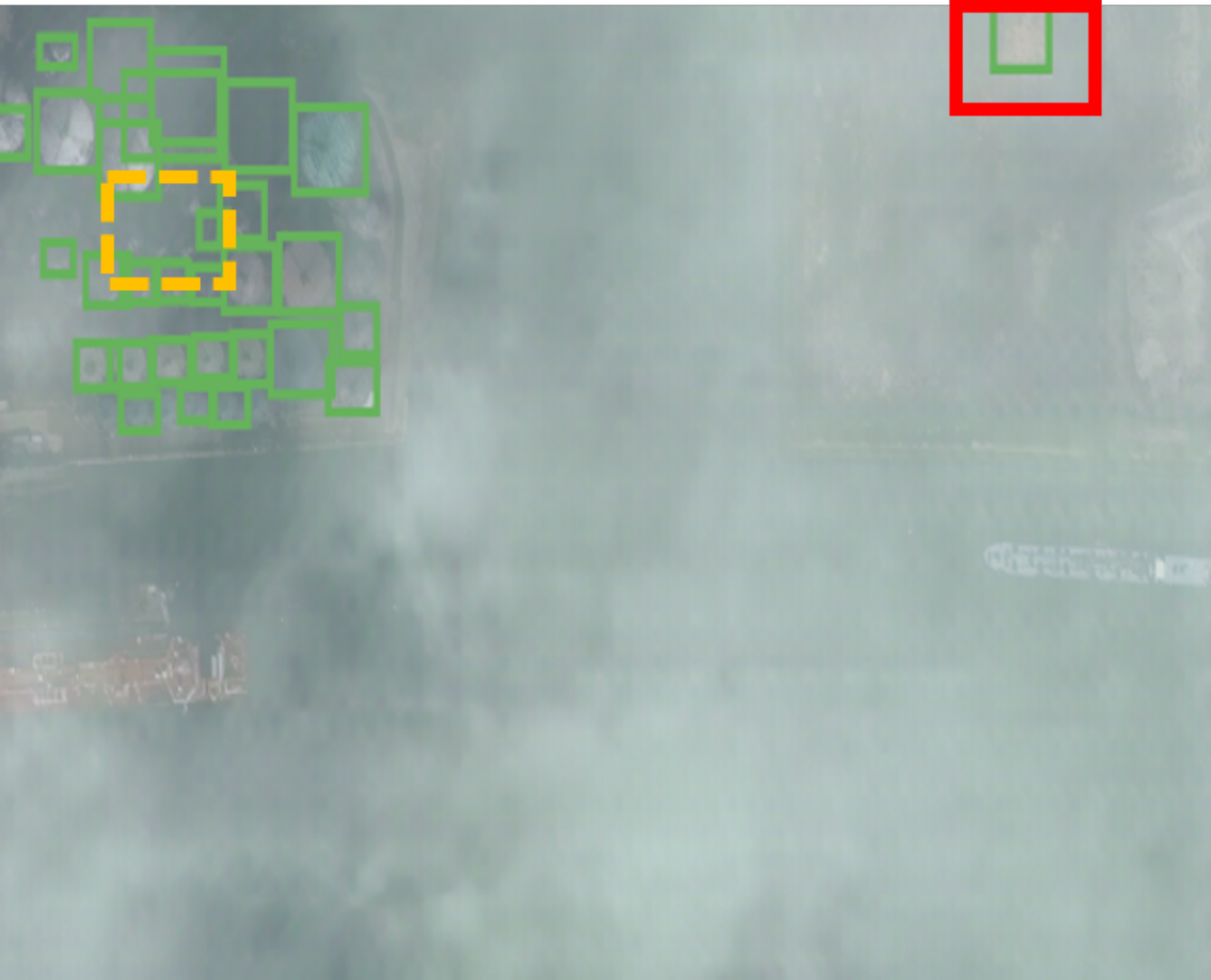} & 
        \includegraphics[width=\linewidth]{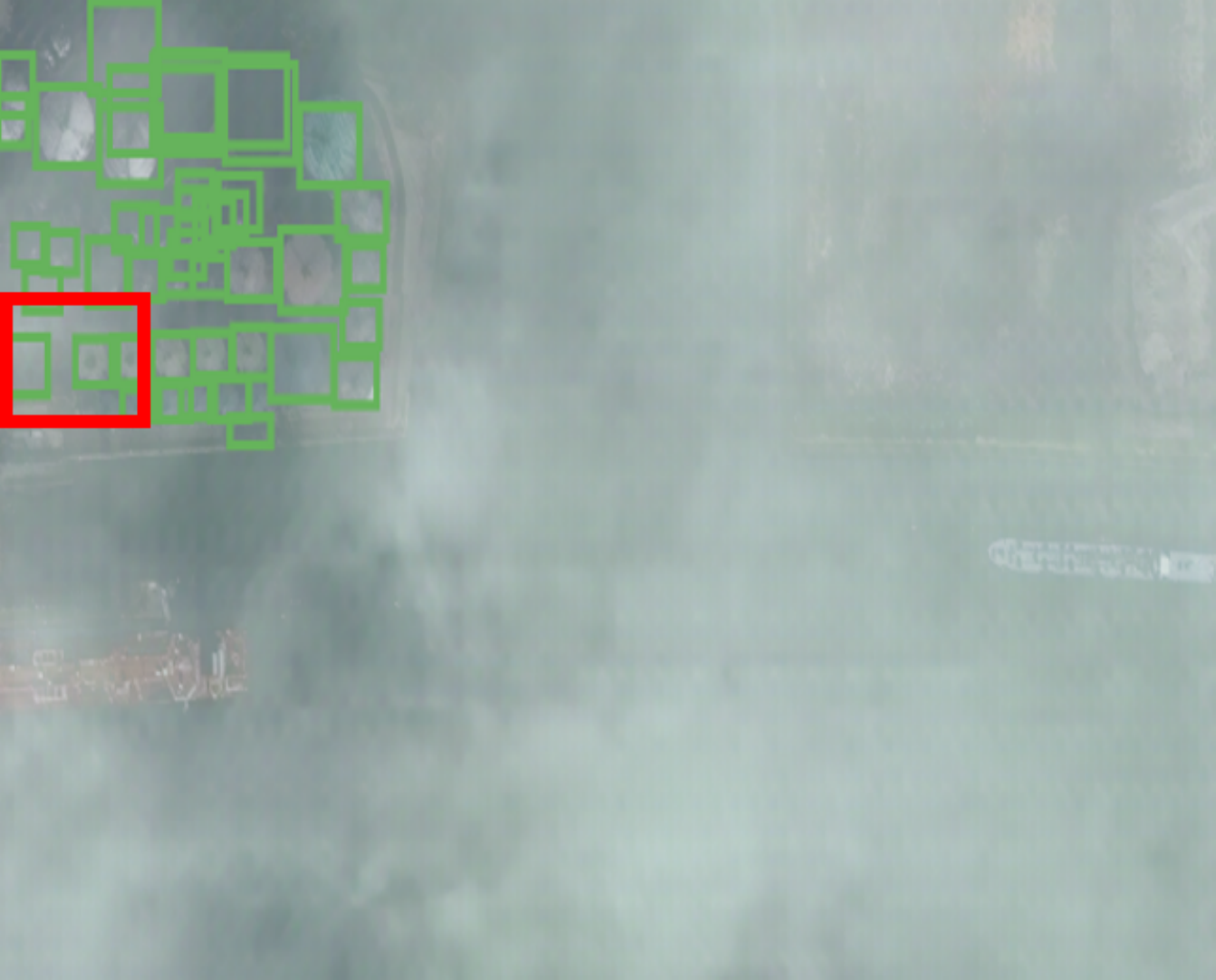} & 
        \includegraphics[width=\linewidth]{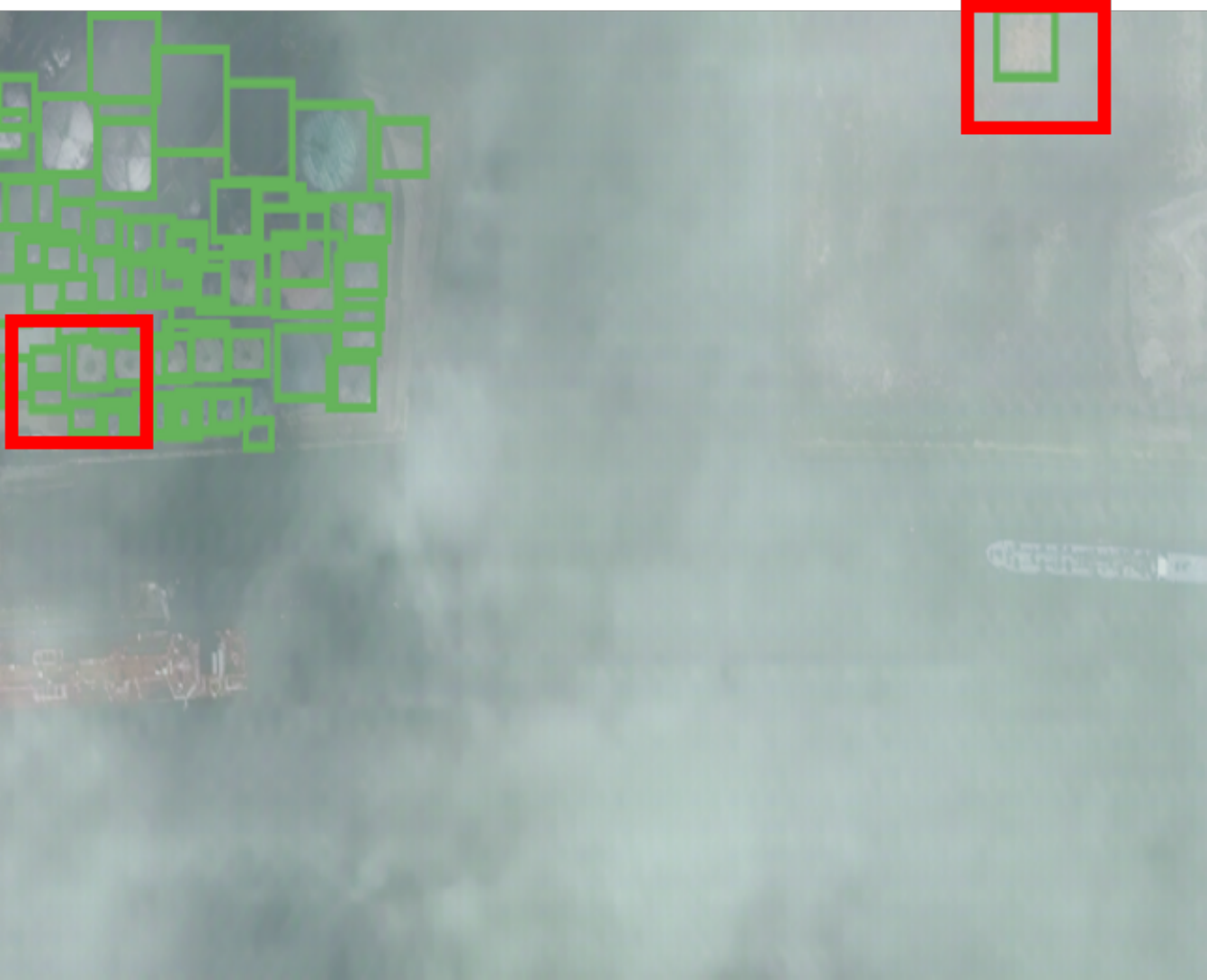} & 
        \includegraphics[width=\linewidth]{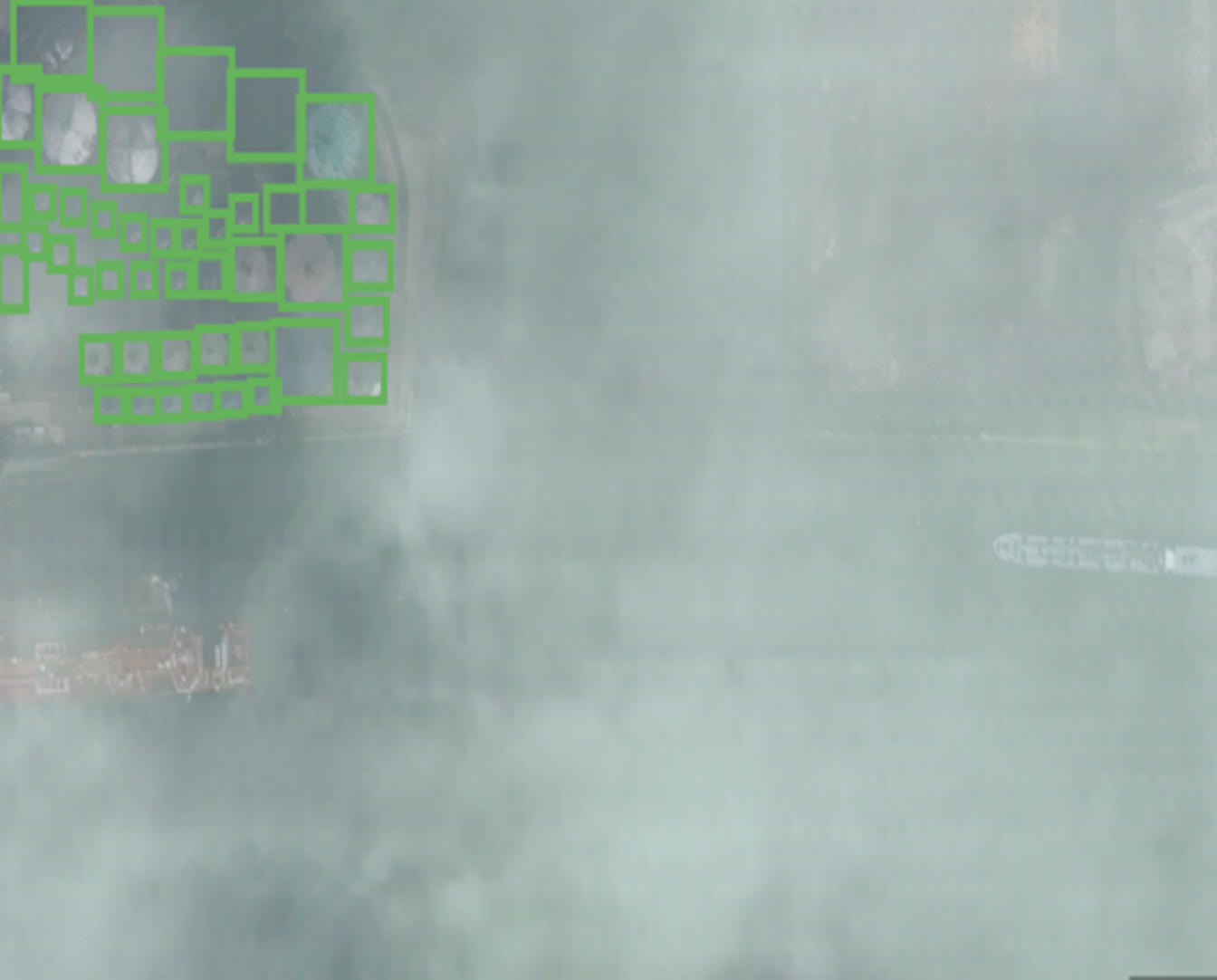} \\
        
        \includegraphics[width=\linewidth]{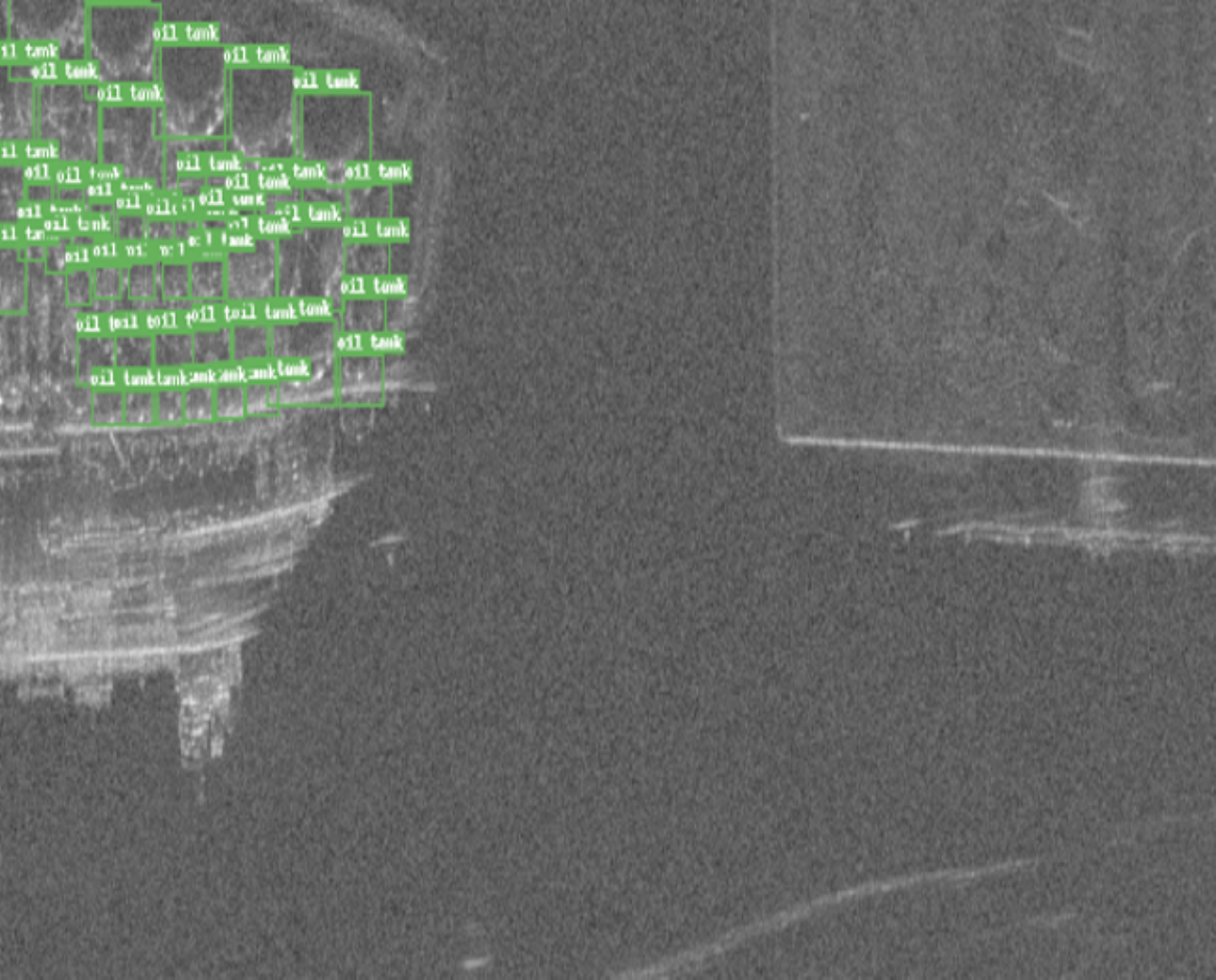} & 
        \includegraphics[width=\linewidth]{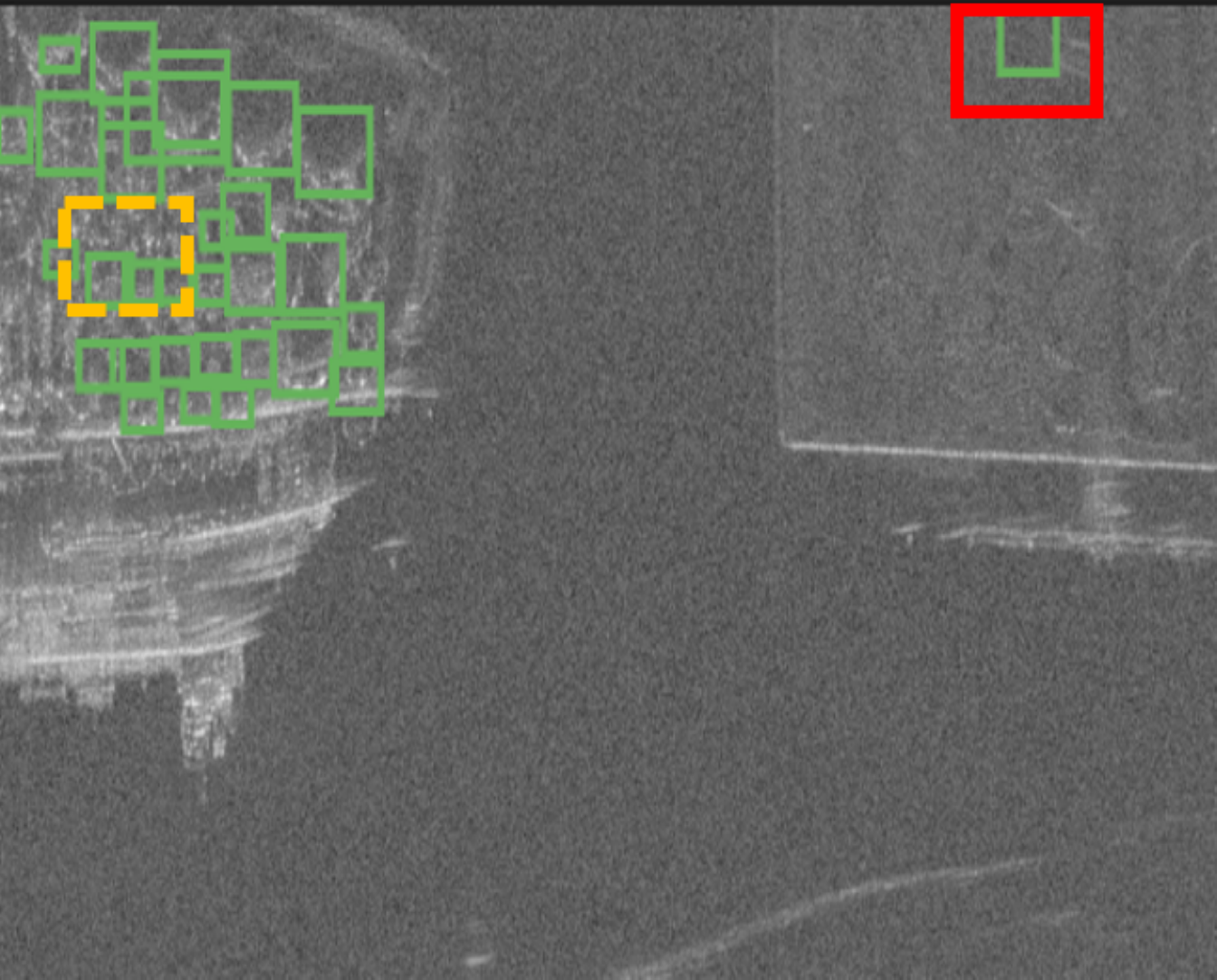} & 
        \includegraphics[width=\linewidth]{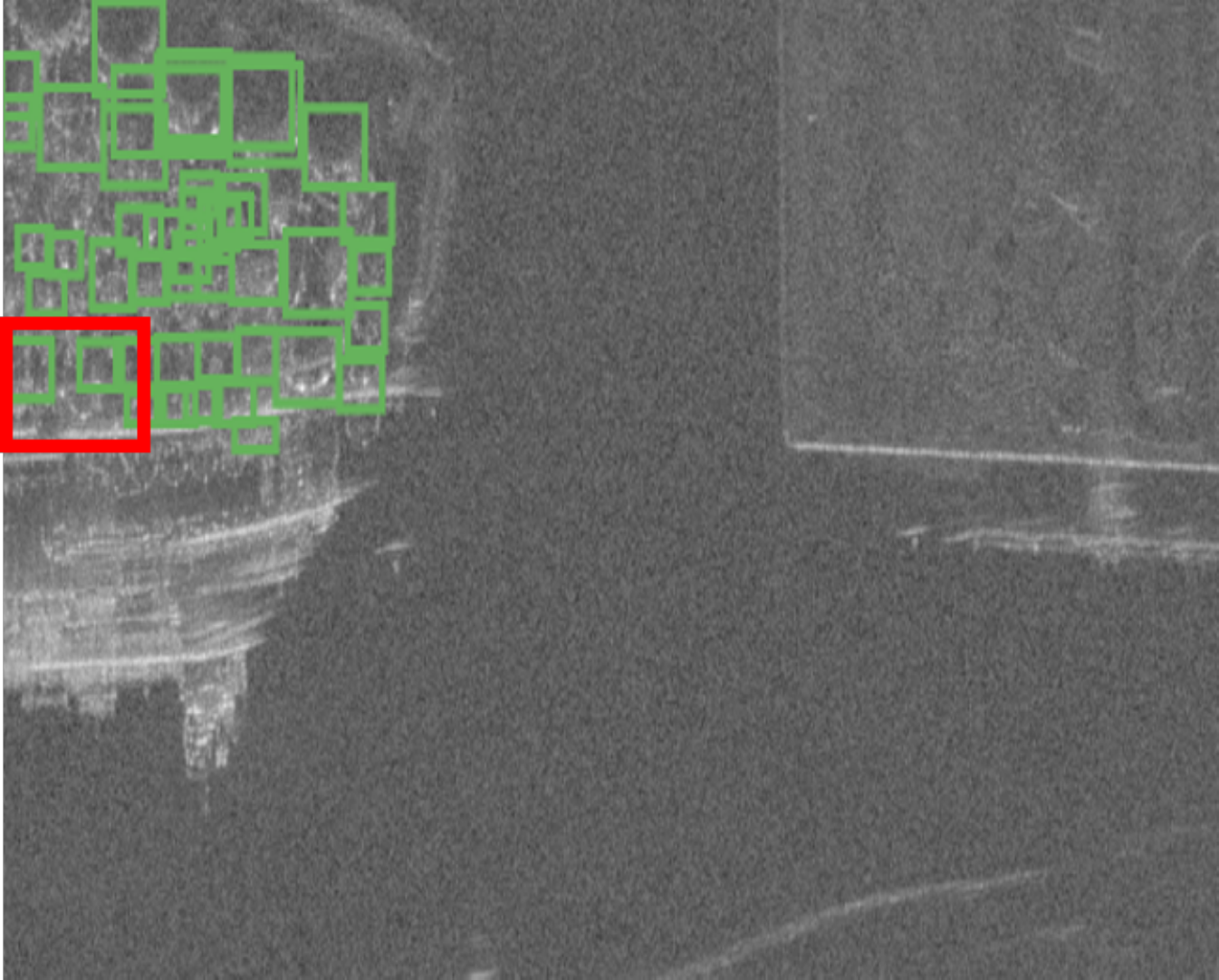} & 
        \includegraphics[width=\linewidth]{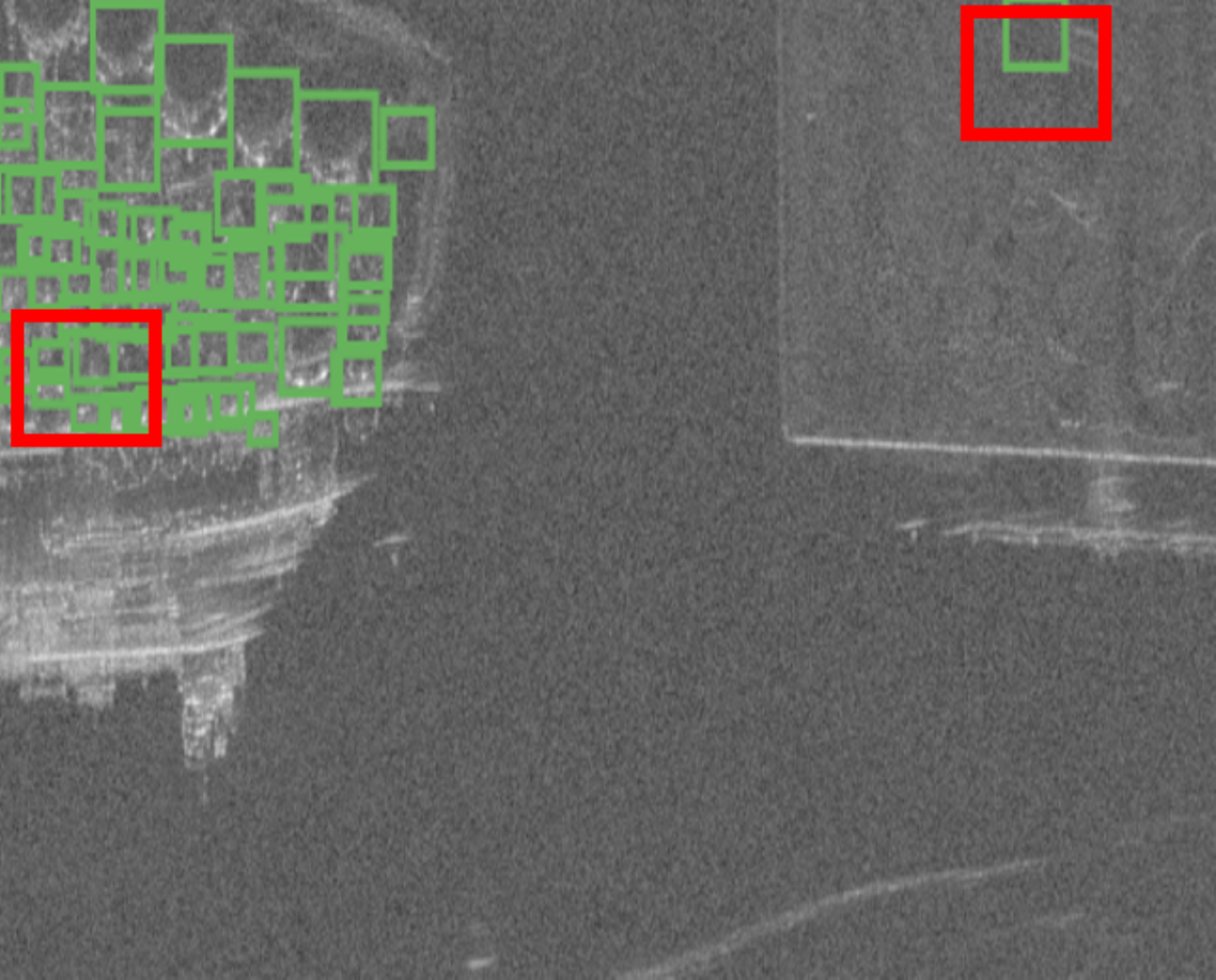} & 
        \includegraphics[width=\linewidth]{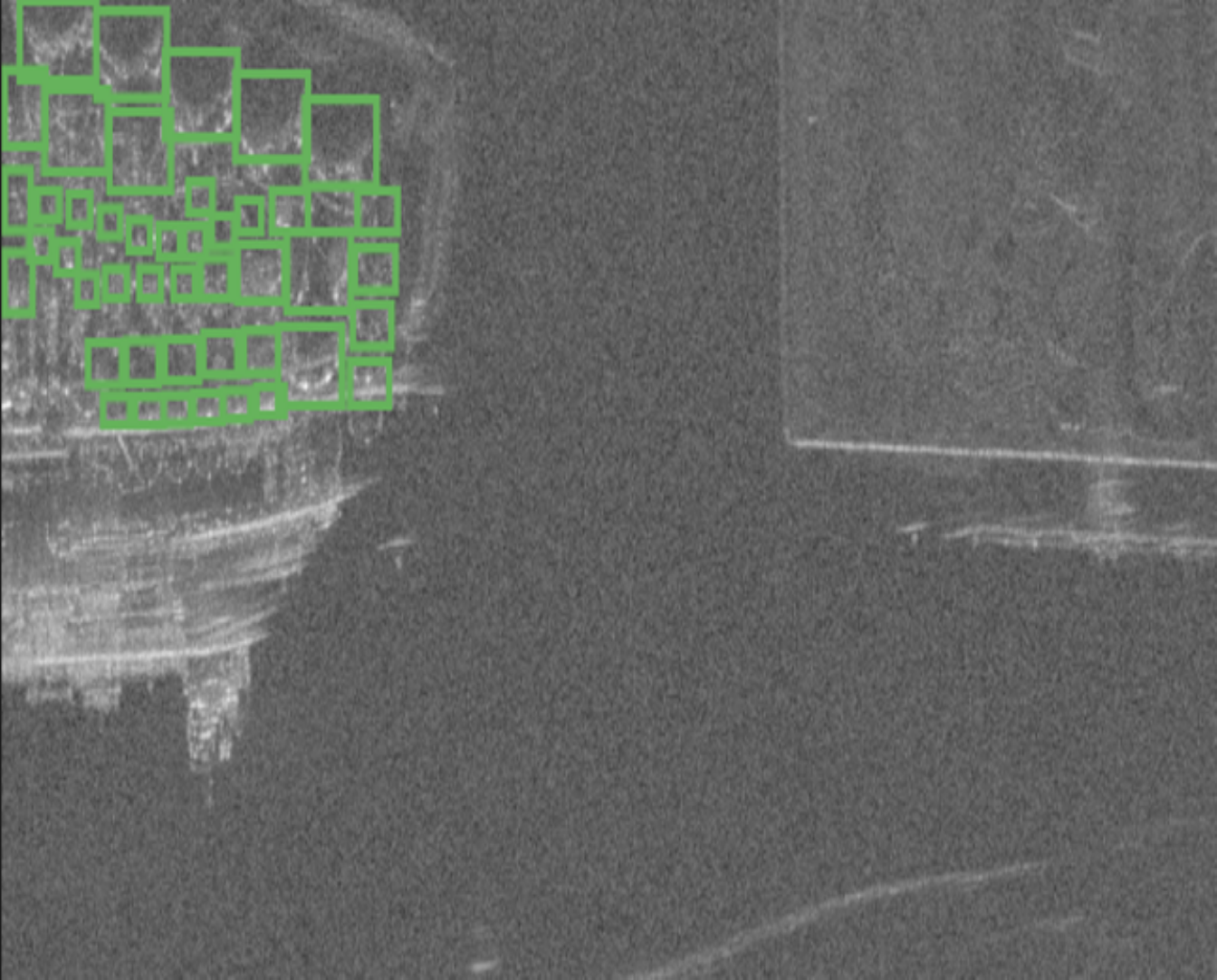} \\

        \includegraphics[width=\linewidth]{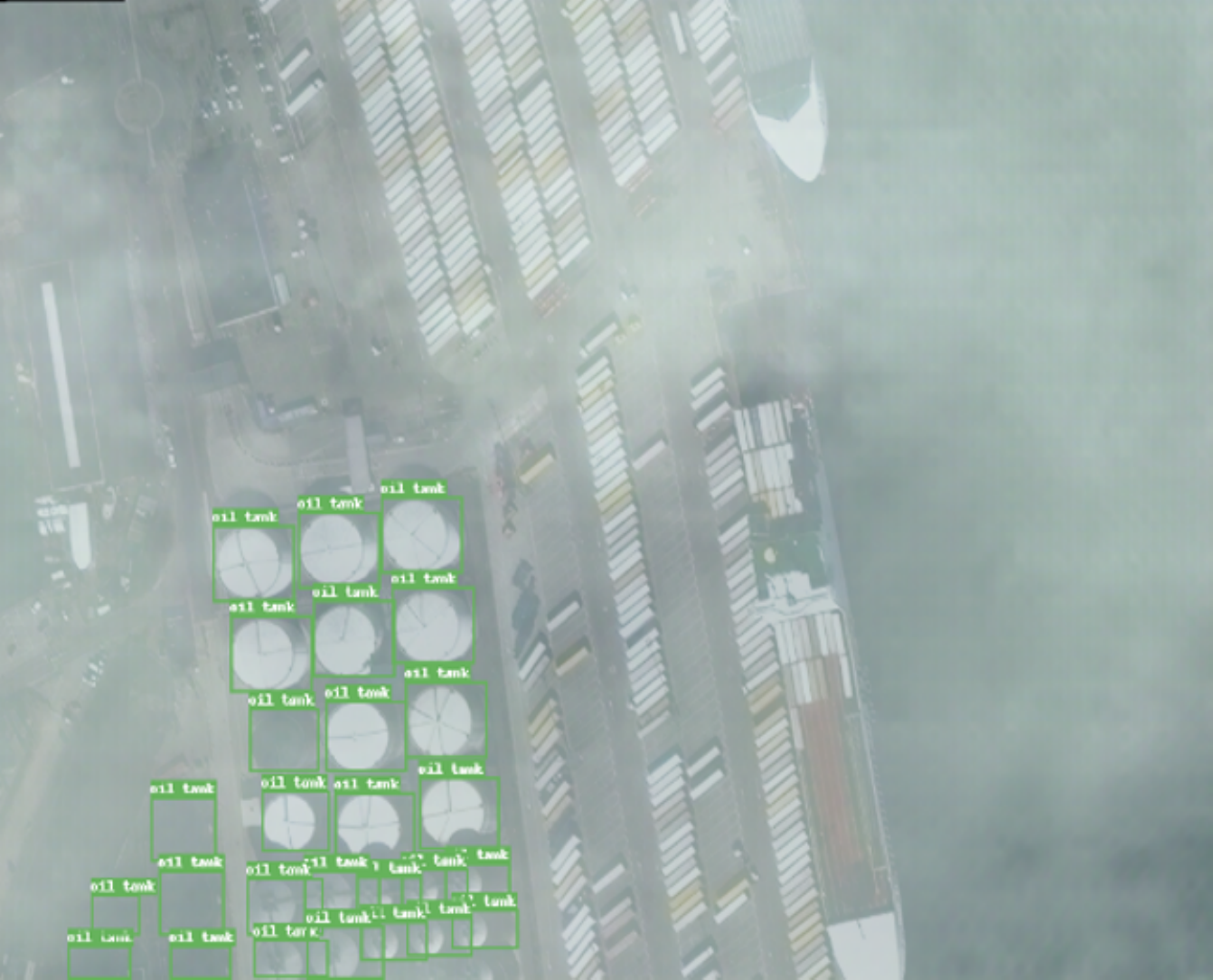} & 
        \includegraphics[width=\linewidth]{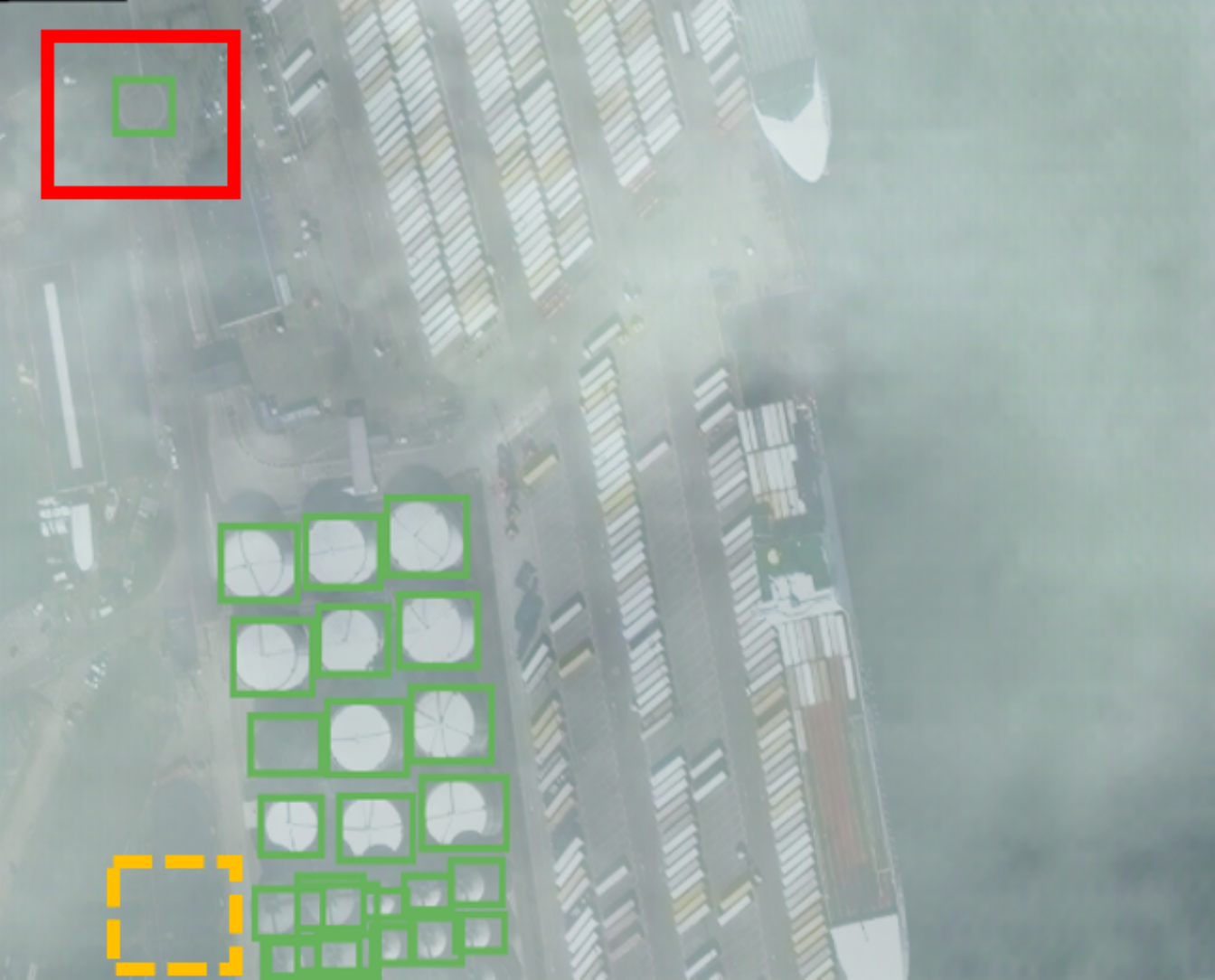} & 
        \includegraphics[width=\linewidth]{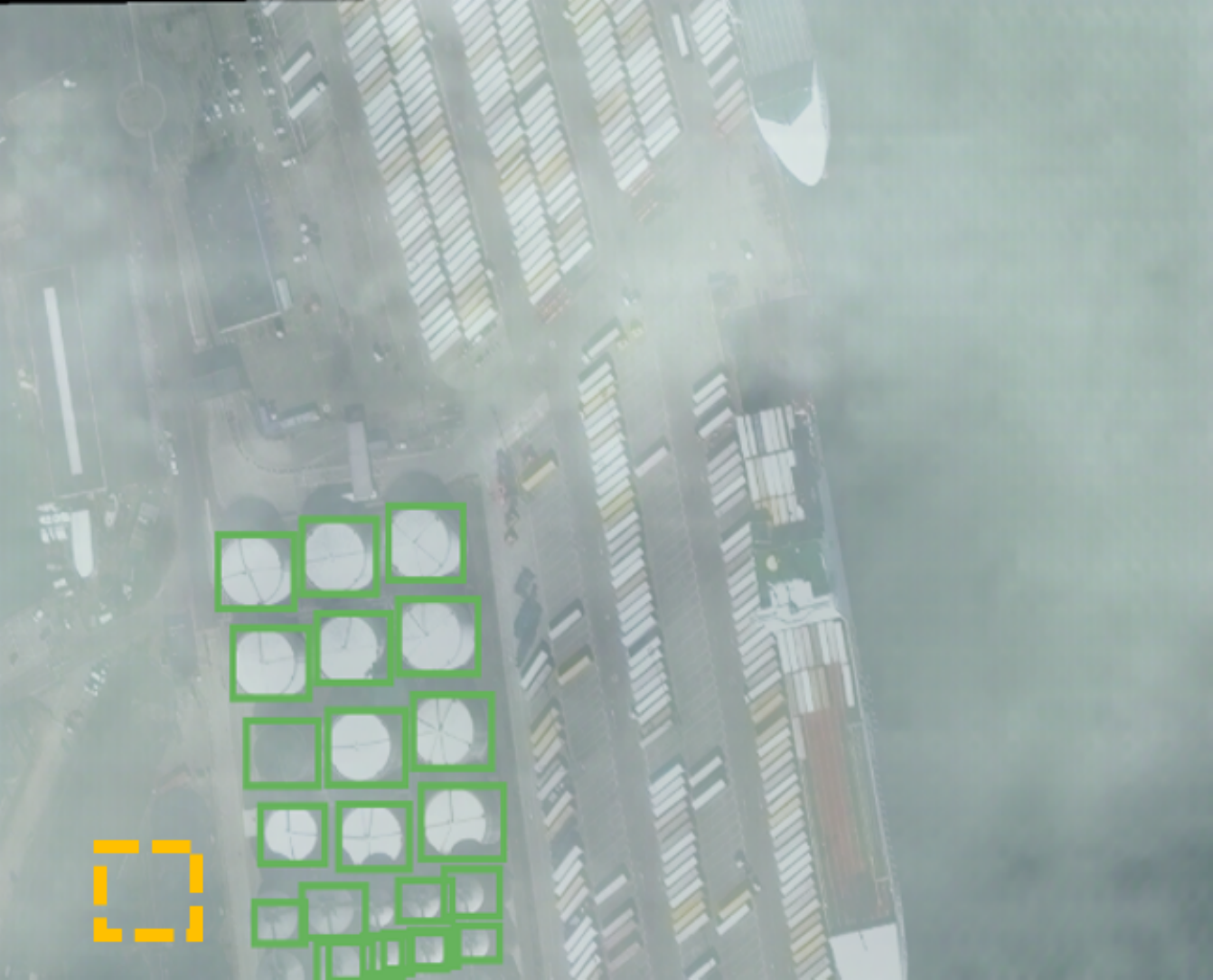} & 
        \includegraphics[width=\linewidth]{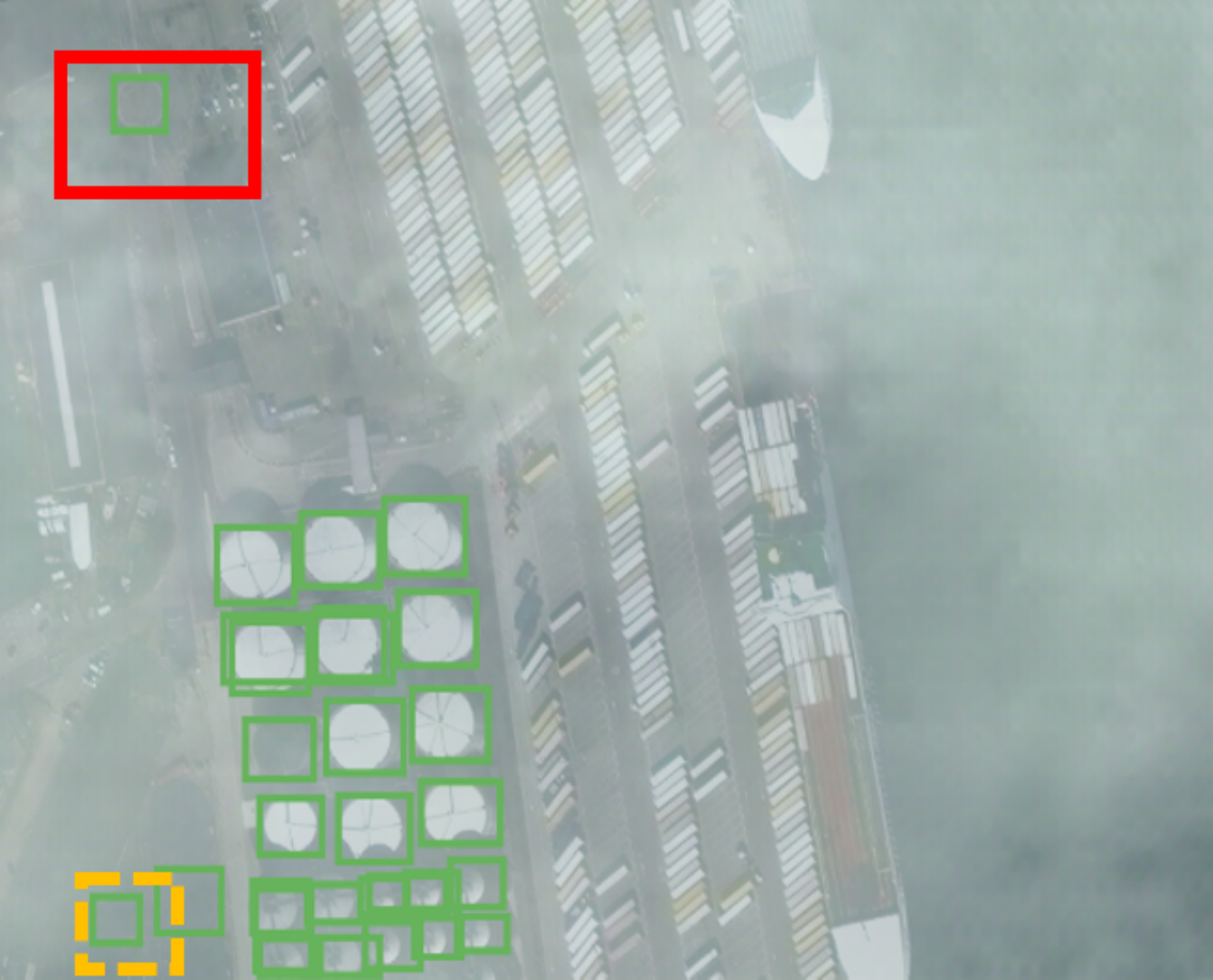} & 
        \includegraphics[width=\linewidth]{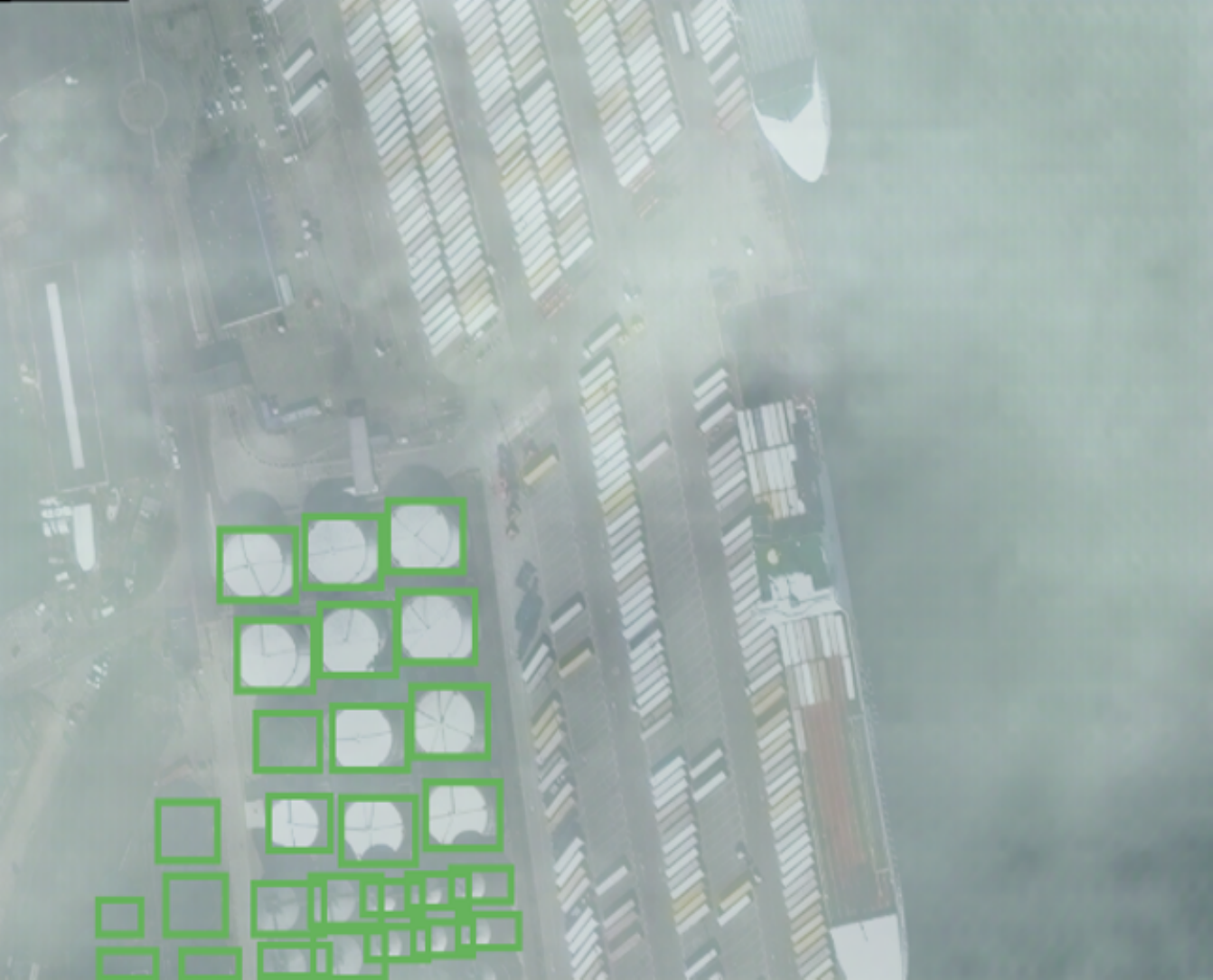} \\

        \includegraphics[width=\linewidth]{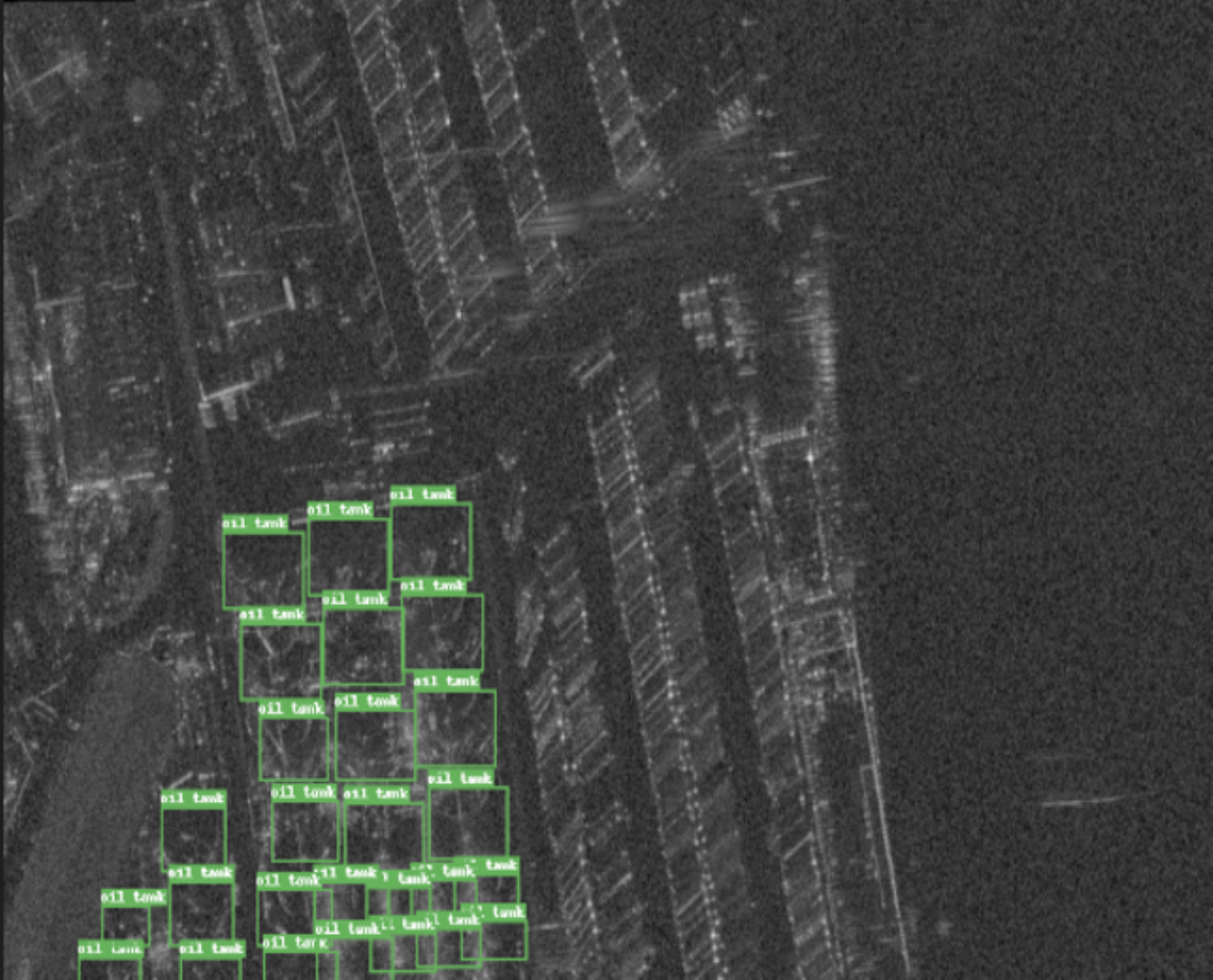} & 
        \includegraphics[width=\linewidth]{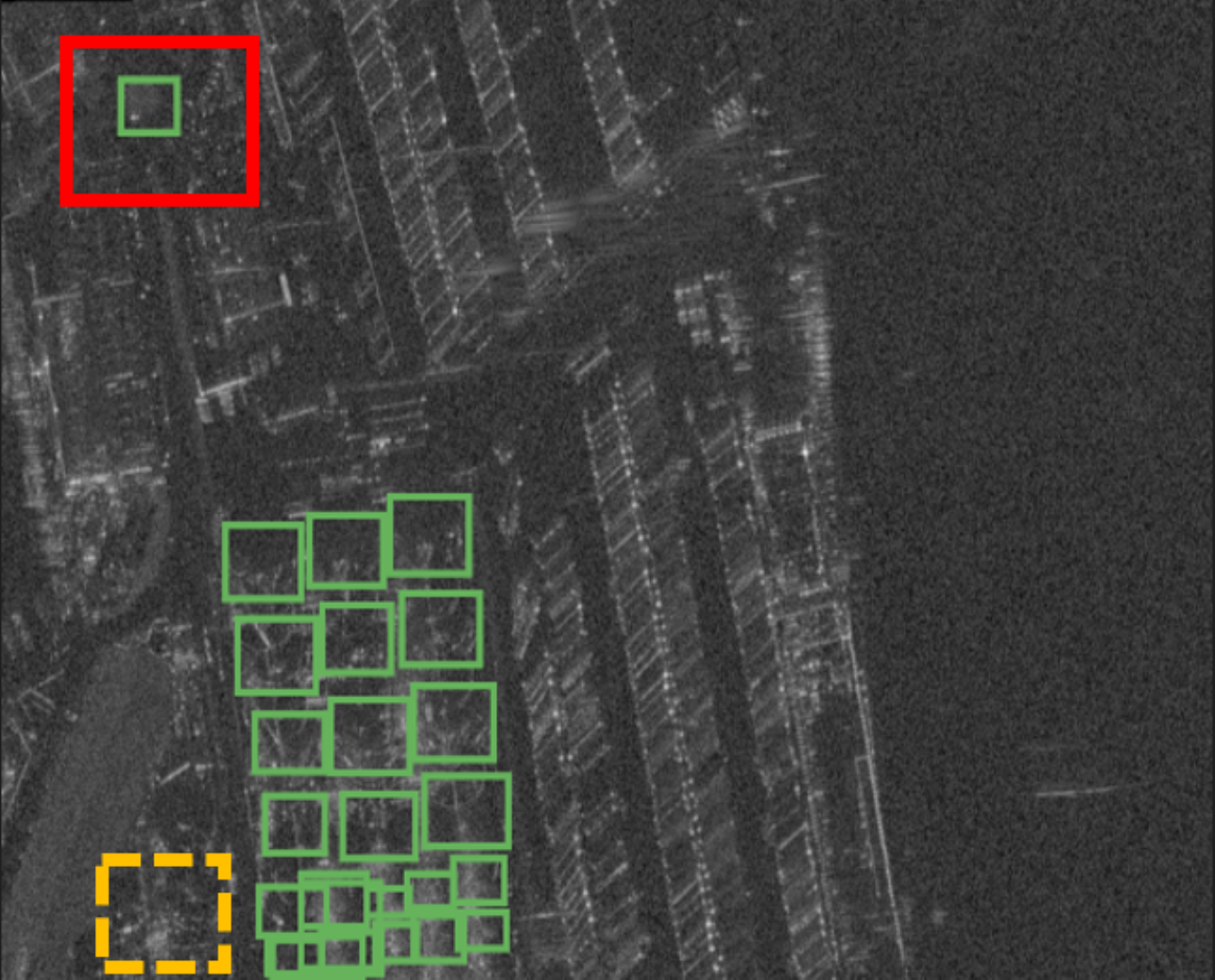} & 
        \includegraphics[width=\linewidth]{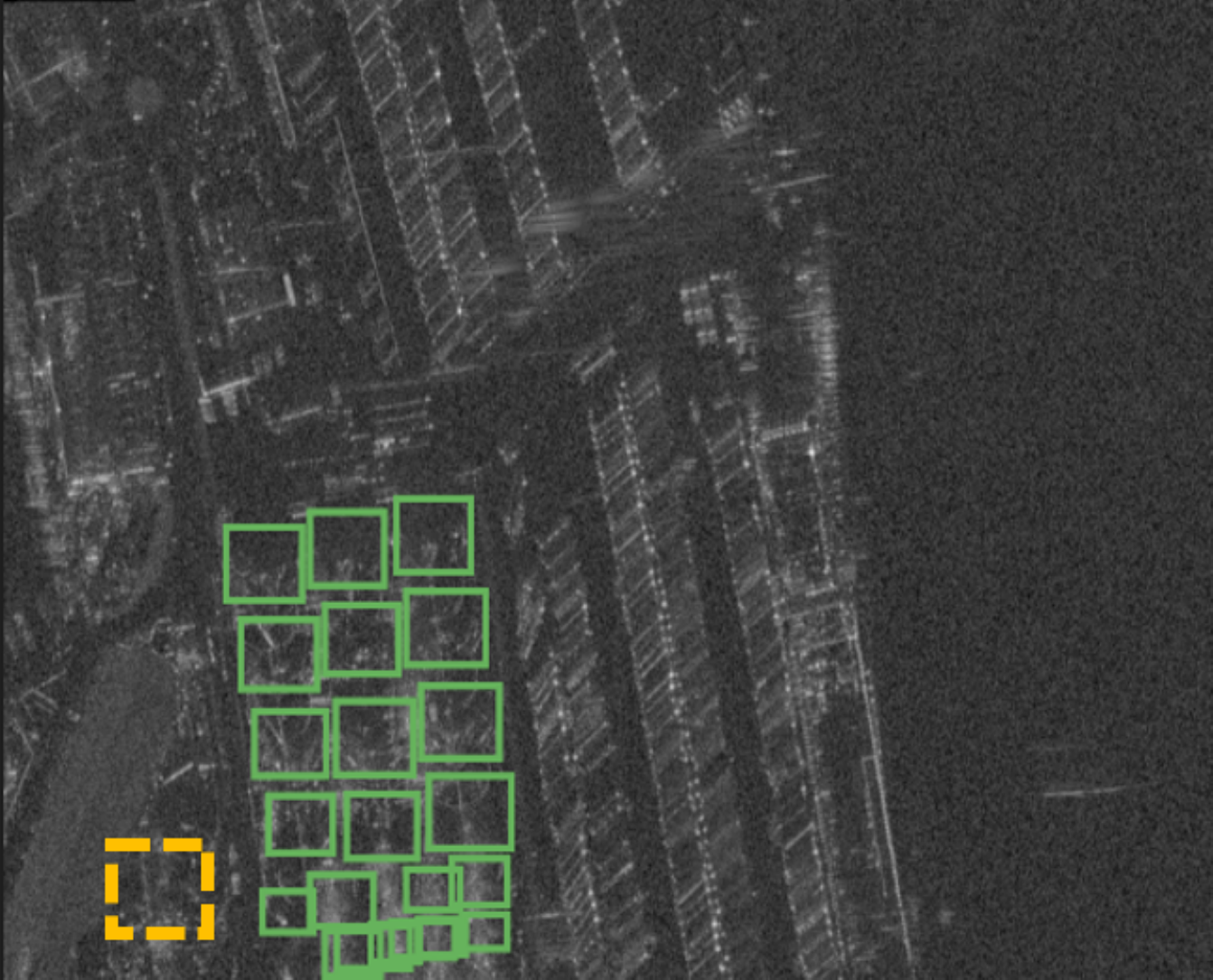} & 
        \includegraphics[width=\linewidth]{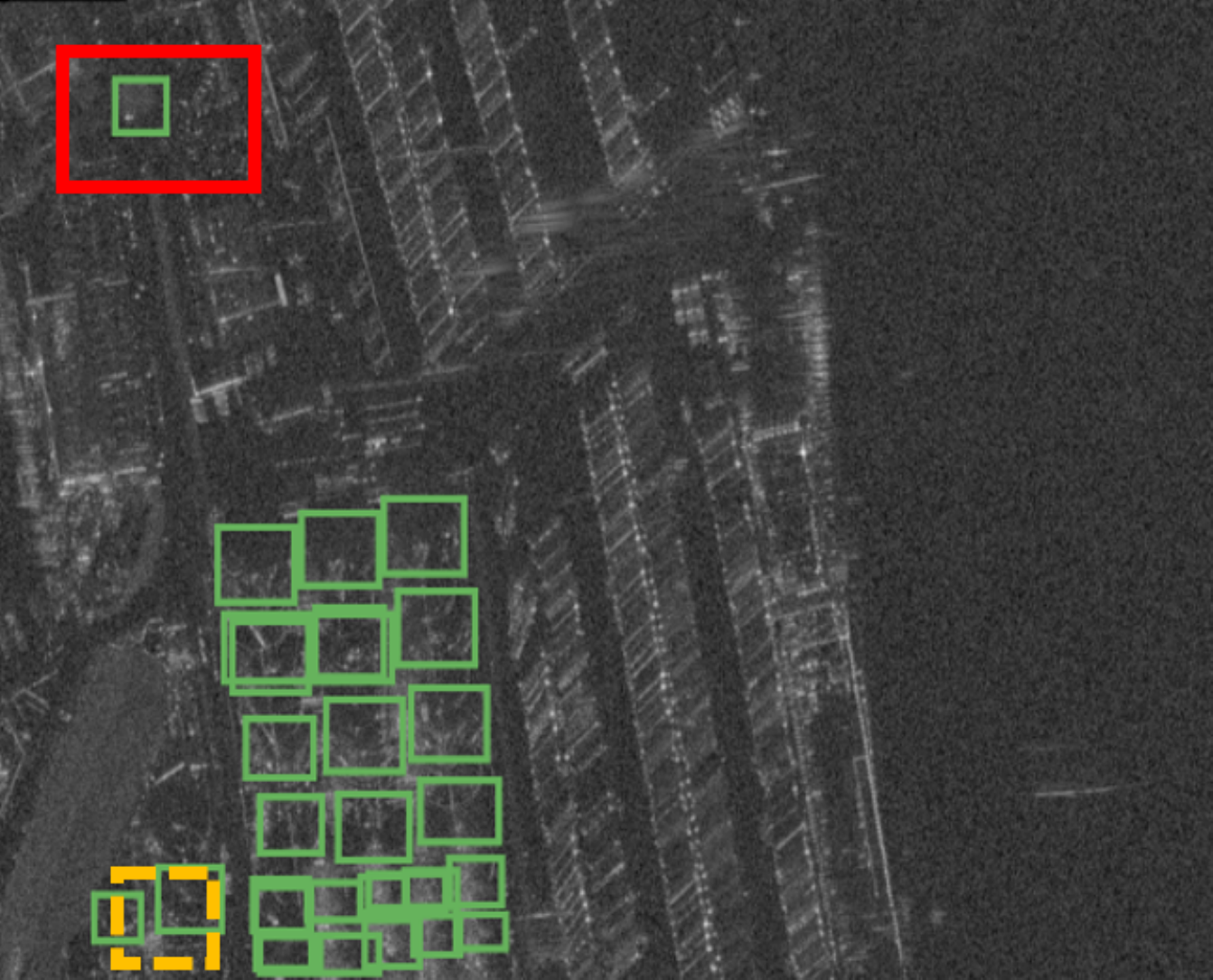} & 
        \includegraphics[width=\linewidth]{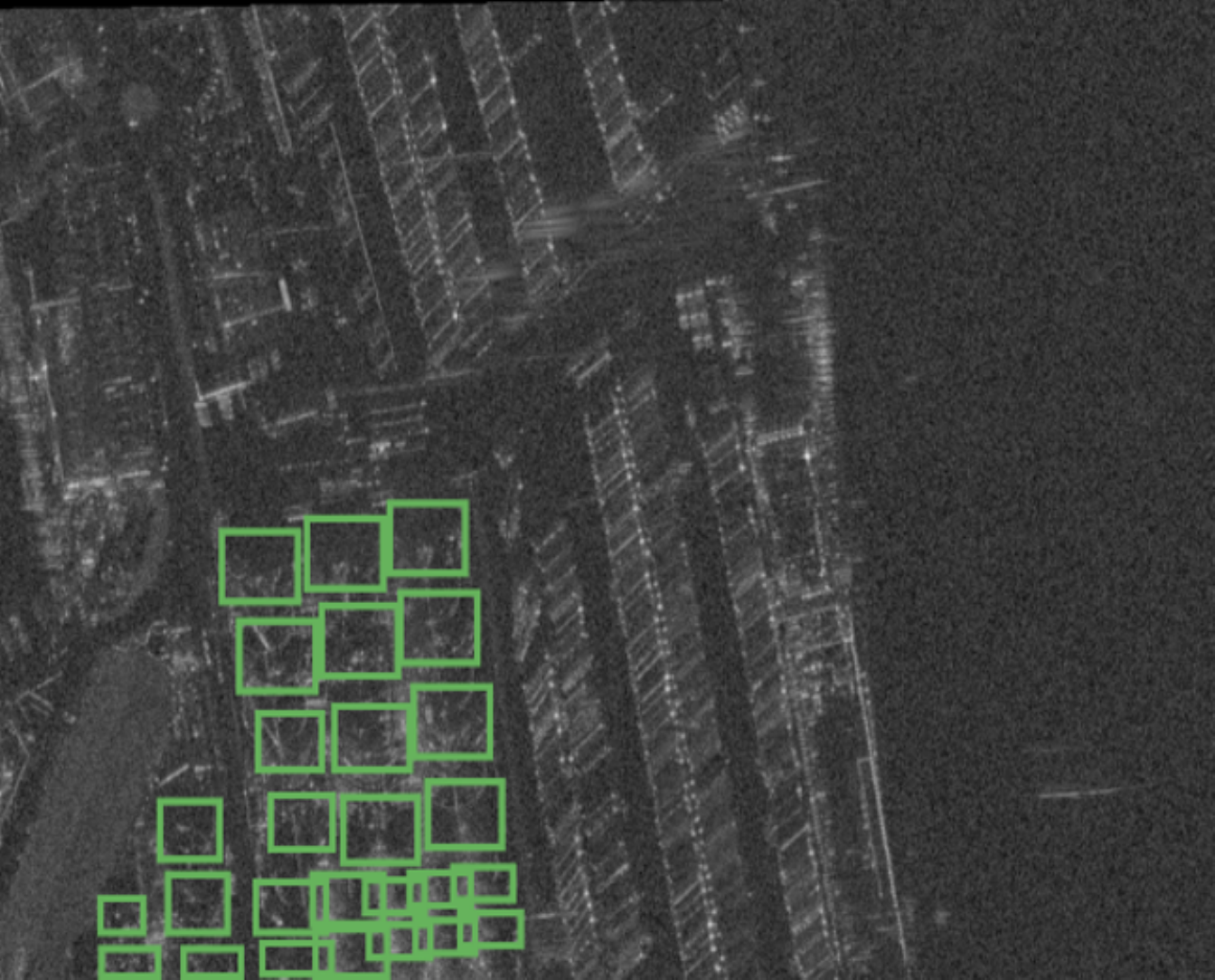} \\

        (a) & (b) & (c) & (d) & (e) \\
    \end{tabular}
    \caption{Illustration of output visualization on the SpaceNet6-OTD-Fog dataset. (a) Ground truth labels. (b) CFT. (c) ICAFusion. (d) MMIDet. (e) Ours. Yellow dashed boxes indicate objects missed during localization, red boxes represent false detections, and green boxes denote visualization boxes with correct predictions.}
    \label{fig:11}
\end{figure*}

\begin{figure*}[htbp]
    \centering
    \begin{tabular}{>{\centering\arraybackslash}m{0.18\textwidth}
                    >{\centering\arraybackslash}m{0.18\textwidth}
                    >{\centering\arraybackslash}m{0.18\textwidth}
                    >{\centering\arraybackslash}m{0.18\textwidth}
                    >{\centering\arraybackslash}m{0.18\textwidth}}
        \includegraphics[width=\linewidth]{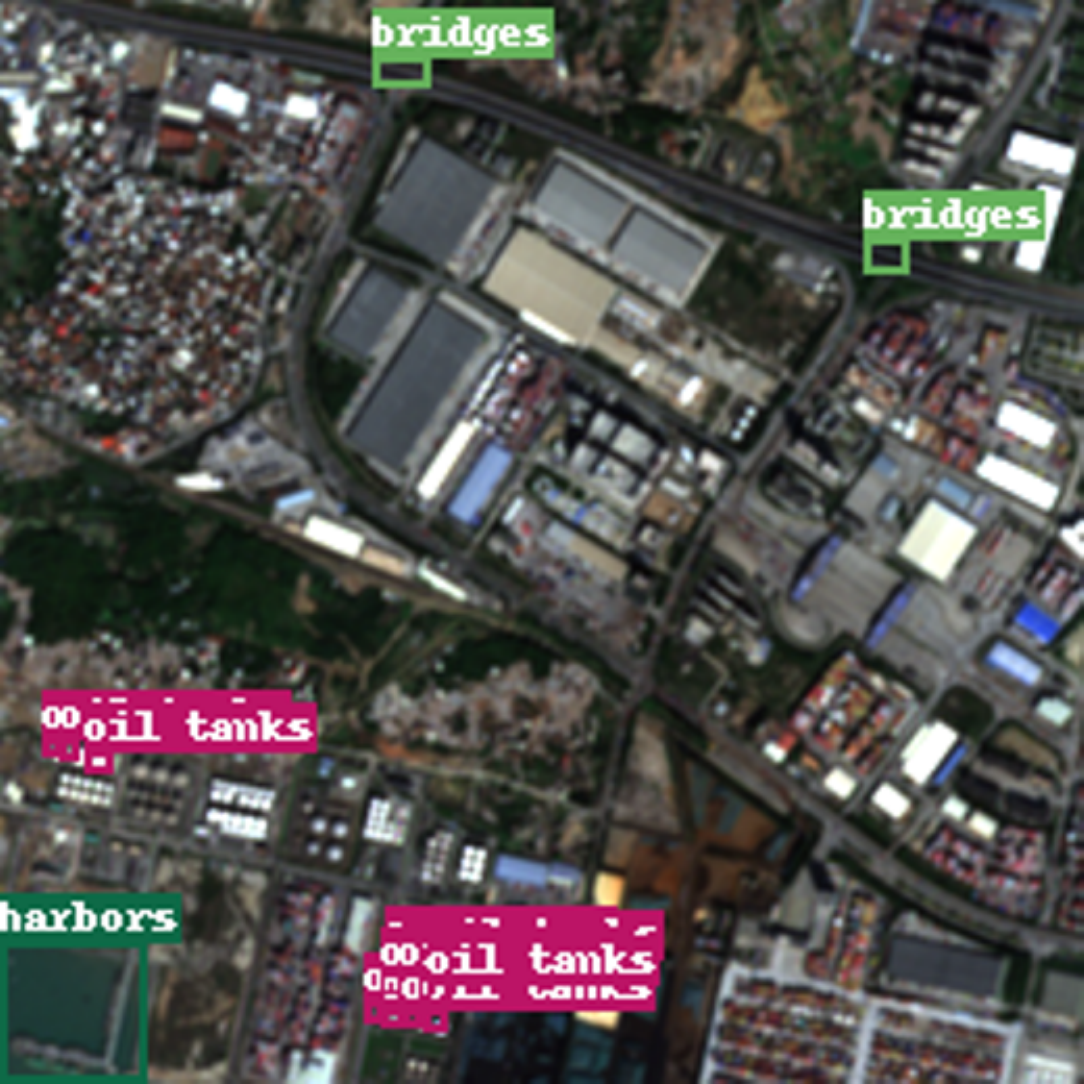} & 
        \includegraphics[width=\linewidth]{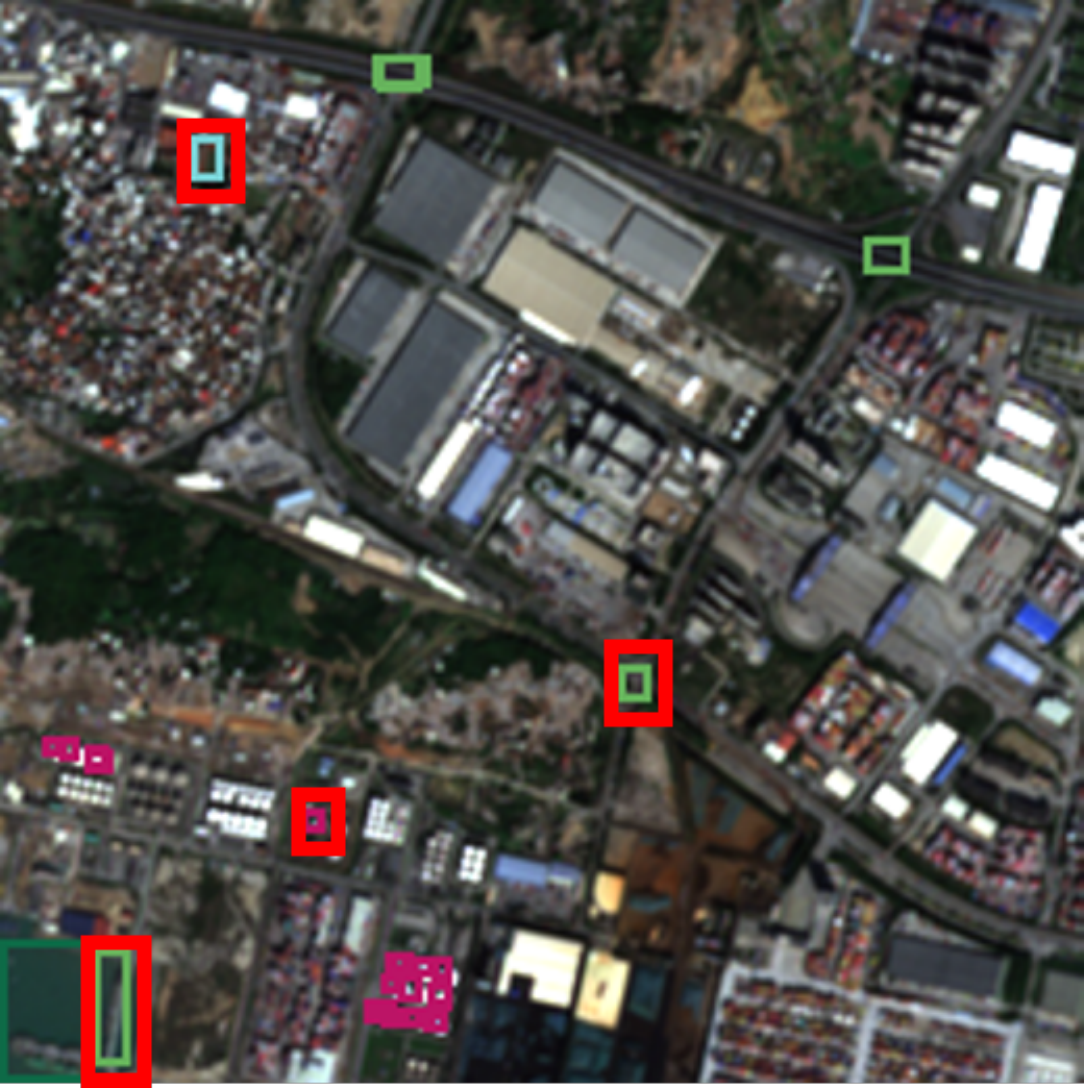} & 
        \includegraphics[width=\linewidth]{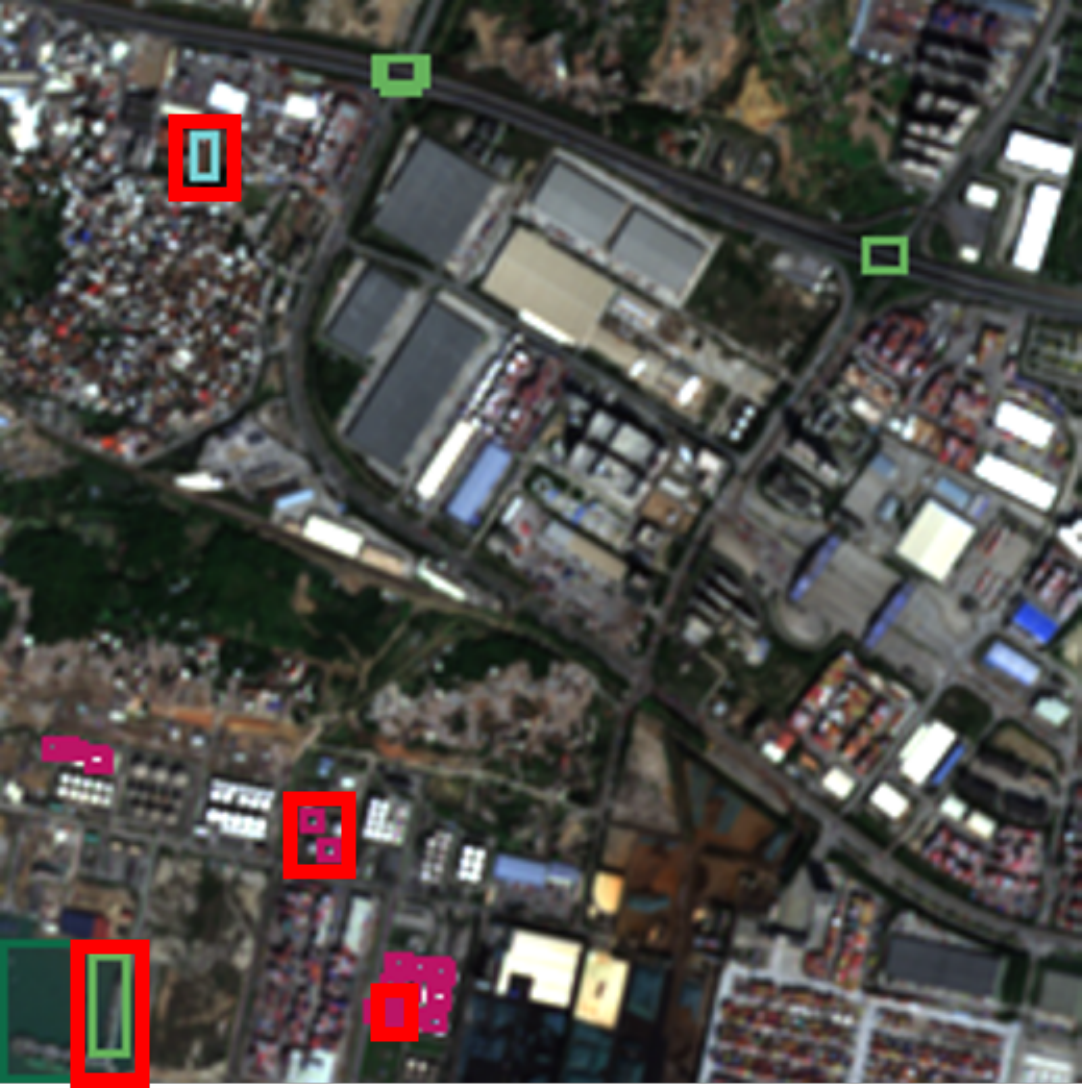} & 
        \includegraphics[width=\linewidth]{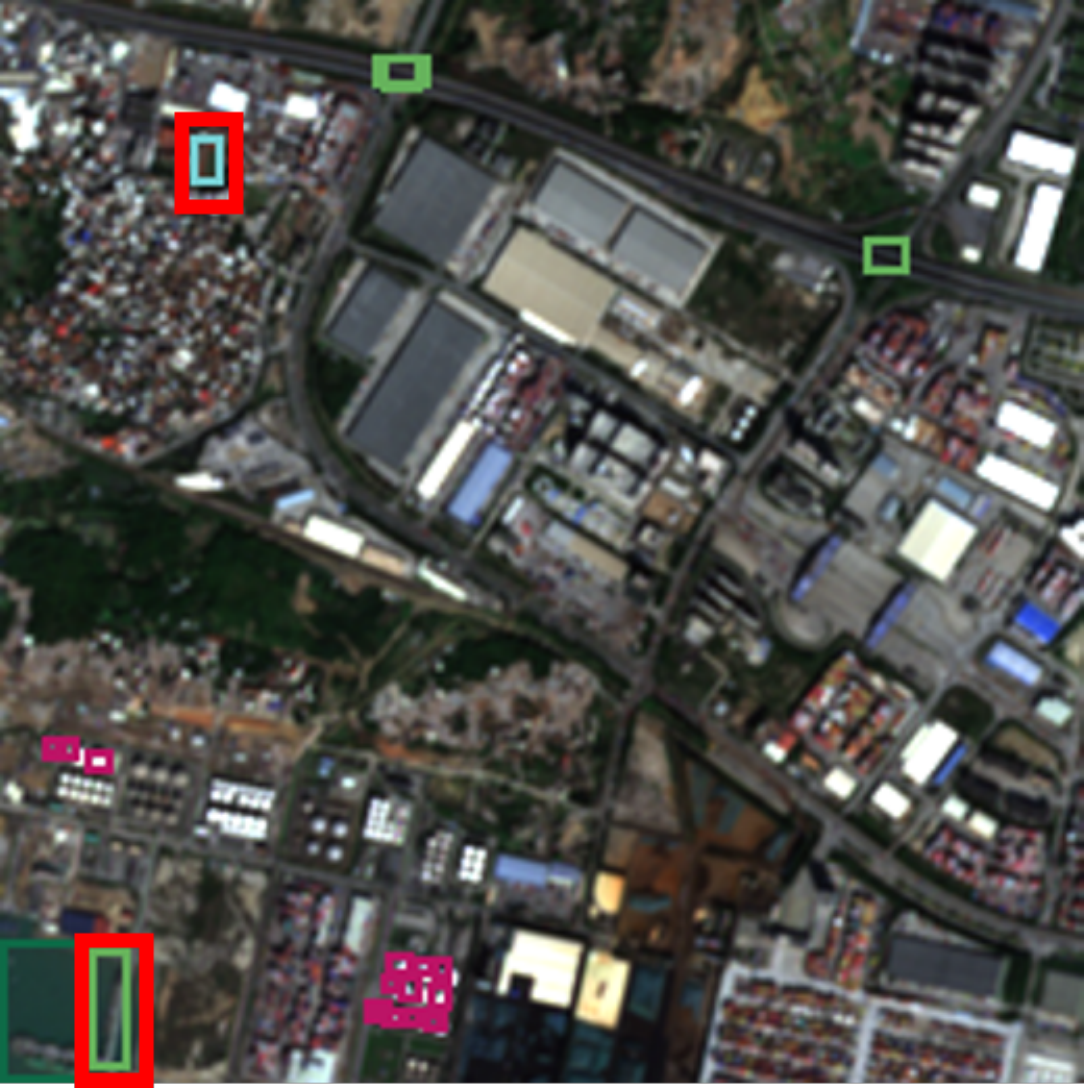} & 
        \includegraphics[width=\linewidth]{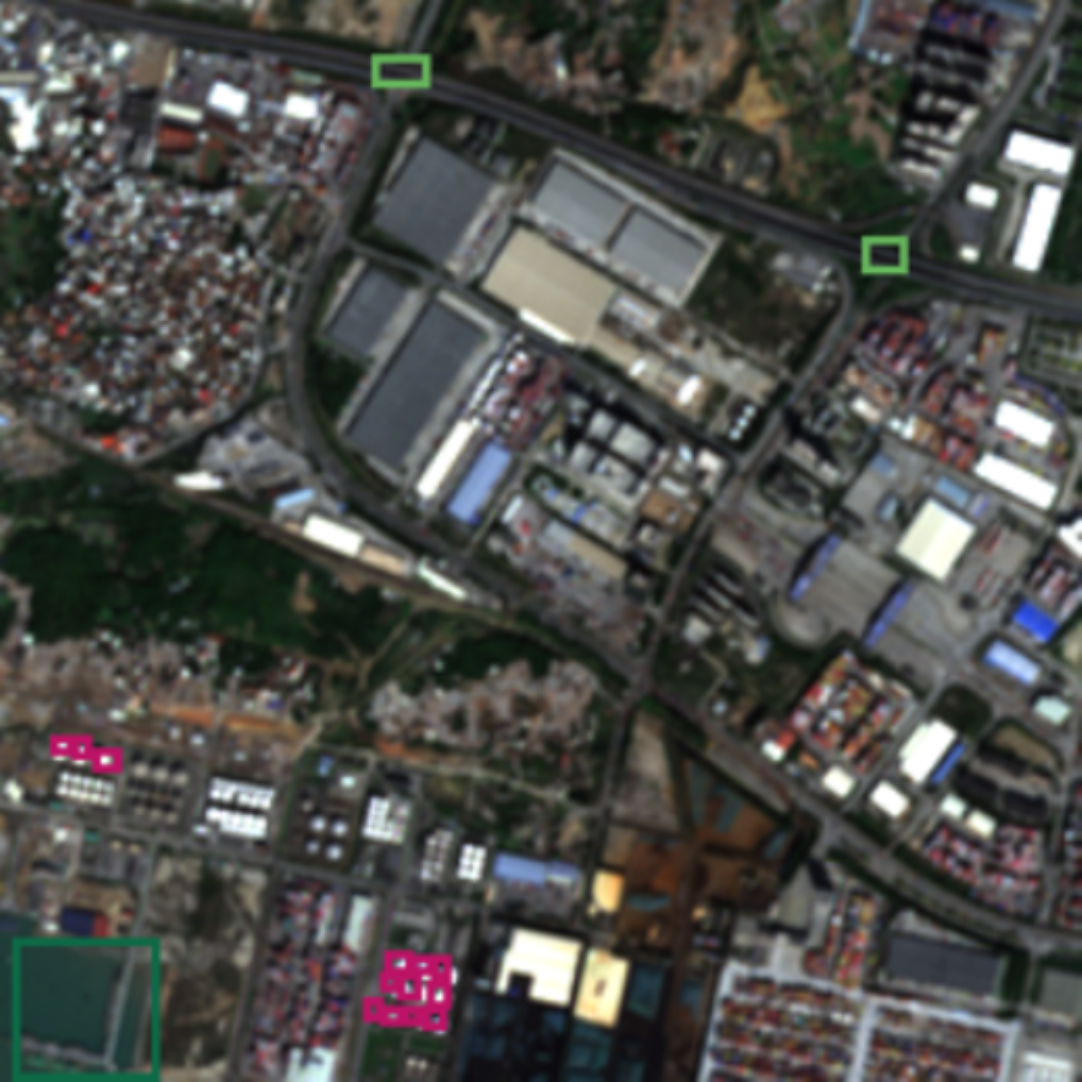} \\
        
        \includegraphics[width=\linewidth]{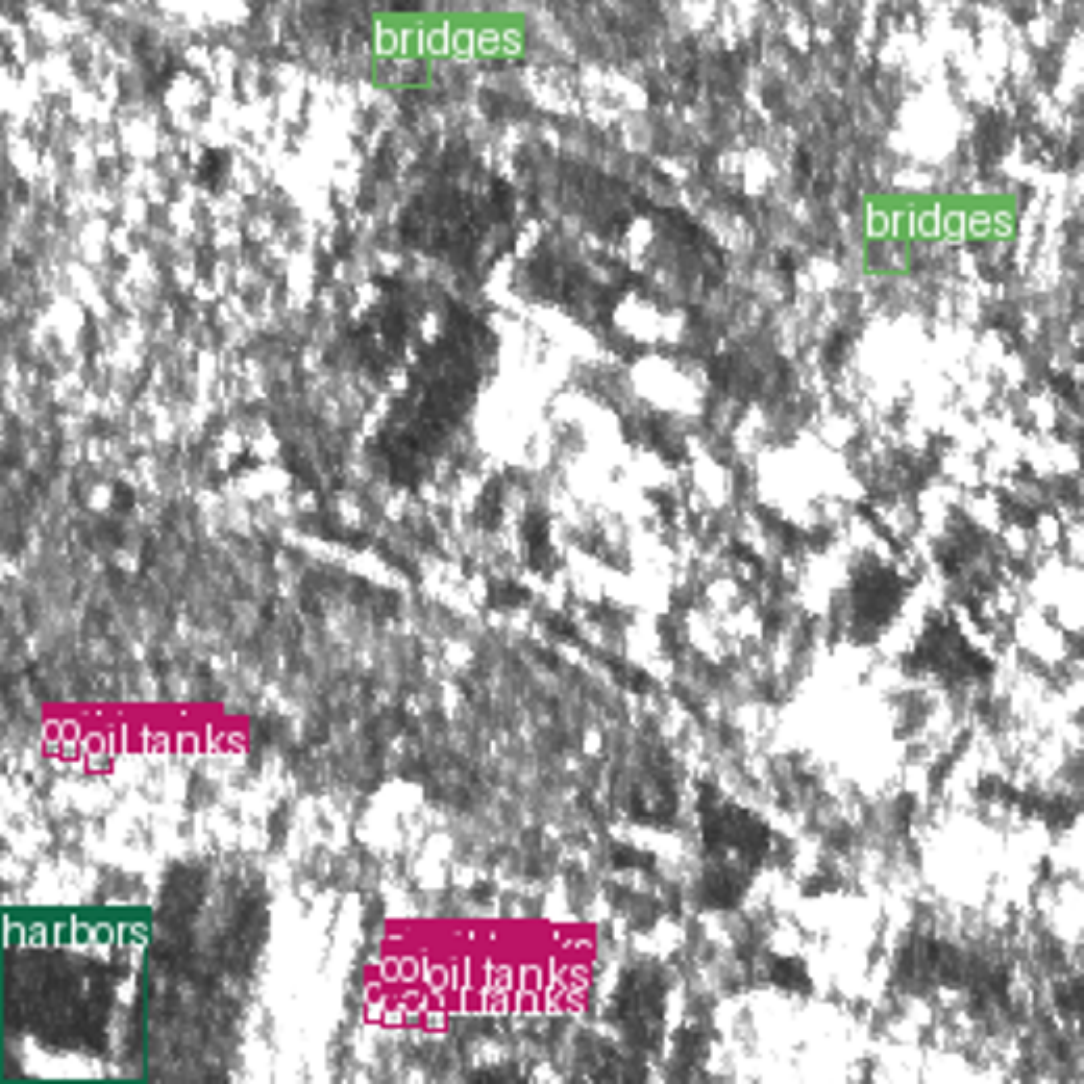} & 
        \includegraphics[width=\linewidth]{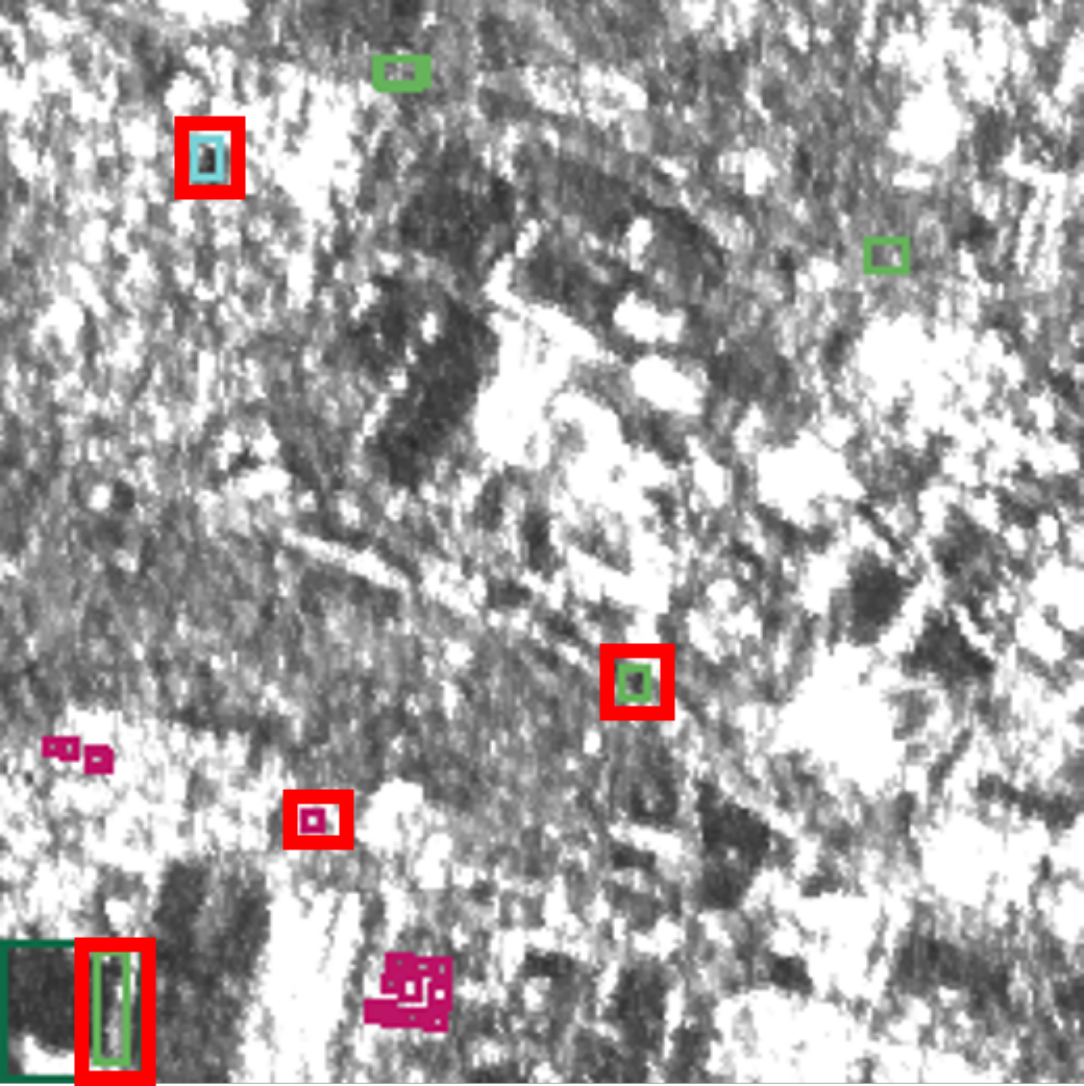} & 
        \includegraphics[width=\linewidth]{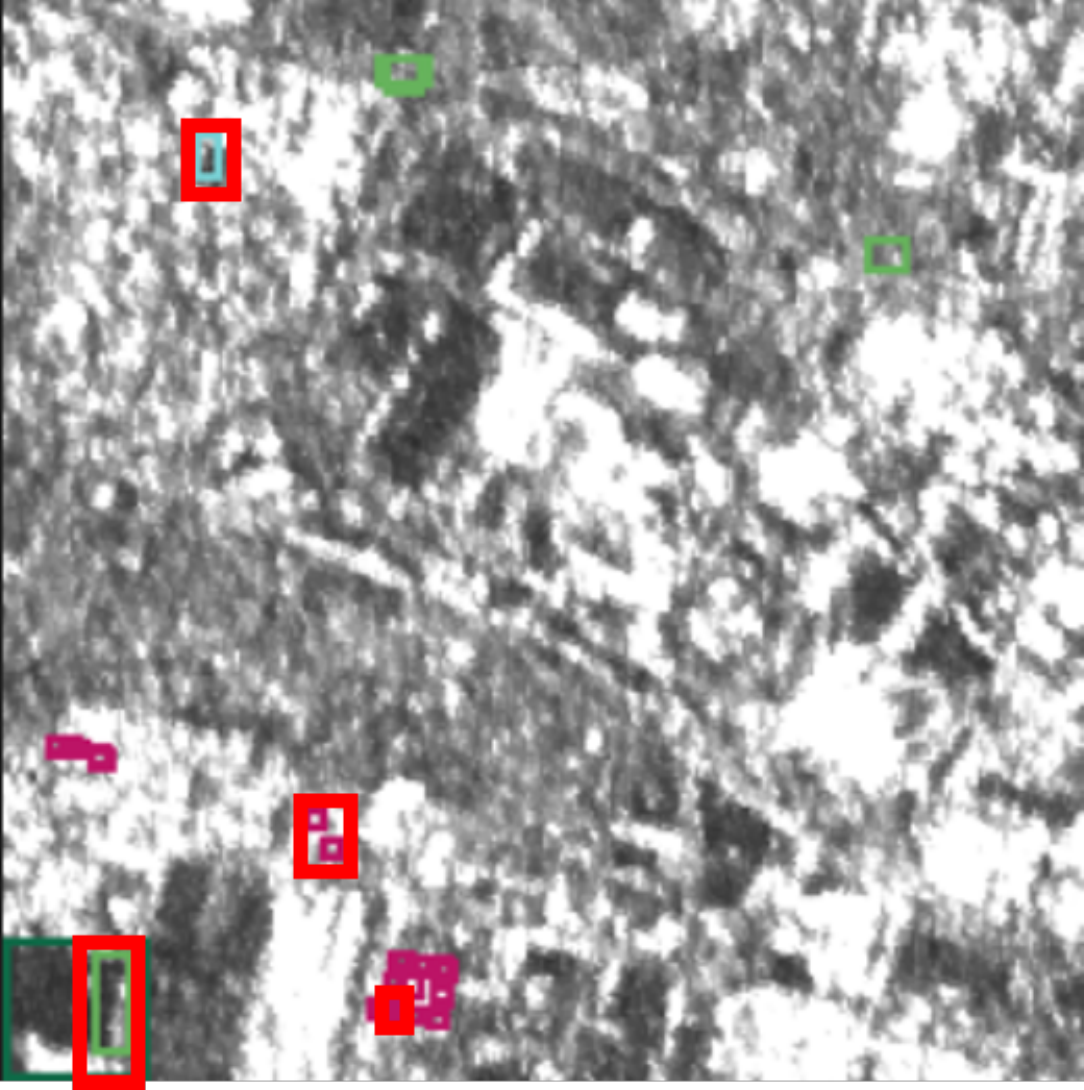} & 
        \includegraphics[width=\linewidth]{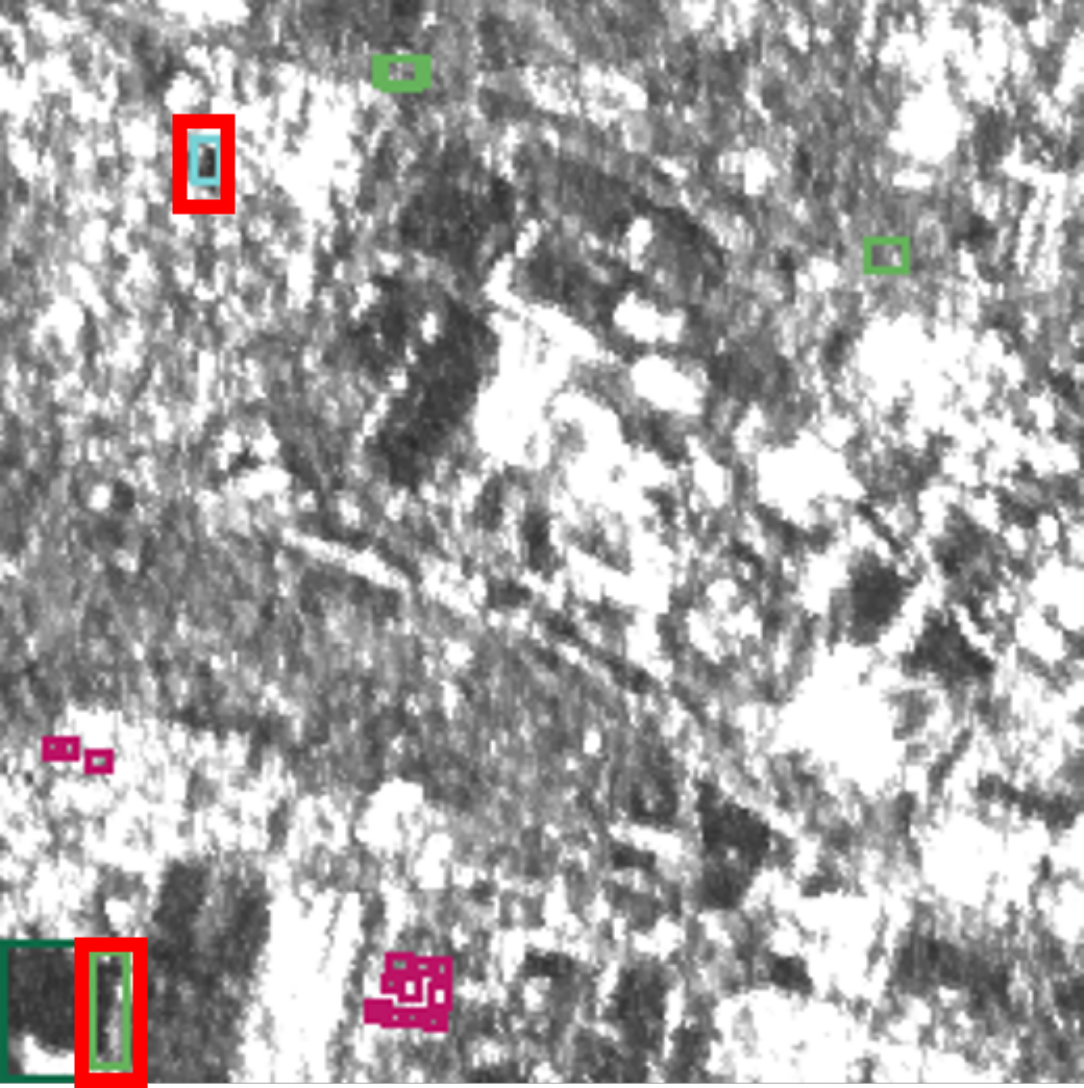} & 
        \includegraphics[width=\linewidth]{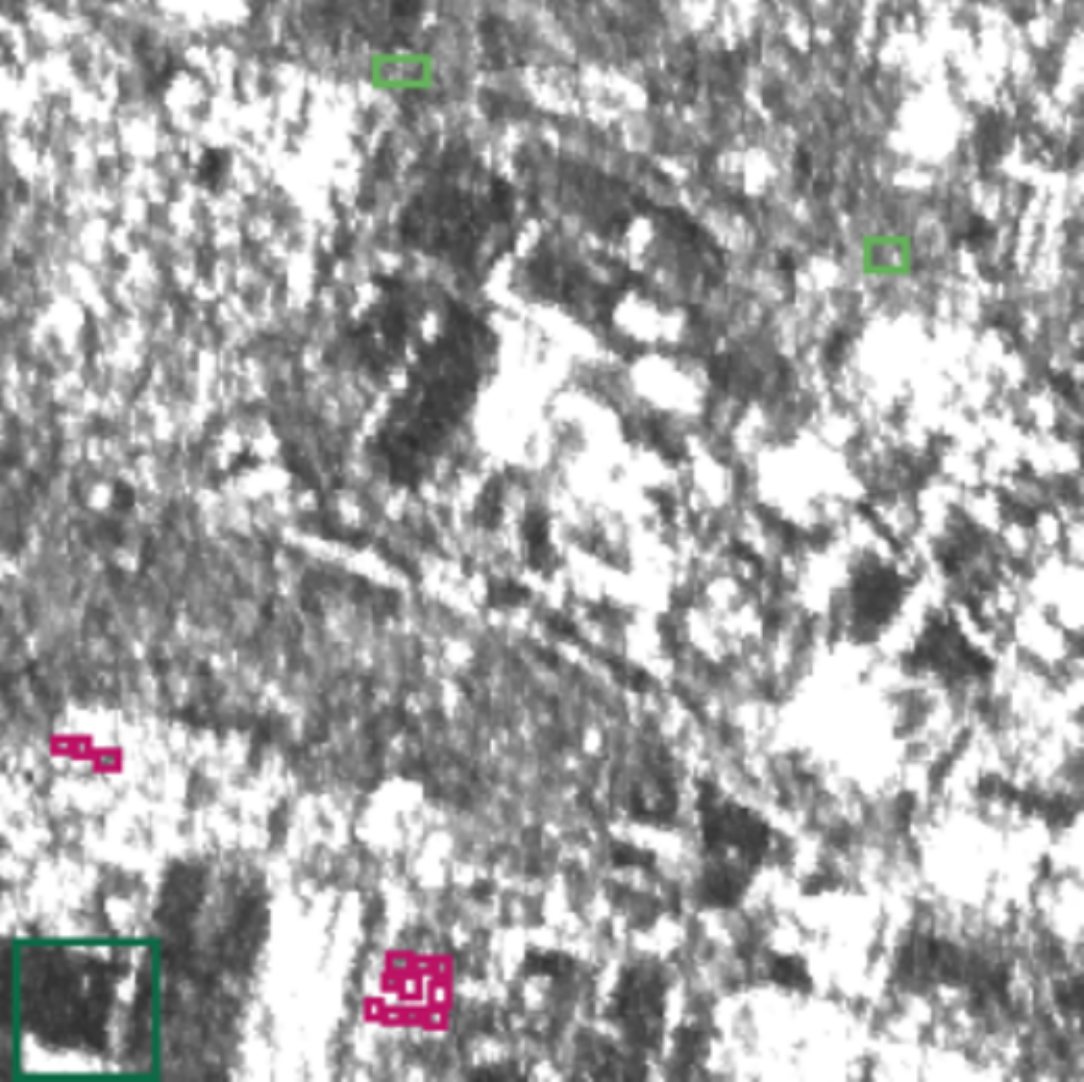} \\

        \includegraphics[width=\linewidth]{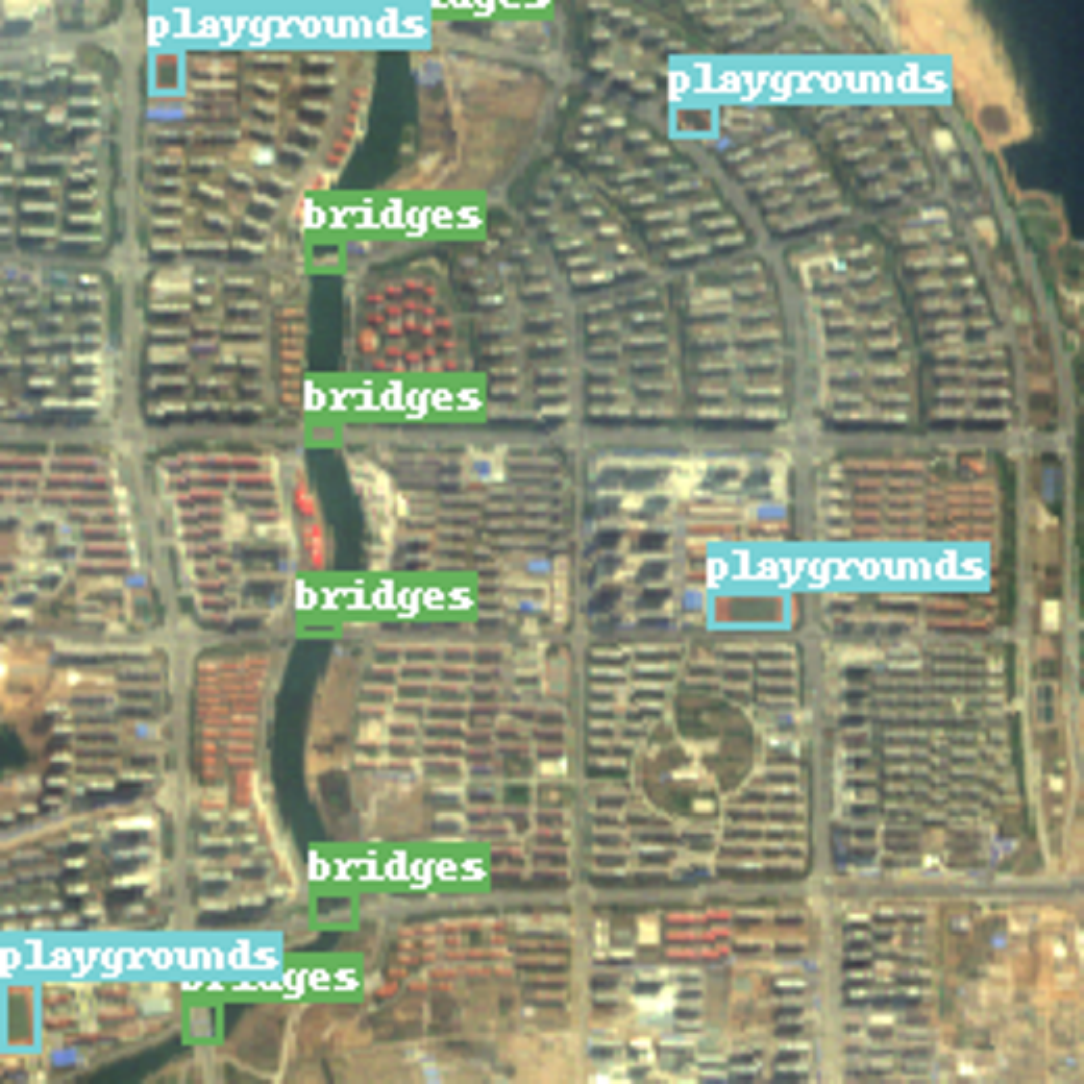} & 
        \includegraphics[width=\linewidth]{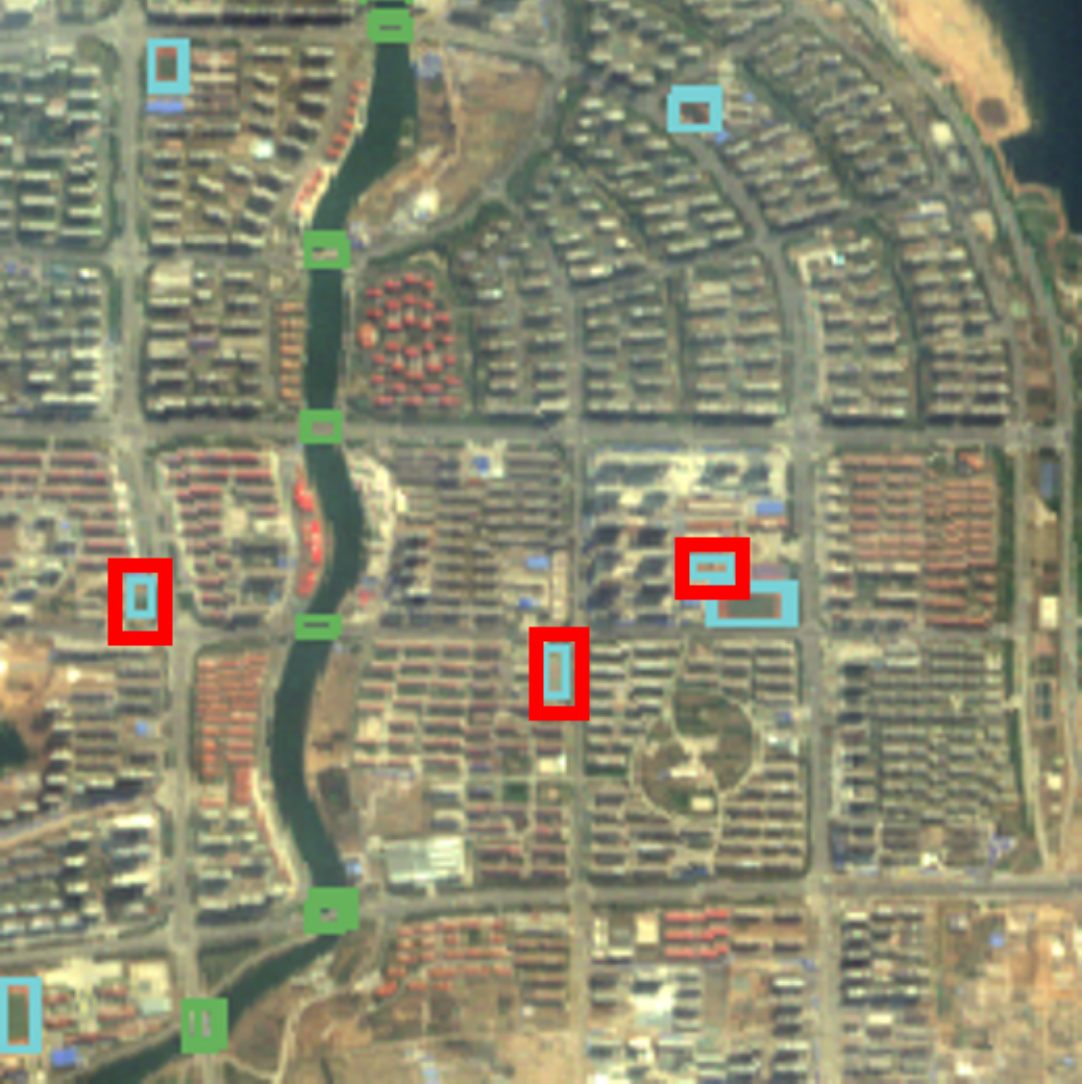} & 
        \includegraphics[width=\linewidth]{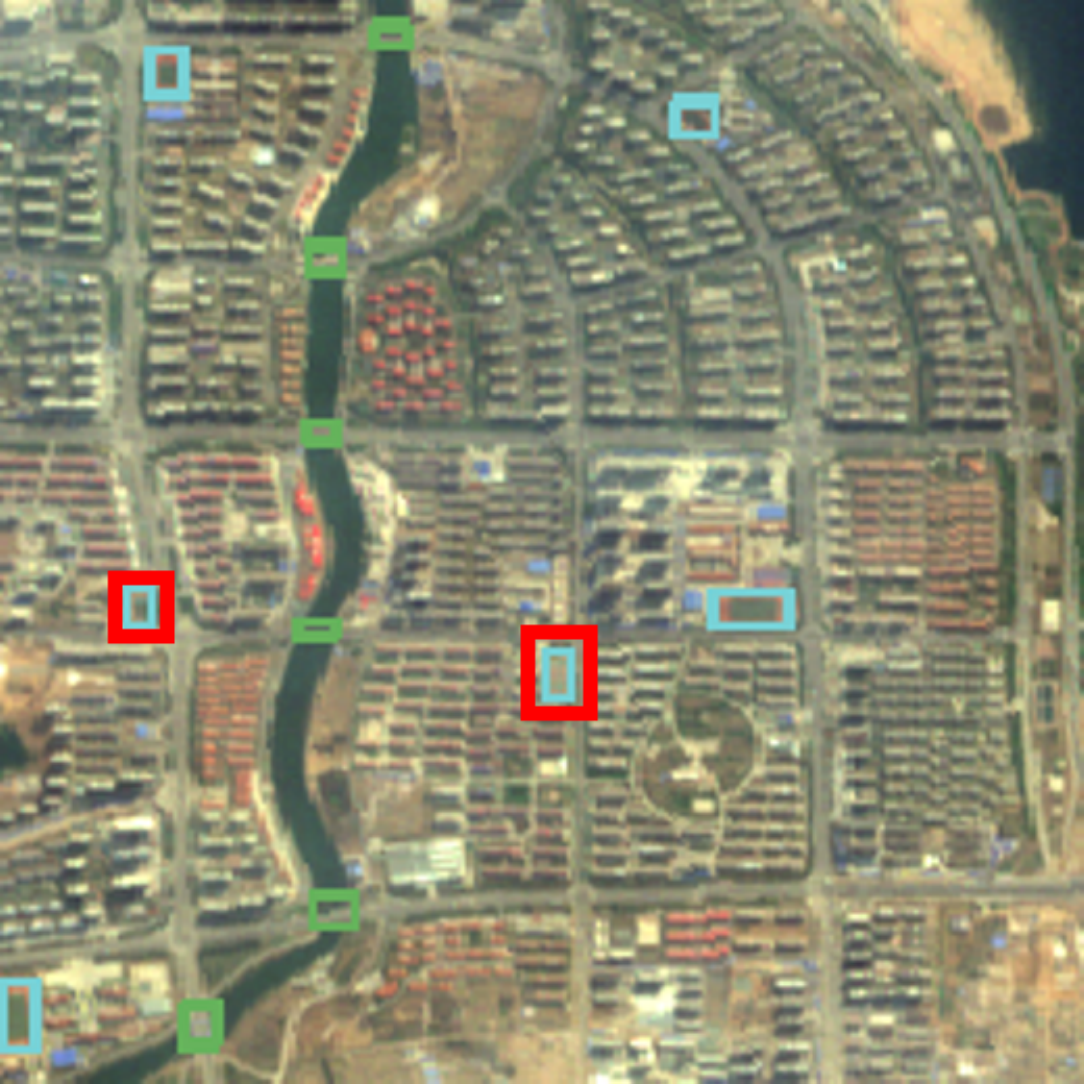} & 
        \includegraphics[width=\linewidth]{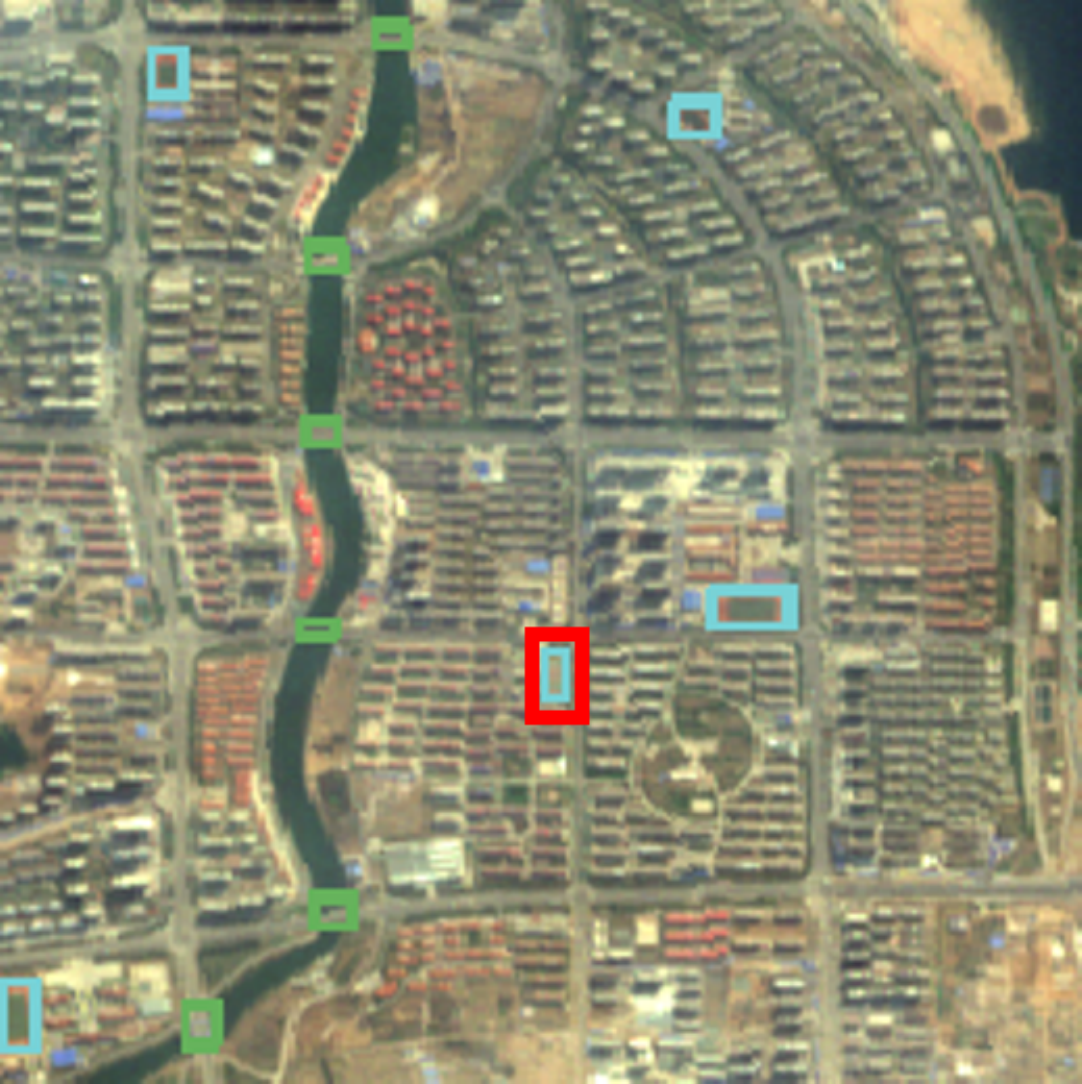} & 
        \includegraphics[width=\linewidth]{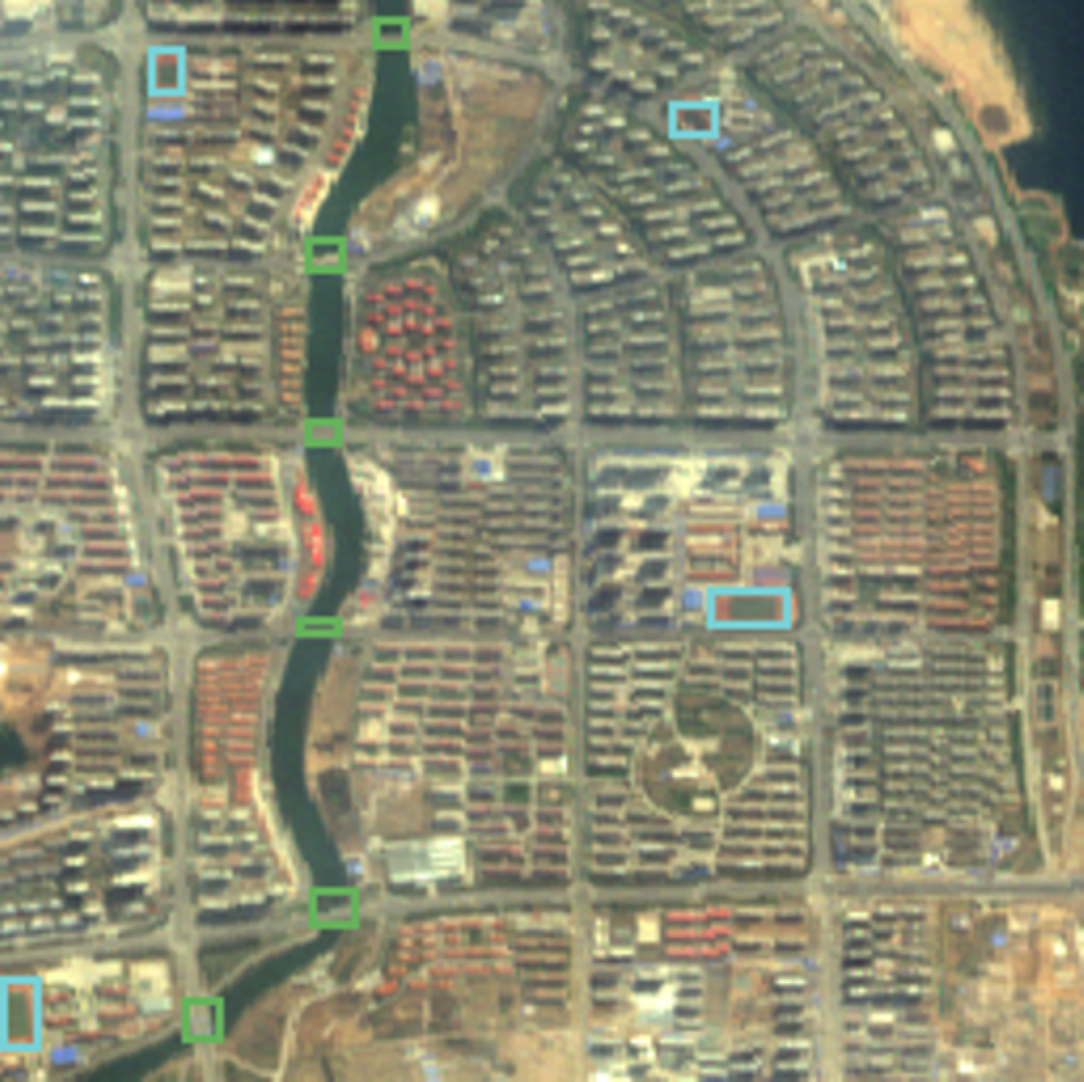} \\

        \includegraphics[width=\linewidth]{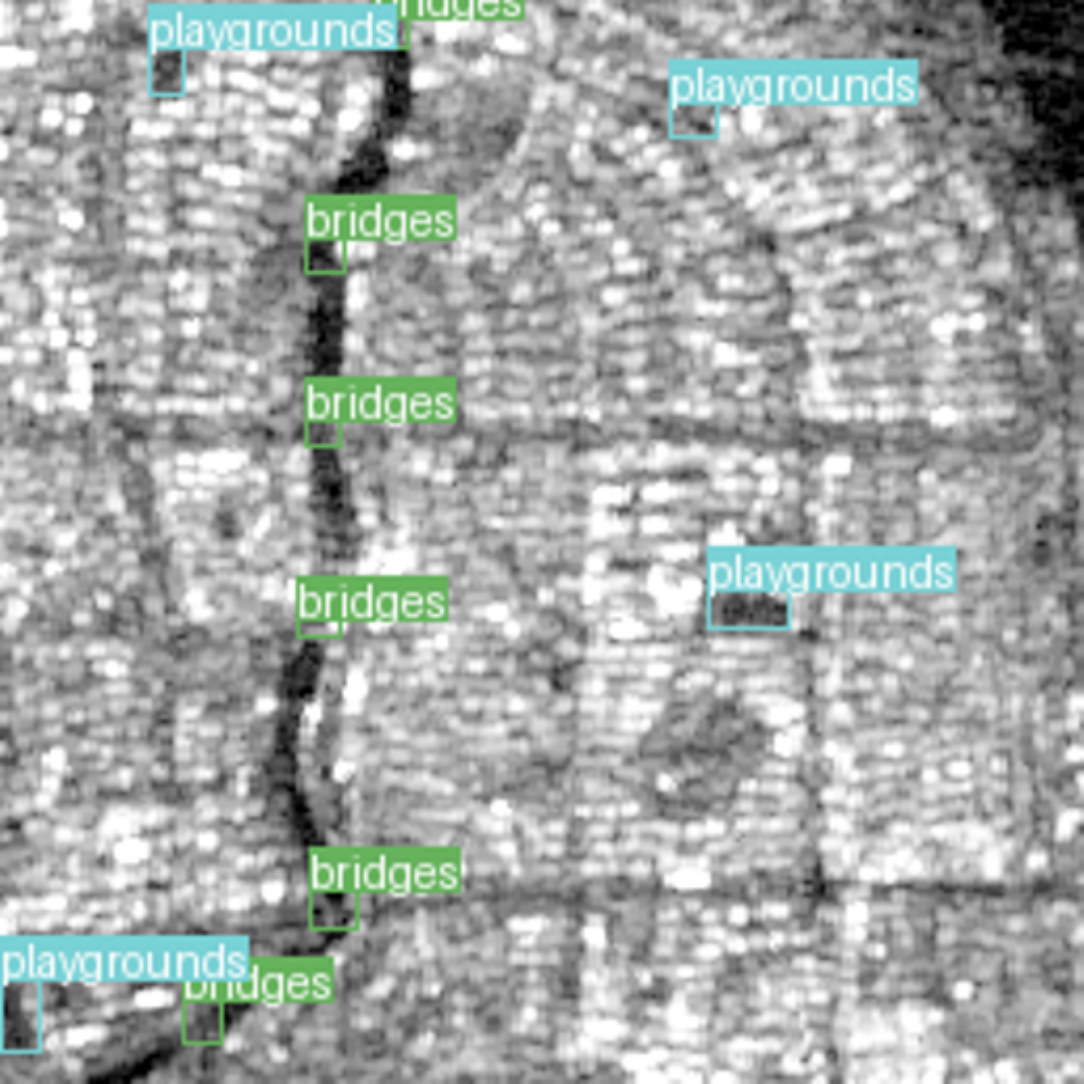} & 
        \includegraphics[width=\linewidth]{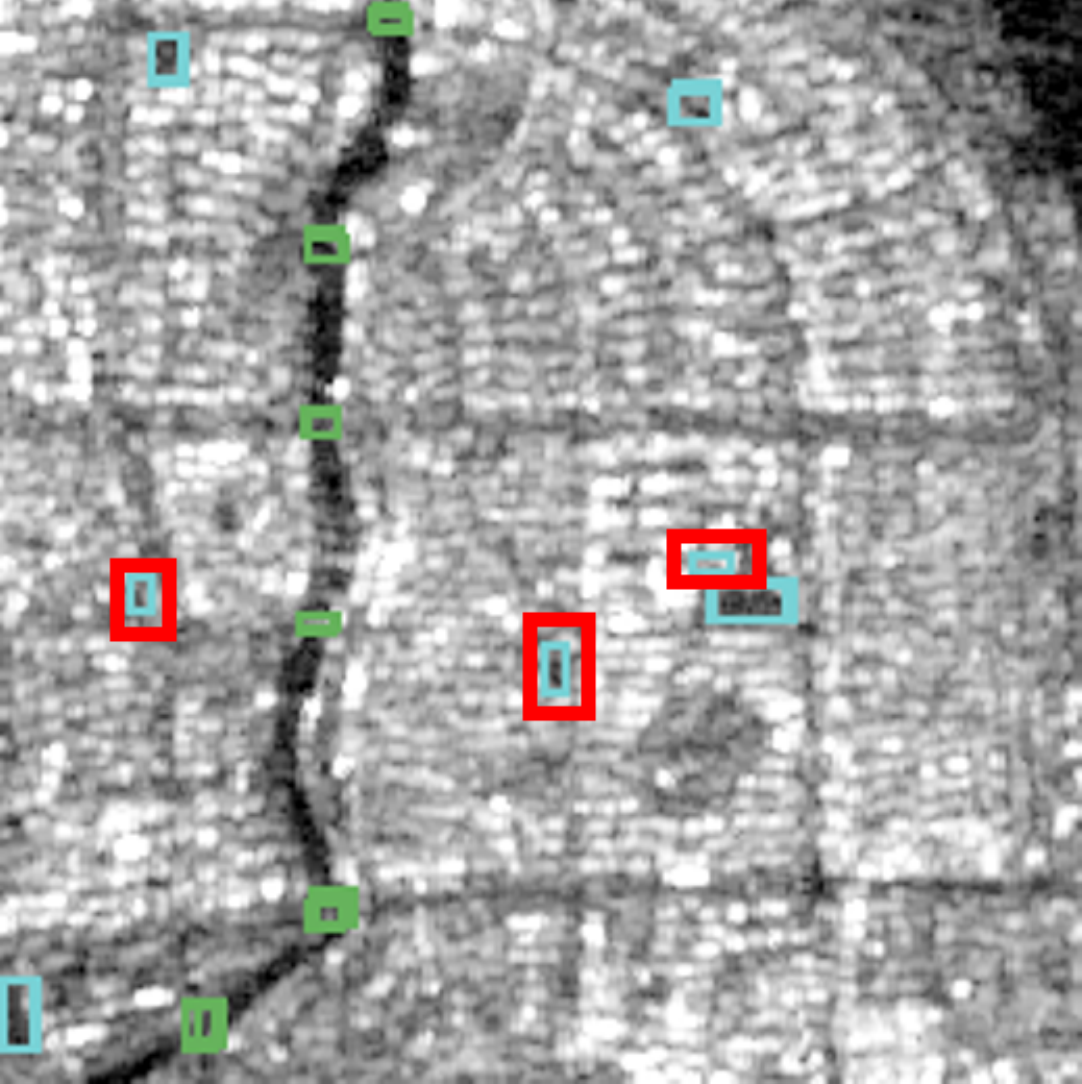} & 
        \includegraphics[width=\linewidth]{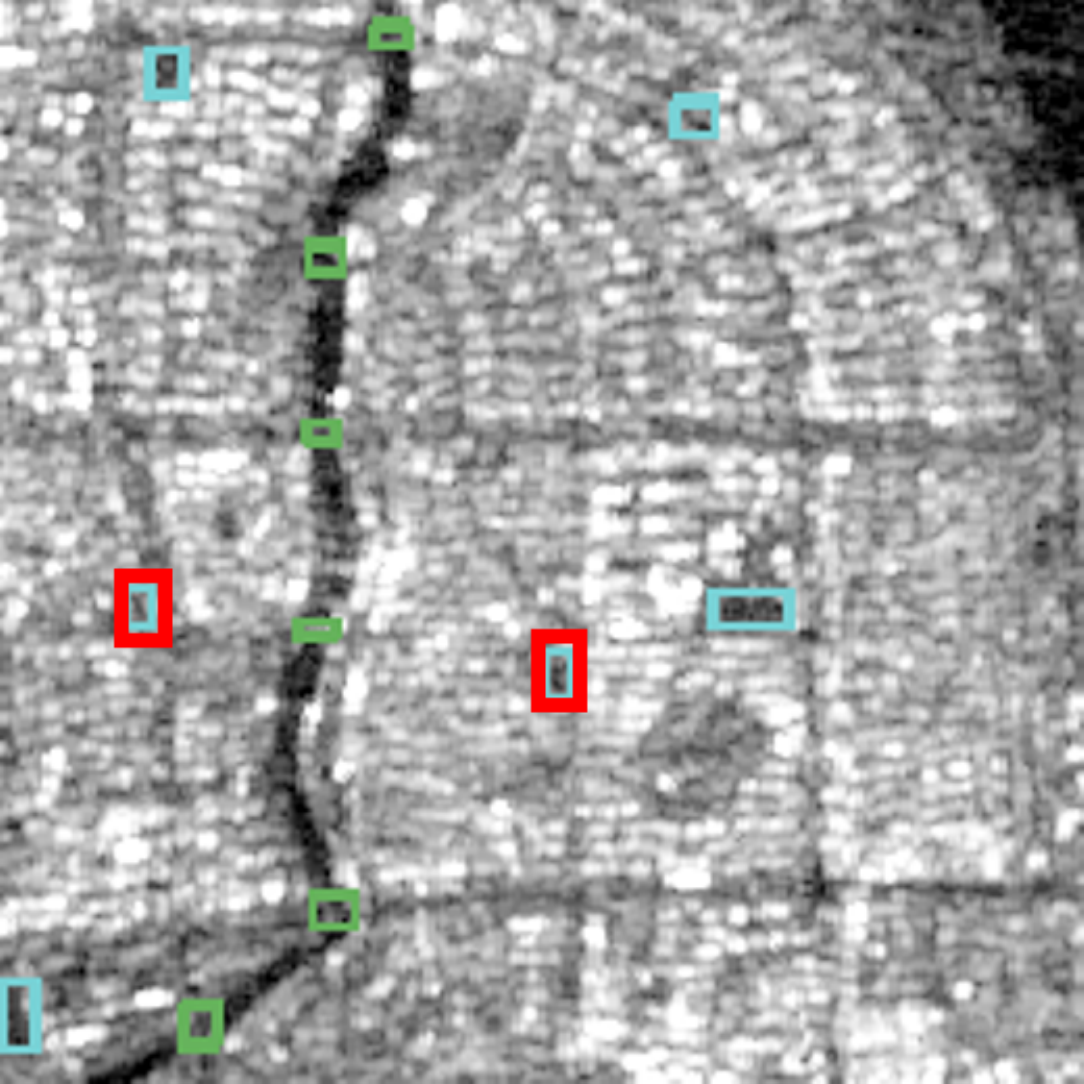} & 
        \includegraphics[width=\linewidth]{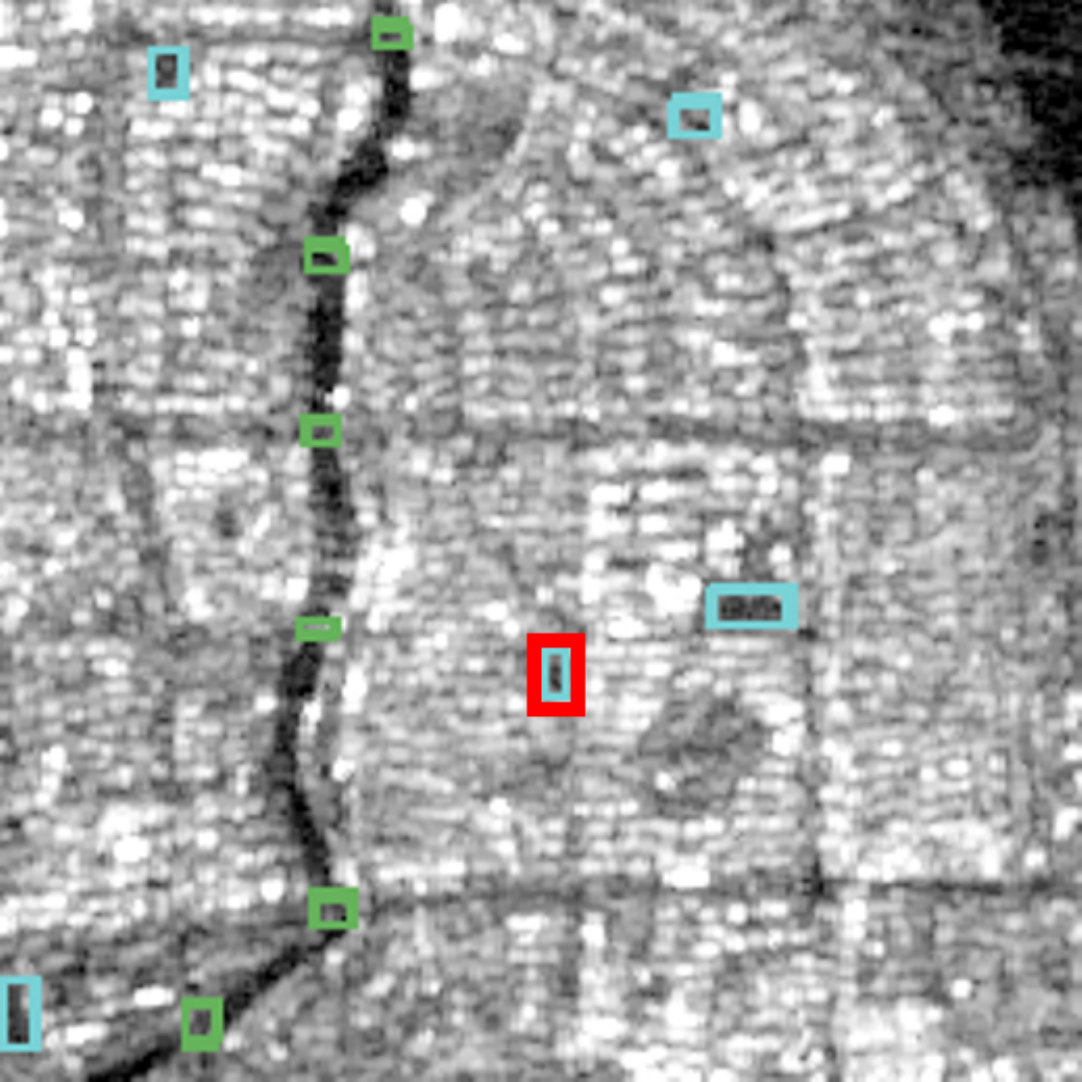} & 
        \includegraphics[width=\linewidth]{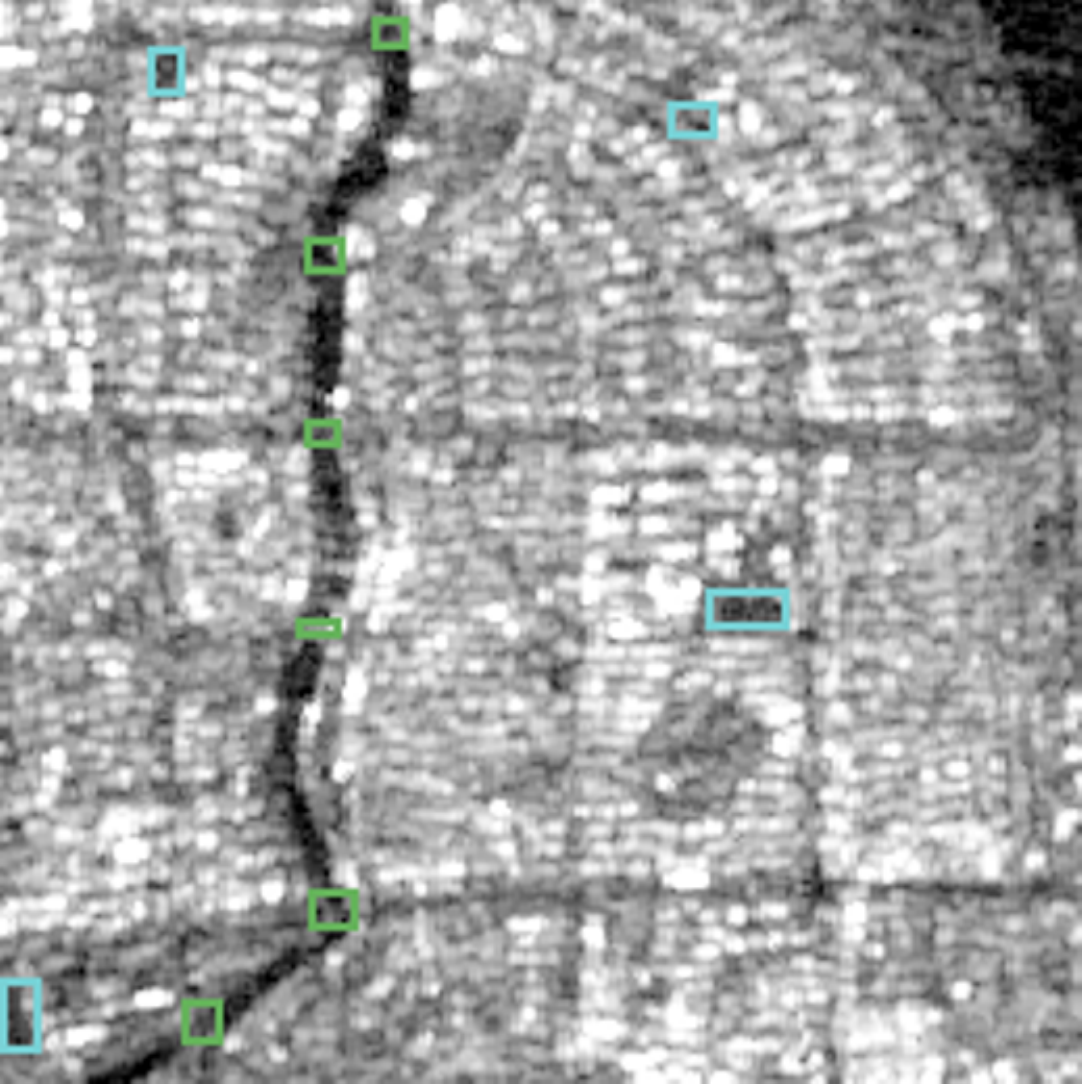} \\

        (a) & (b) & (c) & (d) & (e) \\
    \end{tabular}
    \caption{Illustration of output visualization on the OGSOD-2.0 dataset. (a) Ground truth labels. (b) CFT. (c) ICAFusion. (d) MMIDet. (e) Ours. Yellow dashed boxes indicate objects missed during localization, red boxes represent false detections, and green boxes denote visualization boxes with correct predictions.}
    \label{fig:12}
\end{figure*}

\noindent\textbf{Effectiveness of the DMQA}. 
We first evaluate the DMQA module, which iteratively enhances cross-modal feature associations and quantifies region-wise reliability based on token-representation principles.
As shown in Table~\ref{tab:3} and Table~\ref{tab:4}, integrating DMQA into the baseline increases mAP50/mAP from 80.2\%/42.7\% to 81.3\%/43.5\%.
Under modality missingness, DMQA consistently yields gains. At MR = 0.1, it improves mAP50/mAP by 0.8\%/0.5\% under Zero-filling, and at MR = 0.3 the improvements reach 0.9\%/0.9\%. Even at MR = 0.5, DMQA still delivers 0.9\% and 1.0\% gains. Comparable trends are observed under the INN reconstruction strategy, where the largest gain occurs at MR = 0.3, reaching 1.2\%/1.0\%.

We also visualize the features under the Zero-filling setting with MR=0.3. Comparisons against the ground truth labels in Figure~\ref{fig:5} (a) and Figure~\ref{fig:5} (b) show that the baseline model, shown in Figure~\ref{fig:5} (c), exhibits scattered or misplaced attention and fails to effectively focus on actual target regions. In contrast, our proposed method integrated with the DMQA module, shown in Figure~\ref{fig:5} (d), produces more precise and concentrated attention on the corresponding target regions, including those originally affected by the absence of modalities. A similar improvement can also be observed under the INN setting, where the integration of DMQA enables the model to better utilize the compensated information. These results confirm that the reliability quantification mechanism of DMQA effectively enhances the model’s ability to accurately focus on and identify target regions when modalities are missing.

\noindent\textbf{Effectiveness of the OCNF}. 
We further ablate the OCNF module, which is designed for deep fusion under modality-missing. Zero-filling and reconstruction strategies under random modality-missing conditions often introduce inaccurate signals that conflict with valid cues from the available modality, leading to noticeable inter-modal interference. As shown in Table~\ref{tab:3} and Table~\ref{tab:4}, integrating OCNF into the baseline yields consistent improvements, increasing mAP50/mAP by 0.7\%/0.6\% on the complete dataset. Under the Zero-filling setting, OCNF enhances detection accuracy across all missing rates. At MR = 0.1, the model achieves 79.2\%/41.3\%. When MR increases to 0.3, OCNF brings gains of 0.7\%/0.6\%, and even under the more severe MR = 0.5 condition, additional improvements of 0.6\%/0.8\% are obtained.
Under the INN reconstruction setting, OCNF also demonstrates stable benefits. For instance, at MR = 0.2, OCNF increases mAP50/mAP by 0.6\%/0.8\%, and at MR = 0.4, the improvements reach 0.8\%/0.5\%. These results indicate that OCNF alleviates the adverse effects of inaccurate reconstructed signals and enhances the overall fusion process.

Feature visualizations further support these observations. As shown in Figure~\ref{fig:6} (c), the baseline exhibits clear cross-modal interference, whereas Figure~\ref{fig:6} (e) shows that OCNF preserves inter-modal independence through orthogonally normalized weights. The resulting features present sharper target boundaries and more consistent spatial responses, even in regions affected by missing or degraded modalities. A similar trend is observed under INN reconstruction, where OCNF further mitigates modality entanglement. These results confirm that the orthogonality-enhanced fusion mechanism effectively improves cross-modal feature integration and strengthens the model’s ability to capture reliable target cues under challenging multimodal conditions.

\noindent\textbf{Component-wise Analysis of DMQA}.
We further investigate the contributions of Magnitude and Directional Reliability within the DMQA module. As shown in Figure~\ref{fig:7}, under the Zero-filling setting, incorporating either w/ Length or w/ Direction yields noticeable improvements over the baseline. When MR = 0.0, performance increases from 80\% to 80.5\% and 81.0\%, respectively, while the complete DMQA achieves the highest accuracy at 81.3\%. As the missing rate increases, the performance gap becomes more evident. At MR = 0.3, the baseline drops to 73.8\%, whereas w/ Length, w/ Direction, and the full DMQA reach 74.2\%, 74.5\%, and 74.7\%, respectively. Even under severe corruption at MR = 0.5, the complete DMQA still maintains 71.3\%, outperforming the baseline by 0.9\%.
Consistent tendencies are observed in the INN reconstruction setting, as shown in Figure~\ref{fig:8}. For example, at MR = 0.2, the addition of w/ Length or w/ Direction improves mAP50 from 77.1\% to 77.5\% and 77.6\%, while the complete module achieves 78.0\%. At a higher corruption level of MR = 0.4, the full DMQA continues to offer the best performance at 74.1\%. These results confirm that Magnitude and Directional Reliability are complementary. Their integrated utilization enables more robust cross-modal interaction, particularly under challenging modality-missing conditions.

\noindent\textbf{Hyperparameter Sensitivity of DMQA}. 
To evaluate the sensitivity of DMQA to key hyperparameters, we conduct ablation experiments on the SpaceNet6-OTD-Fog dataset at $\text{MR}=0.3$, investigating the iteration number $I\in\{1,2,3,4,5\}$ and the number of reliability tokens $K\in\{4,8,16,32\}$ under both Zero-filling and INN reconstruction settings. The quantitative results are reported in Figures~\ref{fig:9} and \ref{fig:10}.
With respect to the iteration number, performance consistently improves as $I$ increases and reaches its optimum at $I = 4$ across all token configurations. Beyond this point, the accuracy saturates or slightly declines, indicating that additional iterations provide limited semantic refinement and may introduce redundant updates that weaken feature distinctiveness.
Regarding the token number $K$, the performance exhibits a non-monotonic trend. When the token count is small, such as $K = 4$ or $K = 8$, the representational capacity becomes insufficient and the results degrade accordingly. When the token count is very high, for example $K = 32$, the model introduces redundant and noisy representations, which again leads to performance deterioration. Across all configurations, the setting with $K = 16$ and $I = 4$ achieves the highest accuracy under both conditions, providing an effective balance between representational expressiveness and computational cost.

\noindent\textbf{Visualization of Detection Results}.
To further illustrate the detection capability of the proposed method, we conduct qualitative comparisons between QDFNet and three representative multimodal object detectors. Visualization results on the SpaceNet6-OTD-Fog and OGSOD-2.0 datasets are presented in Figure~\ref{fig:11} and Figure~\ref{fig:12}. As observed, existing methods frequently miss small objects, produce false alarms, or exhibit imprecise localization. Such failures mainly originate from image degradation: RGB images often suffer from occlusions and low contrast due to haze, cloud cover, or illumination variations, whereas SAR images are affected by speckle noise and ghosting artifacts that distort structural cues.

In contrast, QDFNet achieves more reliable detections under these challenging conditions. It can accurately identify small, partially occluded, or visually similar targets, even when one modality provides ambiguous or degraded information. These results demonstrate that QDFNet effectively exploits complementary multimodal cues and maintains discriminative spatial features, leading to more stable and accurate detections in complex imaging environments.

\section{CONCLUSION}
In this work, we propose a novel Quality-aware Dual-Fusion Network (QDFNet), a robust framework for optical-SAR remote sensing object detection under modality degradation and partial modality missing. The proposed QDFNet incorporates a Dynamic Modality Quality Assessment module that explicitly models region-wise modality reliability based on token representations, enabling adaptive suppression of unreliable regions and effective mitigation of degraded or missing modality information. In addition, we design an Orthogonally Constrained Normalized Fusion module that enforces orthogonality during cross-modal feature integration, significantly reducing feature redundancy while preserving complementary information from different modalities.
Extensive experiments on multiple public optical-SAR benchmarks demonstrate that QDFNet consistently outperforms existing multimodal object detection methods, particularly under challenging missing-modality conditions. These results confirm the robustness and effectiveness of the proposed framework for practical remote sensing scenarios, where data quality and modality availability are often uncertain.
In future work, we will investigate extending QDFNet to additional multimodal remote sensing tasks and sensing modalities, as well as exploring lightweight model designs to facilitate deployment in large-scale or resource-constrained remote sensing applications.

\bibliographystyle{unsrt}  
\bibliography{ref}


\ifCLASSOPTIONcaptionsoff
  \newpage
\fi

\end{document}